\documentclass{article}

\PassOptionsToPackage{numbers, compress}{natbib}

 \usepackage[eandd,preprint]{neurips_2026}

\usepackage[utf8]{inputenc} 
\usepackage[T1]{fontenc}    
\usepackage{hyperref}       
\usepackage{url}            
\usepackage{booktabs}       
\usepackage{amsfonts}       
\usepackage{amsmath}
\usepackage{amssymb}
\usepackage{cleveref}
\usepackage{nicefrac}       
\usepackage{microtype}      
\usepackage{xcolor}         

\usepackage{graphicx}
\usepackage{tcolorbox}
\tcbuselibrary{breakable}
\usepackage{colortbl}
\usepackage{tikz}
\usepackage{tablefootnote}
\usepackage{threeparttable}
\usepackage{float}
\usepackage{multirow}
\usepackage{pgf}
\usepackage[table]{xcolor}
\usepackage{pifont}
\usepackage{comment}
\usepackage{fontawesome5}
\usepackage{xspace}
\usepackage{enumitem}
\usepackage{eccvabbrv}  
\usepackage{wrapfig}    

\newcommand{\scc}[1]{%
  \pgfmathparse{(#1)*100}%
  \colorlet{scccolor}{green!70!black!\pgfmathresult!red!70!black}%
  {\color{scccolor}\textbf{#1}}%
}

\newcommand{\cmark}{{\color{green!60!black}\ding{51}}}
\newcommand{\xmark}{{\color{red!70!black}\ding{55}}}

\newcommand{\model}{\textsc{VFig}\xspace}
\newcommand{\dataset}{\textsc{VFig-Data}\xspace}
\newcommand{\bench}{\textsc{VFig-Bench}\xspace}
\newcommand{\benchood}{\textsc{VFig-Bench-OOD}\xspace}
\newcommand{\compds}{\textsc{VFig-Data}-Complex-Diagrams\xspace}
\newcommand{\progds}{\textsc{VFig-Data}-Shapes-and-Arrows\xspace}

\definecolor{linkblue}{RGB}{86, 180, 233}

\title{\includegraphics[height=0.8em]{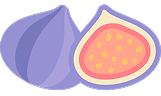}\ \model{}: Vectorizing Complex Diagrams in SVG with Vision-Language Models}

\author{%
  Qijia He$^{\star,1}$ \quad Xunmei Liu$^{\star,1}$ \quad Hammaad Memon$^{\star,1}$ \quad Ziang Li$^{\star,1}$ \quad Zixian Ma$^{\star\dagger,1,2}$ \\
  Jaemin Cho$^{1,2}$ \quad Zhongzheng Ren$^{1,2,3}$ \quad Dan Weld$^{1,2}$ \quad Ranjay Krishna$^{1,2}$ \\[0.3em]
  $^1$University of Washington \quad $^2$Allen Institute for Artificial Intelligence \quad $^3$UNC-Chapel Hill\\
}

\begin{document}

\maketitle



\begin{abstract}

Scalable Vector Graphics (SVG) are essential for technical illustration and digital design, offering resolution independence and semantic editability. In practice, original vector files are frequently lost, leaving only rasterized versions (\eg, PNG, JPEG) that resist modification, while manual reconstruction is prohibitively expensive.
Progress on automating raster-to-SVG conversion has been bottlenecked by two gaps: existing SVG datasets are dominated by icons and decorative graphics that lack the complexity of professional diagrams, and existing benchmarks rely on pixel- or embedding-level similarity that fails to capture structural correctness (\eg, broken connectivity, misplaced arrows). We close both gaps with paired contributions targeting \emph{diagram-centric} figures (\eg, model architectures, flowcharts, schematics).
For training, we introduce \textbf{\dataset}, the largest figure-to-SVG dataset of its kind at 66K pairs, combining real paper figures converted via a describe-and-generate pipeline with programmatic diagrams that supply noise-free supervision over arrow styles, fonts, and geometry.
For evaluation, we introduce \textbf{\bench}, a structure-aware evaluation suite, paired with \textbf{\benchood{}}, an out-of-distribution set of figures manually curated from highly cited arXiv papers. Beyond pixel and embedding similarity, our protocol reports rubric-based VLM-Judge scores and Elo ratings from pairwise human preference evaluation.
Built on these contributions, VFIG is a VLM family trained with a simple-to-complex SFT curriculum followed by RL with rendering-aware rewards. VFIG achieves state-of-the-art open-source performance, outperforming the best open-source VLM baseline by over $30\%$, and matches Claude Sonnet 4.6 on VFIG-BENCH: Gemini-Judge $78.2\%$ vs. $76.7\%$ and GPT-Judge $87.5\%$ vs. $87.4\%$. It remains slightly behind the strongest proprietary models GPT-5.2 and Gemini-3.

\end{abstract}

\begin{figure}[!h]
    \centering
    \includegraphics[width=\linewidth]{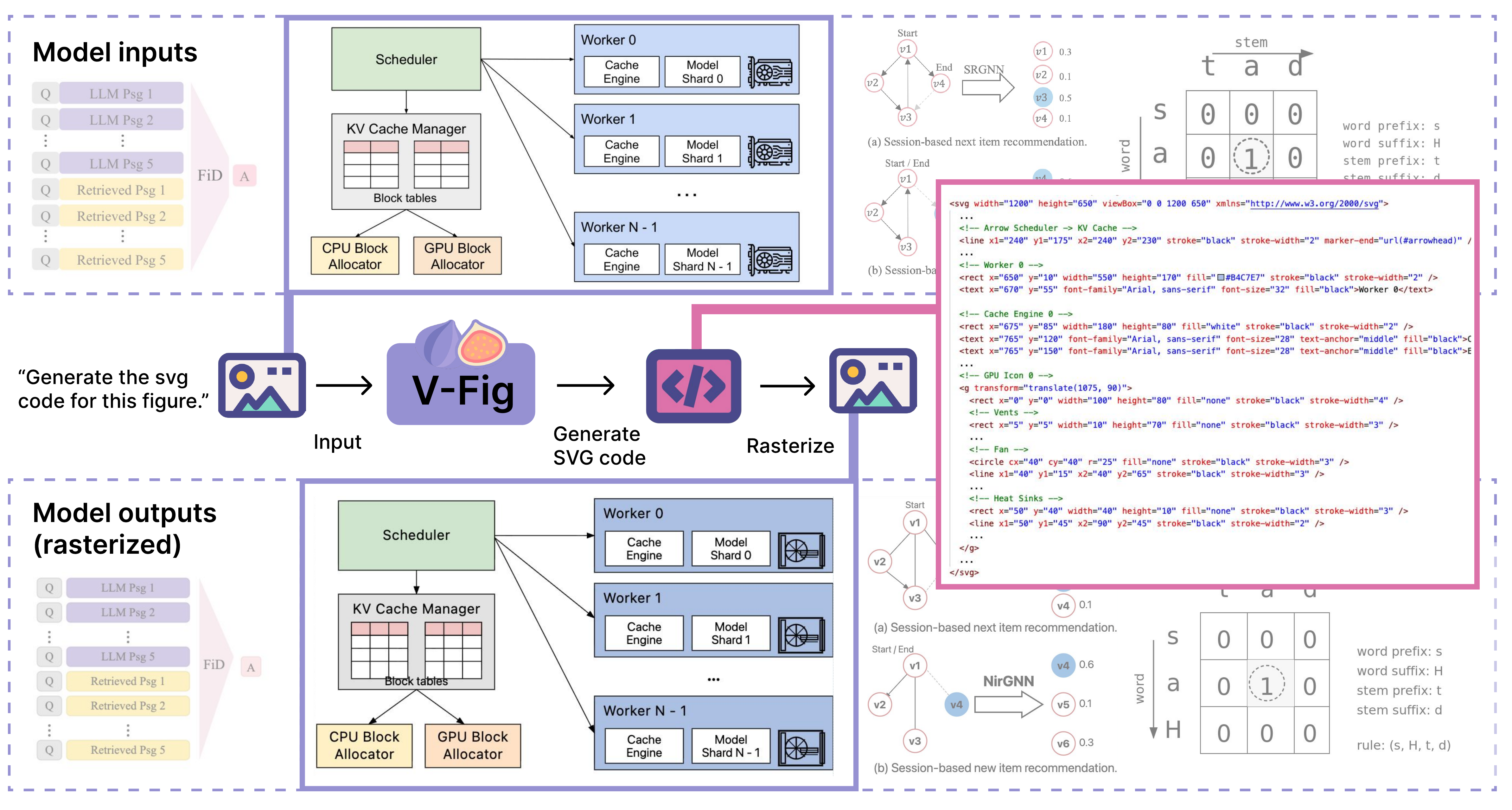}
    \vspace{-2mm}
    \caption{\textbf{Overview of \model}. Given complex raster images (top row) as input, \model generates editable, high-fidelity SVG code (pink box). Rendering the generated SVG (bottom row) produces outputs that closely match the inputs.}
    \vspace{-0.8cm}
    \label{fig:teaser}
\end{figure}
\section{Introduction}
\label{sec:intro}
Scalable Vector Graphics (SVG) serve as a cornerstone of technical illustration and digital design, offering resolution independence, semantic editability, and a text-based structure amenable to both human editing and machine generation, all within a W3C-standard format supported by modern browsers and major graphics editors. From scientific research and engineering to education, design, and media, SVG figures distill complex ideas, processes, and relationships into precise visual forms. Widely recognized AI architectures such as ResNet~\cite{he2016resnet} and Transformers~\cite{vaswani2017attention} are often remembered through their canonical diagrams. These figures combine nested layouts, heterogeneous primitives, precise alignments, and intricate connectivity. Yet in practice, original SVG sources are frequently lost or inaccessible, leaving only flat raster versions (\eg, PNG, JPEG) that resist modification, scaling, and reuse, while manual reconstruction is prohibitively labor-intensive and requires specialized expertise.

Automating this raster-to-SVG conversion would unlock significant practical value: accelerating revision workflows, lowering the barrier to professional visualization, and enabling faithful reuse across platforms. The task requires joint reasoning over visual content (\eg, spatial layouts, styling, and compositional hierarchy) and the structured code needed to reproduce it. This naturally lends itself to a vision-language modeling formulation. We therefore propose a VLM-based approach that takes a raster image as input and generates editable SVG code (Fig.~\ref{fig:teaser}).

Existing figure-to-SVG approaches span classical contour-tracing methods~\cite{selinger2003potrace, vtracer}, learning-based techniques~\cite{deepsvg,diffvg,im2vec}, and more recent LLM/VLM-based generation methods~\cite{starvector,llm4svg,omnisvg, reason-svg, rlrf}, but they are predominantly developed and evaluated on relatively simple graphics such as icons or small diagrams. It remains unclear how well they scale to structurally complex diagram-centric figures encountered in practice, such as model architectures, flowcharts, and hierarchical diagrams. These are precisely the figures where automated reconstruction would be most valuable. However, no public training set or benchmark specifically targets this setting, and open-source VLMs still lag substantially behind proprietary systems such as Gemini-3 and GPT-5.2 on these figures.

Closing this gap requires addressing the unique difficulties of diagram-centric figures. Unlike natural images or isolated icons, scientific diagrams exhibit both high structural complexity (\eg, nested groups, hierarchical layouts) and high element density (\eg, numerous geometric primitives and connectors). These properties introduce four key challenges for figure-to-SVG generation. First, not all visual content is suitable for vectorization, such as natural images, heavy textures, or mathematical equations, requiring careful data curation. Second, SVG sequences grow rapidly with diagram complexity, making long-horizon generation and syntactic consistency difficult. Third, compositional figures with repeated modules and precise alignments are substantially harder to learn from scratch than isolated symbols or icons. Finally, fine-grained geometric and stylistic details are difficult to recover through token prediction alone without visual-level feedback.
Our central contributions are a new training dataset paired with a held-out evaluation suite and an open-source model trained on the dataset that bridges this open-vs-proprietary gap. Concretely, we introduce:

\noindent\textbf{Training dataset (\dataset).} A large-scale dataset of 66K complex diagram-centric figure--SVG pairs from real-world paper diagrams and procedurally generated diagrams. Our pipeline (i) filters out figures unsuitable for vectorization (\eg, natural images, math equations, plots, etc.) via VLM-based image classification, (ii) converts retained figures with a describe-and-generate procedure that optimizes for visual fidelity, and (iii) applies code-level filtering that favors semantic primitives (e.g. \texttt{<rect>}, \texttt{<line>}) over free-form paths to control for editability and sequence length.

\noindent\textbf{Evaluation benchmark (\bench).} An evaluation suite built on the held-out split of \compds{} (\bench) and a manually curated out-of-distribution benchmark (\benchood). The suite follows a coarse-to-fine protocol with three complementary granularities: \emph{pixel-level} metrics (\eg, LPIPS~\cite{zhang2018unreasonable}) for visual fidelity, \emph{component-level} scores (rule-based arrow and shape matching) for structural correctness, and \emph{image-level} VLM judgments (Gemini~\cite{google_gemini3_2025}, GPT~\cite{openai_gpt5_2_2025}) for holistic quality.

\noindent\textbf{Open-source model (\model).} On top of these datasets we train and evaluate \model{}, a family of VLMs with a simple-to-complex SFT curriculum (primitive-level generation first, then multi-panel and hierarchical compositions) followed by RL with rendering-aware, structure-focused rewards over alignment, grouping, connectivity, and layout.

In experiments (Sec.~\ref{sec: experiments}), \model{} achieves the strongest open-source results on \bench{} and two existing diagram-centric benchmarks, Molmo2-Diagram~\cite{Molmo2} and SVG-Diagrams~\cite{starvector}, outperforming the best open-source VLM baseline by over $30\%$. On \bench{}, it reaches Gemini-Judge and GPT-Judge scores of $78.2\%$ and $87.5\%$, respectively, compared with $76.7\%$/$87.4\%$ for Claude-Sonnet-4.6 and $91.5\%$/$94.8\%$ for Gemini-3-Pro, substantially narrowing the gap to proprietary systems. Our ablations further show that the simple-to-complex SFT curriculum significantly improves rendering success. RL with rendering-aware rewards brings additional gains on pixel-level metrics (SSIM, LPIPS), image-level metrics (VLM-Judge scores), as well as clean and render rate. Rubric-based VLM-judge rewards also produce larger structural improvements than lower-level reward objectives.

\begin{figure}[!t]
    \centering
    \includegraphics[width=\linewidth]{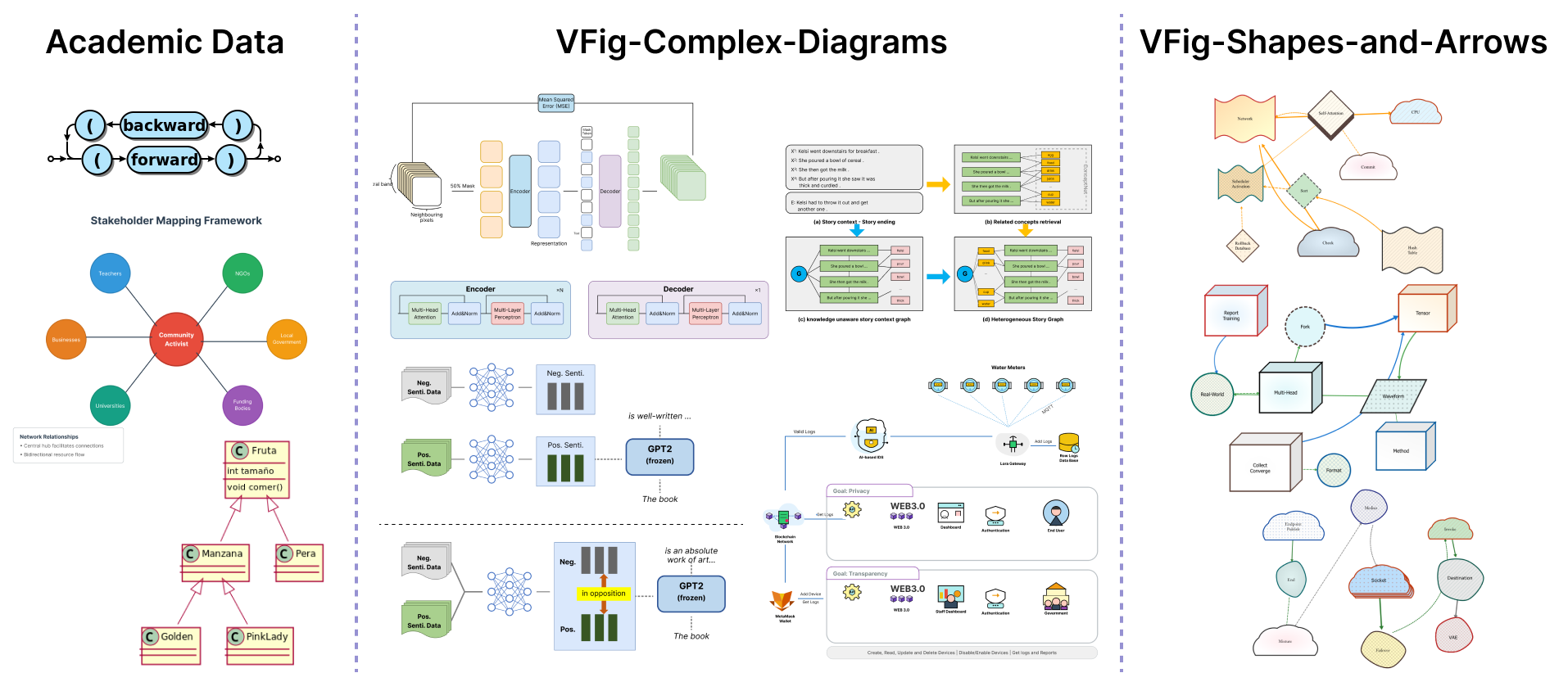}
    \caption{\textbf{Examples of \dataset and academic data.} We show the three sources: simple diagrams from academic datasets, complex diagram layouts, and a curated set of basic shapes and arrows to support structured SVG generation.}
    \vspace{-7mm}
    \label{fig:training_data}
\end{figure}
\section{Related Work}
\label{sec: related_work}

We discuss the most closely related LLM/VLM-based SVG generation methods and diagram-centric datasets here. Earlier non-LLM approaches, including tracing-based vectorization, differentiable rendering, and sequential decoder models, are reviewed in Appendix~\ref{appendix:extended_related_work}.

\noindent\textbf{LLM/VLM-based SVG and Diagram Code Generation.}
Recent work reframes SVG generation as multimodal code synthesis. StarVector~\cite{starvector} trains a multimodal LLM, LLM4SVG~\cite{llm4svg} introduces semantic command tokens, OmniSVG~\cite{omnisvg} unifies SVG token and coordinate generation, and InternSVG~\cite{wang2026internsvgunifiedsvgtasks} unifies SVG understanding, editing, and generation. Other reasoning- and feedback-based extensions incorporate explicit design rationales~\cite{reason-svg}, RL from rendering feedback~\cite{rlrf}, visual self-feedback~\cite{liang2026renderintheloopvectorgraphicsgeneration}, or internal visual guidance~\cite{zhang2025duetsvgunifiedmultimodalsvg}. Related directions cover human-readable command prediction~\cite{zhang2023beyondpixels}, diffusion or optimization-based text-to-SVG~\cite{Jain_2023_CVPR,Xing_2024_CVPR}, hybrid LLM--diffusion pipelines~\cite{Wu_2025_CVPR}, text-guided icon synthesis~\cite{wu2023iconshop}, and text-to-TikZ figure generation with RL feedback~\cite{greisinger2026tikzillascalingtexttotikzhighquality}. The closest peers to our recipe, RLRF~\cite{rlrf} and Reason-SVG~\cite{reason-svg}, target generic SVG rather than diagrams in their data and reward design, and have not released code or weights at submission time. More broadly, these methods face open challenges in long-horizon consistency, syntactic validity, and editability.

\noindent\textbf{Datasets \& Evaluation.}
Existing SVG datasets~\cite{starvector,llm4svg,omnisvg,wang2026internsvgunifiedsvgtasks,clouatre2019figr,zou2024vgbench} largely focus on icons, emojis, and general SVG tasks, with limited emphasis on diagram-centric figure-to-SVG conversion. Paper2Fig~\cite{ocr-vqgan} offers 100K arXiv figures with architecture and methodology diagrams but is curated for text-aware reconstruction rather than figure-to-SVG, lacking paired SVG code. Closely related, SliDer~\cite{hazimeh2025semanticdocumentderenderingsvg} reconstructs editable SVGs from raster document images, and DaVinci~\cite{zeng2026davinci} parses scientific diagrams into structured visual representations. We introduce \dataset{} and \bench{} to fill this gap, providing a diverse training set and unified evaluation suite for diagram-centric figure-to-SVG.

\begin{figure}[t]
    \centering
    \includegraphics[width=\linewidth]{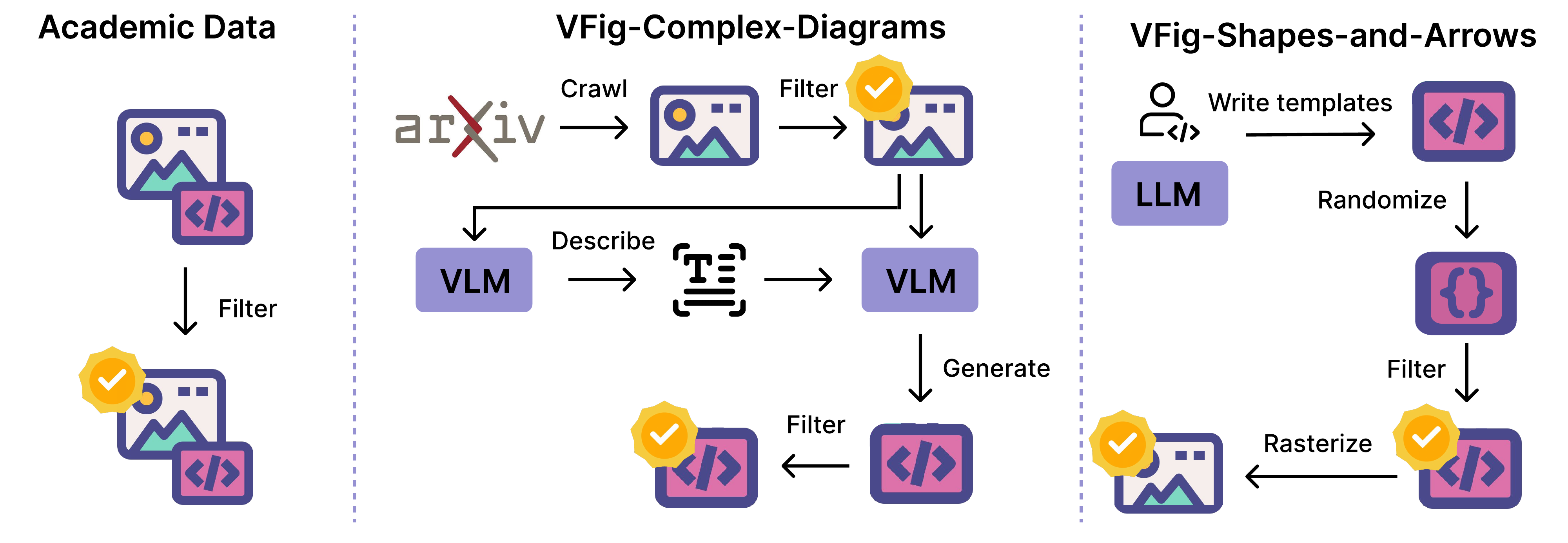}
    \caption{\textbf{Data generation and filtering pipelines.} We show the data generation and filtering processes for curated academic figures, complex diagrams created through a VLM-based describe-and-generate pipeline from crawled images, and shapes and arrows produced by LLM-generated templates with randomized elements.}
    \vspace{-0.7cm}
    \label{fig:data_gen}
\end{figure}
\section{\dataset}
\label{sec: dataset}

To enable realistic diagram-centric figure-to-SVG generation, we curate \dataset, a large-scale dataset of 66K filtered image-SVG pairs. Unlike prior SVG datasets that focus predominantly on icons or decorative graphics, \dataset targets diagram-centric figures (\eg, model architectures, flowcharts, schematics) (Fig.~\ref{fig:training_data}). To our knowledge, \dataset{} is the first dataset of this scale purpose-built for structured diagram-centric figure generation. We describe our data generation pipeline (Sec.~\ref{sec:data_generation}), filtering procedure (Sec.~\ref{sec:data_filtering}), and the full training mixture including 78K additional samples from academic SVG datasets (Sec.~\ref{sec:dataset_stats}).

\subsection{Data Generation}
\label{sec:data_generation}
Specifically, \dataset contains two complementary subsets: (1) \compds: real-world diagram-centric figures from scientific papers collected as raster images and converted into structured SVG (Fig.~\ref{fig:training_data}, center), and (2) \progds: programmatically generated diagrams featuring diverse shapes, connectors, and spatial layouts (Fig.~\ref{fig:training_data}, right). We develop a dedicated data pipeline for each source, detailed below.

\noindent\textbf{\compds.}
Scientific papers are a rich source of complex diagrams, so we develop an automated pipeline to collect diagram-centric figures from \text{arXiv} papers published after January~2025. For each paper, we extract referenced figure assets from the \LaTeX{} source by parsing \texttt{\textbackslash includegraphics} occurrences, keeping PDF/PNG/JPG/JPEG formats and rasterizing embedded PDFs via \texttt{PyMuPDF} for a unified representation (Additional details in Appendix~\ref{appendix:data}). 

Then, we convert them to structured SVG with a two-step \emph{describe-and-generate} pipeline (Fig.~\ref{fig:data_gen}, center): the \emph{describe} step prompts a VLM to produce a structured description of geometric elements, text, spatial layout, and inter-object relationships that mirrors SVG primitives; the \emph{generate} step then conditions the VLM on both the original image and this description to produce SVG code. From a side-by-side comparison of $20+$ VLM backbones in an internal sandbox, a unified Gemini-3-Pro describe $\rightarrow$ Gemini-3-Pro generate pipeline is preferred in $88.7\%$ of pairwise human ratings and adopted as our final configuration. Full ablations against single-pass generation are in Appendix~\ref{sec:compds-generation}. When training, we use the rendered version of the SVG generated by this pipeline to ensure the ground truth SVG exactly matches the input image. While this practice guarantees pixel-level image--SVG alignment, it does not fix structural issues in the generated code itself: residual errors remain, most visibly in arrow connectivity. We partially mitigate these with code-level filtering and our noise-free \progds{}, and leave connectivity-aware sample filtering for future work.


\noindent\textbf{\progds.}
\label{sec:programmatic_figure_gen}
To complement \compds{} with noise-free supervision over fine-grained attributes (\eg arrow styles, fonts, fill patterns, geometric variations) that are hard to recover from raster images alone, we construct \progds, a programmatically generated dataset (Fig.~\ref{fig:training_data}, right). Diagrams are synthesized directly in SVG using 19 layout templates with randomized shapes, arrows, fonts, and styles then rendered to form paired image--SVG data; every visual attribute is recorded as structured metadata, providing precise ground truth (additional details in Appendix~\ref{app:progds}).

\subsection{Data Filtering} 
\label{sec:data_filtering}
To ensure high-quality training data, we apply two rigorous filters: (1) \textit{image filtering}, which removes figures unsuitable for vectorization, and (2) \textit{Code filtering}, which discards outputs dominated by free-form paths that would cause token explosion due to verbose coordinate definitions. We provide more details about filtering in Appendix~\ref{app: img_filter} and \ref{app:code_filtering}.

\noindent\textbf{Image Filtering.}
In image filtering, we remove figures dominated by natural images or math equations (which are better represented in \LaTeX{}), as well as plots and tables. We use Gemini-3 Flash to classify each figure into one of four categories: \texttt{KEEP}, \texttt{IMAGE}, \texttt{MATH}, and \texttt{PLOT}, where \texttt{KEEP} allows all types of diagrams with a small portion of images and math notations. We retain only figures classified as \texttt{KEEP}. We apply this filtering to both newly collected arXiv figures and the Paper2Fig~\cite{ocr-vqgan} figures to obtain clean, diagram-centric samples for \compds.

\noindent\textbf{Code Filtering.}
Beyond image-level filtering, we apply SVG code filtering to remove \texttt{<path>}-dominated or structurally noisy outputs. We prioritize reducing \texttt{<path>} elements for two reasons: (1) they often contain extremely long coordinate sequences with excessive floating-point precision, leading to prohibitive token counts under a VLM's tokenizer; and (2) they frequently bundle semantically distinct elements into monolithic definitions, hindering downstream editability. 
Concretely, we apply a ratio-based heuristic that retains diagrams dominated by structural primitives and connectors while removing path-heavy artistic SVGs, followed by light cleaning to standardize formatting and discard corrupted samples. 

\subsection{Training Mixture \& Statistics}
\label{sec:dataset_stats}

Beyond \compds and \progds, we incorporate two academic SVG datasets to strengthen primitive-aware generation: SVG-Diagrams~\cite{starvector}, filtered to \texttt{<text>}-bearing diagram-style layouts, and Molmo2-Diagram, the diagram-focused subset of Molmo2-SynMultiImageQA~\cite{Molmo2}. Table~\ref{tab:data_sources} summarizes all four sources along four metrics:
\textbf{Structural Complexity} (per-SVG count of geometric elements) and
\textbf{Element Complexity} (geometric + text elements, log-scaled) capture diagram size and density;
\textbf{SVG Cleanliness} (share of semantic primitives like \texttt{rect}, \texttt{circle}, \texttt{line}) and
\textbf{Path Dominance} (share of tracing-style \texttt{path}/\texttt{polygon} elements) capture code quality, where higher Cleanliness and lower Path Dominance indicate more structured, editable SVG.
Formal definitions of these metrics can be found in Appendix~\ref{app: svg complexity metrics}.

\begin{table*}[t]
\centering
\caption{\textbf{Summary of the training mixture}. We present the statistics of all datasets used in model training, highlighting the higher structural complexity in \compds.}
\setlength{\tabcolsep}{4pt}
\renewcommand{\arraystretch}{1.15}
\resizebox{\linewidth}{!}{%
\begin{tabular}{
l | c c c |
@{\hspace{6pt}} c
@{\hspace{6pt}} c
@{\hspace{6pt}} c
@{\hspace{6pt}} c
}
\textbf{Data Source}
& \shortstack{\textbf{Image}\\\textbf{Filtering}}
& \shortstack{\textbf{Code}\\\textbf{Filtering}}
& \shortstack{\textbf{Final}\\\textbf{Size}}
& \shortstack{\textbf{Structural}\\\textbf{Complexity}}
& \shortstack{\textbf{Element}\\\textbf{Complexity}}
& \shortstack{\textbf{SVG}\\\textbf{Cleanliness}}
& \shortstack{\textbf{Path}\\\textbf{Dominance}} \\
\midrule
(academic) SVG-Diagrams\cite{starvector}
& \xmark & \cmark & 27.6k
& 32.0 & 3.4 & 0.8 & 0.2 \\

(academic) Molmo2-Diagram \cite{Molmo2}
& \cmark & \cmark & 51.2k
& 26.9 & 3.3 & 0.9 & 0.1 \\

\progds
& \xmark & \cmark & 6.5k
& 22.2 & 3.7 & 0.5 & 0.5 \\

\compds
& \cmark & \cmark & 60.0k
& 55.3 & 4.0 & 0.8 & 0.2 \\
\end{tabular}%
}
\label{tab:data_sources}
\vspace{-3mm}
\end{table*}
\section{\model{} Model}
\label{sec: method}

Given a figure image $x$, \model{} produces a structured SVG program $y$ that reconstructs the visual content while preserving compositional structure. We train with supervised fine-tuning followed by reinforcement learning with rendering-aware rewards (full objectives, rubric details, and hyperparameters in Appendix~\ref{appendix:setup}).

\noindent\textbf{Simple-to-complex SFT.}\label{sec:sft}
Direct SFT on complex diagram-centric figures destabilizes optimization, since the model must simultaneously learn low-level primitive generation and high-level compositional reasoning. We instead train in two stages: (i) on structurally simpler diagrams (SVG-Diagrams~\cite{starvector}, Molmo2-Diagram~\cite{Molmo2}, and \progds) to establish primitive-level generation and basic layout understanding, then (ii) on \compds{} for compositional reasoning and structural fidelity.

\noindent\textbf{RL with rendering-aware rewards.}\label{sec:rl_reward}
SFT optimizes token-level likelihood, leaving perceptual layout and connectivity errors unaddressed. We further optimize the SFT policy with GRPO~\cite{deepseekmath}. Each sampled SVG is rendered via CairoSVG and scored by a rubric-based VLM judge (Gemini-3-Flash) on four axes: \textbf{Presence} ($r_{\text{pres}}$, required elements), \textbf{Layout} ($r_{\text{layout}}$, alignment), \textbf{Connectivity} ($r_{\text{conn}}$, arrow endpoints), and \textbf{Details} ($r_{\text{det}}$, text and styling). The final reward is the average of the four scores, $R = \tfrac{1}{4}(r_{\text{pres}} + r_{\text{layout}} + r_{\text{conn}} + r_{\text{det}})$. Invalid SVGs receive zero reward, naturally penalizing malformed outputs. For each input $x$, GRPO samples $G$ candidates $\{\hat{y}_i\}_{i=1}^{G}$, computes per-sample rewards $R_i = R(x,\hat{y}_i)$, and forms group-normalized advantages $A_i = (R_i - \mu)/\sigma$ from the group's reward mean $\mu$ and standard deviation $\sigma$, propagating structural credit jointly across the four axes. The policy is updated under KL regularization against the frozen SFT reference $\pi_{\text{ref}}$:
\begin{equation}
\mathcal{L}_{\text{GRPO}} = \mathbb{E}_{x\sim\mathcal{D},\,\{\hat{y}_i\}\sim\pi_\theta}\!\left[\frac{1}{G}\sum_{i=1}^{G} A_i \log \pi_\theta(\hat{y}_i \mid x)\right] - \beta\,\mathrm{KL}\!\left(\pi_\theta(\cdot\mid x) \,\|\, \pi_{\text{ref}}(\cdot\mid x)\right).
\label{eq:grpo}
\end{equation}
We use a rubric VLM judge rather than pixel metrics (\eg, SSIM) because diagram fidelity requires structural correctness beyond perceptual similarity. For example, a misplaced arrow may have little effect on SSIM while significantly changing the diagram semantics. On 100 annotated examples, our judge correlates with human ratings at $r{=}0.89$ overall (per-axis rubric details and Pearson breakdown in Appendix~\ref{app:training_objectives}). During RL, we randomly sample training examples from the same SFT training data, without introducing additional RL-specific data.

\section{Experiments}
\label{sec: experiments}
We evaluate \model along four axes: VLM capability on diagram-centric figure-to-SVG, the effect of simple-to-complex curriculum SFT, the contribution of RL with visual feedback, and the impact of reward designs, including reward signals' source and granularity.

\subsection{Experimental Setup}
\label{sec:exp_setup}

\noindent\textbf{Training.}
We fine-tune recent VLMs, including Qwen3-VL-4B/8B~\cite{bai2025qwen3vltechnicalreport}, Qwen2.5-VL-3B~\cite{bai2025qwen25vltechnicalreport}, and InternVL3.5-4B~\cite{zhu2025internvl3exploringadvancedtraining}, using LoRA-based parameter-efficient fine-tuning~\cite{hu2021loralowrankadaptationlarge} applied only to the language model. We then run GRPO~\cite{deepseekmath} on the SFT checkpoints with $n{=}8$ rollouts per prompt, using KL regularization against the SFT reference and an entropy bonus. Additional training and compute details (full hyperparameters, GPU configurations, and wall-clock times) are in Appendix~\ref{appendix:setup}.

\begin{table*}[t]
\centering
\caption{\textbf{Benchmark results across three datasets.} Scores are in $[0,1]$, displayed $\times 100$ for readability. \textbf{ViSim}: visual similairity: avg cosine similarity of DINO/CLIP/SigLIP embeddings; \textbf{Gemini} / \textbf{GPT}: VLM-as-judge structural fidelity scores from Gemini-3-flash and GPT-5.2 respectively; \textbf{Clean} and \textbf{Render}: SVG cleanliness and successful rendering rate. Best among open-source VLMs in \textbf{bold}, second-best \underline{underlined}; OmniSVG-4B and Starvector-8B use their official default decoding.}
\label{tab:main_benchmark}
\small
\setlength{\tabcolsep}{2pt}
\renewcommand{\arraystretch}{1.3}
\newcommand{\graytxt}{\color{black!55}}
\resizebox{\textwidth}{!}{
\begin{tabular}{lrrrrrrr|rrrrrrr|rrrrrrr}

& \multicolumn{7}{c|}{\textbf{\bench}}
& \multicolumn{7}{c|}{\textbf{Molmo2-Diagram~\cite{Molmo2}}}
& \multicolumn{7}{c}{\textbf{SVG-Diagrams~\cite{starvector}}} \\

\cmidrule(lr){2-8}
\cmidrule(lr){9-15}
\cmidrule(lr){16-22}

\textbf{Model}

& \rotatebox{60}{SSIM$\uparrow$}
& \rotatebox{60}{LPIPS$\downarrow$}
& \rotatebox{60}{ViSim$\uparrow$}
& \rotatebox{60}{Gemini$\uparrow$}
& \rotatebox{60}{GPT$\uparrow$}
& \rotatebox{60}{Clean$\uparrow$}
& \rotatebox{60}{Render$\uparrow$}

& \rotatebox{60}{SSIM$\uparrow$}
& \rotatebox{60}{LPIPS$\downarrow$}
& \rotatebox{60}{ViSim$\uparrow$}
& \rotatebox{60}{Gemini$\uparrow$}
& \rotatebox{60}{GPT$\uparrow$}
& \rotatebox{60}{Clean$\uparrow$}
& \rotatebox{60}{Render$\uparrow$}

& \rotatebox{60}{SSIM$\uparrow$}
& \rotatebox{60}{LPIPS$\downarrow$}
& \rotatebox{60}{ViSim$\uparrow$}
& \rotatebox{60}{Gemini$\uparrow$}
& \rotatebox{60}{GPT$\uparrow$}
& \rotatebox{60}{Clean$\uparrow$}
& \rotatebox{60}{Render$\uparrow$}

\\
\midrule

\rowcolor{gray!20}
\multicolumn{22}{c}{\textit{Classical raster-to-vector}} \\

\graytxt VTracer
& \graytxt 95.0 & \graytxt 9.2 & \graytxt 93.8 & \graytxt 81.0 & \graytxt 86.6 & \graytxt 0.0 & \graytxt 99.7
& \graytxt 94.2 & \graytxt 11.3 & \graytxt 88.6 & \graytxt 71.3 & \graytxt 80.1 & \graytxt 0.0 & \graytxt 100.0
& \graytxt 88.5 & \graytxt 13.0 & \graytxt 90.3 & \graytxt 78.7 & \graytxt 82.4 & \graytxt 0.0 & \graytxt 100.0 \\

\midrule
\rowcolor{gray!20}
\multicolumn{22}{c}{\textit{Closed-source VLMs}} \\

\graytxt Claude-Sonnet-4.6
& \graytxt 71.0 & \graytxt 47.1 & \graytxt 93.2 & \graytxt 76.7 & \graytxt 87.4 & \graytxt 85.6 & \graytxt 96.2
& \graytxt 73.2 & \graytxt 35.9 & \graytxt 93.5 & \graytxt 80.7 & \graytxt 88.9 & \graytxt 84.8 & \graytxt 98.8
& \graytxt 59.3 & \graytxt 42.0 & \graytxt 91.2 & \graytxt 70.2 & \graytxt 78.0 & \graytxt 72.8 & \graytxt 99.3 \\

\graytxt GPT-5.2
& \graytxt 72.7 & \graytxt 36.4 & \graytxt 95.7 & \graytxt 80.8 & \graytxt 90.8 & \graytxt 73.1 & \graytxt 99.5
& \graytxt 76.3 & \graytxt 28.3 & \graytxt 95.5 & \graytxt 85.4 & \graytxt 93.3 & \graytxt 79.2 & \graytxt 100.0
& \graytxt 60.6 & \graytxt 34.9 & \graytxt 93.6 & \graytxt 72.8 & \graytxt 83.4 & \graytxt 68.8 & \graytxt 98.4 \\

\graytxt Gemini-3-Flash
& \graytxt 77.2 & \graytxt 25.8 & \graytxt 96.4 & \graytxt 88.9 & \graytxt 93.7 & \graytxt 78.8 & \graytxt 99.0
& \graytxt 82.8 & \graytxt 16.2 & \graytxt 96.5 & \graytxt 91.8 & \graytxt 95.5 & \graytxt 83.3 & \graytxt 99.2
& \graytxt 67.2 & \graytxt 24.5 & \graytxt 95.0 & \graytxt 87.9 & \graytxt 90.7 & \graytxt 68.0 & \graytxt 99.1 \\

\graytxt Gemini-3-Pro
& \graytxt 75.6 & \graytxt 30.3 & \graytxt 96.4 & \graytxt 91.5 & \graytxt 94.8 & \graytxt 78.7 & \graytxt 90.2
& \graytxt 78.4 & \graytxt 24.4 & \graytxt 95.9 & \graytxt 90.6 & \graytxt 95.1 & \graytxt 81.4 & \graytxt 93.0
& \graytxt 63.7 & \graytxt 31.1 & \graytxt 94.3 & \graytxt 87.1 & \graytxt 90.2 & \graytxt 66.9 & \graytxt 94.5 \\

\midrule
\rowcolor{gray!20}
\multicolumn{22}{c}{\textit{Open-source VLMs}} \\

InternVL3.5-4B
& 61.4 & 69.8 & 67.5 & 14.3 & 16.7 & 48.6 & 34.3
& 65.7 & 63.3 & 65.9 & 16.8 & 19.7 & 12.2 & 15.2
& 48.9 & 63.0 & 70.5 & 20.4 & 20.9 & 24.8 & 46.4 \\

Qwen3-VL-4B
& 70.8 & 57.4 & 85.7 & 40.3 & 52.8 & 79.4 & 47.6
& 72.2 & 51.2 & 85.9 & 48.0 & 60.0 & 77.4 & 62.9
& 61.4 & 55.4 & 80.5 & 43.1 & 46.8 & 59.1 & 49.5 \\

OmniSVG-4B
& 69.5 & 60.1 & 50.5 & 3.8 & 4.0 & 0.0 & 81.9
& 70.5 & 54.5 & 50.4 & 9.3 & 9.8 & 0.0 & 89.4
& 58.6 & 57.3 & 56.9 & 8.9 & 8.9 & 0.0 & 87.5 \\

InternSVG-8B
& 45.5 & 80.6 & 41.4 & 0.0 & 0.0 & 7.6 & 37.0
& 38.8 & 81.2 & 36.7 & 0.1 & 0.0 & 4.6 & 43.5
& 34.7 & 78.7 & 46.7 & 0.9 & 0.3 & 2.1 & 51.1 \\

Starvector-8B
& 69.9 & 38.0 & 85.1 & 50.6 & 59.0 & 54.4 & 9.3
& 75.5 & 31.0 & 84.5 & 61.2 & 68.8 & 73.1 & 13.4
& {\textbf{67.7}} & 29.9 & 90.5 & 66.6 & 73.5 & 42.6 & 54.3 \\

\rowcolor{gray!20}
\multicolumn{22}{c}{\textit{Ours}} \\

\model-4B (SFT)
& 76.3 & 26.4 & 95.1 & 71.5 & 84.7 & 78.4 & 88.4
& 78.3 & 22.6 & 93.7 & 71.7 & 83.4 & 82.8 & 96.6
& 63.3 & 31.1 & 90.7 & 60.4 & 70.2 & 71.0 & 93.9 \\

\model-4B (SFT+RL)
& {\textbf{77.6}} & {\textbf{21.6}} & \underline{95.6} & \underline{78.2} & {\underline{87.5}} & {\textbf{85.6}} & {\underline{95.2}}
& \textbf{80.6} & \textbf{16.3} & \textbf{95.3} & {\underline{81.0}} & {\textbf{89.0}} & {\textbf{86.5}} & \textbf{99.6}
& 65.4 & \underline{25.3} & \underline{92.1} & \underline{67.7} & {\textbf{77.8}} & {\textbf{80.5}} & {\textbf{98.0}} \\

\model-8B (SFT)
& 76.3 & 26.0 & 95.4 & 74.4 & 86.7 & 75.8 & 90.7
& 78.3 & 22.0 & 94.1 & 73.4 & 84.5 & 81.6 & 96.4
& 63.3 & 30.1 & 90.1 & 60.4 & 71.0 & 69.6 & 95.9 \\

\model-8B (SFT+RL)
& \underline{77.4} & \underline{22.3} & {\textbf{96.0}} & \textbf{80.0} & \textbf{89.0} & \underline{81.9} & {\textbf{96.0}}
& {\underline{80.3}} & {\underline{17.3}} & {\underline{95.2}} & \textbf{81.4} & \underline{88.8} & \underline{84.2} & {\underline{99.4}}
& \underline{66.0} & {\textbf{25.1}} & {\textbf{92.4}} & \textbf{68.2} & \underline{77.5} & \underline{74.9} & \underline{97.0} \\

\end{tabular}
}
\vspace{-3mm}
\end{table*}

\noindent\textbf{Baselines.}
We compare against proprietary VLMs (Gemini-3-Flash, Gemini-3-Pro~\cite{google_gemini3_2025}, GPT-5.2~\cite{openai_gpt5_2_2025}, Claude-Sonnet-4.6~\cite{anthropic2026claudesonnet46}), open-source general VLMs (~\cite{bai2025qwen3vltechnicalreport,zhu2025internvl3exploringadvancedtraining}) and SVG-specialized VLMs (OmniSVG~\cite{omnisvg}, StarVector~\cite{starvector}, InternSVG~\cite{internsvg}), and the classical raster-to-vector tool VTracer~\cite{vtracer}.

\noindent\textbf{Benchmarks.}
We evaluate on three diagram-centric benchmarks: \bench (392 figures held out from \compds), Molmo2-Diagrams (500 samples held out from Molmo2-Diagram~\cite{Molmo2}), and SVG-Diagrams (474 examples from the StarVector~\cite{starvector} test split). We also introduce \benchood{}, a 198-figure out-of-distribution benchmark curated from the top-50 most-cited arXiv papers across 12 CS subfields (\eg, LLMs, computer vision, robotics, and RL/planning) published in 2023--2024, queried through Semantic Scholar to ensure disjointness from the training data.

\noindent\textbf{Metrics.}
We compare rasterized predictions with ground-truth images along four automatic metric families: (i) \emph{pixel similarity} via SSIM~\cite{ssim} and LPIPS~\cite{zhang2018unreasonable}; (ii) \emph{visual similarity} via VisualSim, the average cosine similarity of DINO~\cite{dino}, CLIP~\cite{clip}, and SigLIP~\cite{siglip} embeddings; (iii) \emph{rubric-based VLM-Judge}, averaged over four axes (presence, layout, connectivity, details) using Gemini-3-Flash and GPT-5.2 as a second judge to mitigate self-preference bias; and (iv) \emph{code quality} via svg cleanliness and render rate. For \progds{} we additionally report a rule-based score over shape and arrow correctness~\ref{app: rule based eval}. On \benchood{} we further report Elo ratings from a standard pairwise human preference evaluation following Chatbot Arena~\cite{chiang2024chatbot}: for each trial, an annotator views the ground-truth figure alongside two anonymized SVG reconstructions (Fig.~\ref{fig:ood_results}). Full human evaluation details are in Appendix~\ref{appendix:human_evaluations}.

\subsection{Results}
\begin{figure*}[!ht]
    \centering
    \includegraphics[width=\linewidth]{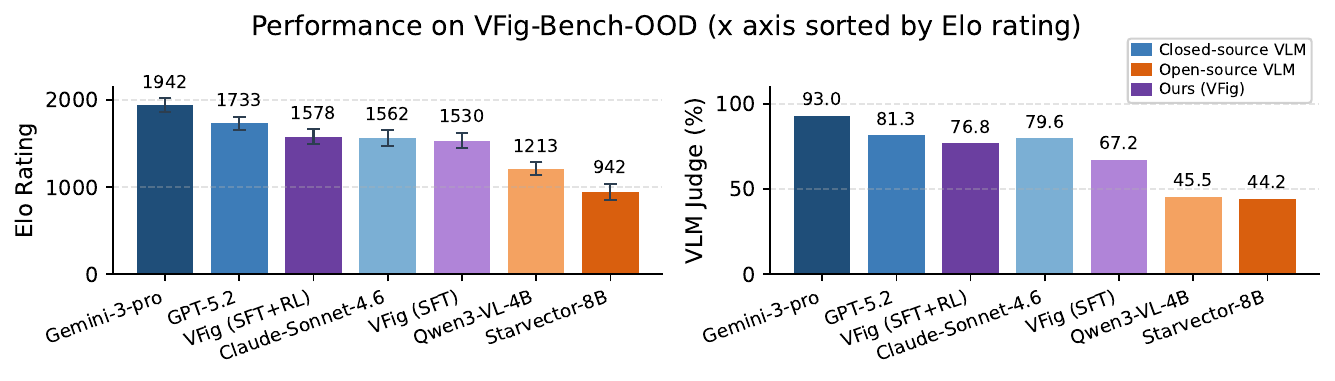}
    \vspace{-5mm}
    \caption{\textbf{Out-of-distribution generalization on \benchood{}} (198 manually curated figures from highly cited arXiv papers). Automatic metrics are shown alongside bootstrap-averaged human Elo from pairwise evaluation.}
    \label{fig:ood_results}
    \vspace{-3mm}
\end{figure*}

We report main results across our four diagram-centric benchmarks (Tab.~\ref{tab:main_benchmark}, Fig.~\ref{fig:ood_results}), ablations on the SFT curriculum (Tab.~\ref{tab:sft_ablation}) and RL reward design (Tab.~\ref{tab:rl_ablation}). Our takeaways include: \model{} (SFT+RL) achieves the best open-source scores on every metric except SVG-Diagram SSIM, narrowing the gap to proprietary models; Simple-to-complex curriculum SFT substantially raises render rate and both VLM-judge scores over one-stage SFT; RL with rendering-aware rewards adds further gains across all metrics, and rubric-based VLM rewards outperform other lower-level reward objectives.

\noindent\textbf{Main results.}
Table~\ref{tab:main_benchmark} reports performance across the three benchmarks. Our SFT+RL models obtain the best scores among open-source baselines on every metric except SVG-Diagram SSIM, where Starvector-8B leads. On \bench{}, \model-4B (SFT+RL) reaches Gemini-Judge $78.2$ / GPT-Judge $87.5$, trailing GPT-5.2 ($80.8$ / $90.8$) and Gemini-3-Pro ($91.5$ / $94.8$); scaling to \model-8B (SFT+RL) further narrows the gap ($80.0$ / $89.0$). Adding RL on top of SFT yields consistent gains across all metrics on all three benchmarks; qualitative comparisons are shown in Fig.~\ref{fig:visual_examples}.

\noindent\textbf{Generalization to OOD diagrams.}
We evaluate on \benchood{} with automatic metrics and bootstrap-averaged human Elo (Fig.~\ref{fig:ood_results}). On VLM-Judge (averaged over Gemini and GPT), \model{} (SFT+RL) reaches $76.8$, exceeding the best open-source baseline by over $30$ points and approaching Claude-Sonnet-4.6 and GPT-5.2, while remaining behind Gemini-3-Pro. Human Elo ratings rank the models Gemini-3-Pro $>$ GPT-5.2 $>$ \model{} (SFT+RL) $>$ Claude-Sonnet-4.6 $>$ \model{} (SFT) $>$ Qwen3-VL-4B $>$ Starvector-8B. 
The Elo and VLM-Judge rankings agree on six of seven positions; the only swap is between \model{} (SFT+RL) and Claude-Sonnet-4.6, where humans rank \model{} (SFT+RL) above Claude while the VLM judge orders them in the opposite direction.

\begin{table*}[!h]
\vspace{-3mm}
\centering
\caption{\textbf{SFT ablations.} Metrics averaged across \bench{} + Molmo2-Diagram + SVG-Diagram. Green deltas show the gain over the baselines.}
\begin{minipage}[t]{0.46\textwidth}
\centering
\textbf{(a) Backbone ablation}
\label{tab:sft_backbone_ablation}
\scriptsize
\setlength{\tabcolsep}{2pt}
\renewcommand{\arraystretch}{1.0}
\vspace{1mm}
\begin{tabular}{lllll}
\toprule
\textbf{Backbone} &
\multicolumn{1}{l}{\textbf{SSIM}$\uparrow$} &
\multicolumn{1}{l}{\textbf{Gemini}$\uparrow$} &
\multicolumn{1}{l}{\textbf{GPT}$\uparrow$} &
\multicolumn{1}{l}{\textbf{Render}$\uparrow$} \\

\midrule
InternVL3.5-4B        & 58.7 & 17.2 & 19.1 & 32.0 \\
\quad + SFT          & \underline{68.1}{\color{green!60!black}$^{+9.4}$} &
\underline{52.4}{\color{green!60!black}$^{+35.2}$} & \underline{67.0}{\color{green!60!black}$^{+47.9}$} & \underline{79.0}{\color{green!60!black}$^{+47.0}$} \\

\midrule
Qwen3-VL-4B           & \underline{68.1} & 43.8 & 53.2 & 53.3 \\
\quad + SFT          & \textbf{72.7}{\color{green!60!black}$^{+4.6}$} & \textbf{67.9}{\color{green!60!black}$^{+24.1}$} & \textbf{79.4}{\color{green!60!black}$^{+26.2}$} & \textbf{93.3}{\color{green!60!black}$^{+40.0}$} \\

\bottomrule
\end{tabular}
\end{minipage}%
\hfill\begin{minipage}[t]{0.52\textwidth}
\centering
\textbf{(b) Curriculum ablation}
\label{tab:sft_ablation}
\scriptsize
\setlength{\tabcolsep}{2pt}
\renewcommand{\arraystretch}{1.0}
\vspace{1mm}
\begin{tabular}{lllll}
\toprule
\textbf{Curriculum} &
\multicolumn{1}{l}{\textbf{SSIM}$\uparrow$} &
\multicolumn{1}{l}{\textbf{Gemini}$\uparrow$} &
\multicolumn{1}{l}{\textbf{GPT}$\uparrow$} &
\multicolumn{1}{l}{\textbf{Render}$\uparrow$} \\
\midrule
one-stage SFT          & \textbf{75.0} & 65.0 & 77.4 & 74.9 \\
simp-to-comp SFT & 72.7{\color{red}$^{-2.3}$} & 67.9{\color{green!60!black}$^{+2.9}$} & 79.4{\color{green!60!black}$^{+2.0}$} & \underline{93.3}{\color{green!60!black}$^{+18.4}$} \\
\midrule

one-stage + RL          & 74.0 & \underline{71.6} & \underline{81.8} & \textbf{97.0} \\
simp-to-comp + RL  & \underline{74.4}{\color{green!60!black}$^{+0.4}$} & \textbf{74.1}{\color{green!60!black}$^{+2.5}$} & \textbf{83.7}{\color{green!60!black}$^{+1.9}$} & \textbf{97.0}{\color{green!60!black}$^{+0.0}$} \\


\bottomrule
\end{tabular}
\end{minipage}%

\vspace{-3mm}
\end{table*}

\noindent\textbf{SFT Ablations.}
Table~\ref{tab:sft_backbone_ablation} compares two backbones, Qwen3-VL-4B and InternVL3.5-4B, before and after our simple-to-complex SFT. SFT delivers large gains on both models. Render improves by $+40.0$ and $+47.0$ points, respectively. For Qwen3-VL-4B, VLM-judge scores increase by $+24.1$ on Gemini and $+26.2$ on GPT. InternVL3.5-4B improves even more, with gains of $+35.2$ on Gemini and $+47.9$ on GPT. These results confirm that the improvements come from our training data rather than a specific backbone. Table~\ref{tab:sft_ablation} then compares one-stage SFT against the simple-to-complex curriculum on Qwen3-VL-4B, both before and after RL. With SFT alone, the curriculum raises Render ($+18.4$) and both VLM-judge scores (Gemini $+2.9$, GPT $+2.0$), indicating that learning primitive-level generation and simple layouts first stabilizes training before adapting to complex diagram-centric figures. After RL, Render saturates for both variants ($97.0$), but the curriculum still wins on SSIM ($+0.4$) and both VLM-judge scores (Gemini $+2.5$, GPT $+1.9$). Per-backbone breakdowns including Qwen2.5-VL and Qwen3-VL-8B, and the backbone-scale ablation, are deferred to Appendix~\ref{appendix: ablation}.

\begin{wraptable}[18]{r}{0.5\textwidth}
\vspace{-3mm}
\centering
\vspace{-2mm}
\caption{\textbf{RL ablation.} Metrics averaged across \bench{} + Molmo2-Diagram + SVG-Diagram. We ablate on (i) rubric components and (ii) the reward signal. A training-data ablation is in Appendix~\ref{appendix: ablation}.}
\label{tab:rl_ablation}
\footnotesize
\setlength{\tabcolsep}{3pt}
\renewcommand{\arraystretch}{1.0}
\begin{tabular}{lrrrr}
\toprule
\textbf{Variant}
& \textbf{SSIM}$\uparrow$
& \textbf{Gemini}$\uparrow$
& \textbf{GPT}$\uparrow$
& \textbf{Render}$\uparrow$ \\
\midrule
\rowcolor{gray!20}
\multicolumn{5}{l}{\textit{Rubric components in Gemini reward model}} \\
\textbf{Full (P+L+C+D)}
& \underline{74.7} & {\textbf{75.9}} & {\textbf{85.0}} & {\textbf{97.8}} \\
$-$~Presence      & 74.4 & 74.6 & 83.9 & 95.1 \\
$-$~Connectivity  & 74.5 & \underline{75.0} & \underline{84.2} & 96.8 \\
$-$~Layout        & 74.2 & 74.0 & 83.4 & 95.7 \\
$-$~Details       & 74.4 & 74.4 & 83.4 & 96.1 \\
No Rubric         & 74.3 & 72.7 & 82.6 & 97.2 \\
\midrule
\rowcolor{gray!20}
\multicolumn{5}{l}{\textit{Alternative reward signals}} \\
Gemini Full + Pixel    & {\textbf{75.2}} & 73.1 & 82.8 & 95.7 \\
DINO Reward       & 74.0 & 72.1 & 81.9 & 96.4 \\
GPT Full Reward        & 74.4 & 73.9 & 83.1 & \underline{97.6} \\
\bottomrule
\end{tabular}
\end{wraptable}
\noindent\textbf{RL Ablations.}
Table~\ref{tab:rl_ablation} ablates two design dimensions: (i) the rubric decomposition and (ii) the reward signal source. Our full reward decomposes into four structural axes---\emph{presence}, \emph{layout}, \emph{connectivity}, and \emph{details}. Removing any one axis degrades judge-based metrics, with the largest drops when removing \emph{layout} or \emph{details}, highlighting the importance of spatial arrangement and fine-grained structural fidelity. Replacing the four-axis rubric with a single Gemini score (\emph{No Rubric}) further drops VLM judges by $2-3$ points, showing that explicit rubric decomposition is itself a useful inductive bias for RL feedback. Augmenting the rubric reward with pixel-level objectives (Gemini+Pixel, fully defined in Appendix~\ref{app:training_objectives}) slightly improves SSIM but reduces judge-based scores, suggesting a tradeoff between pixel reconstruction and structural optimization. DINO embedding similarity hurts judge-based metrics, indicating that embedding-level visual matching provides weaker structural supervision than a rubric-based VLM judge. Comparing rubric judges, the Gemini-rubric reward yields the best score under \emph{both} Gemini and GPT judges, while the GPT-rubric reward is worse under both---ruling out any judge-favoritism bias and showing that Gemini-rubric supervision is stronger across the board.

\begin{figure*}[t]
\vspace{-3mm}
    \centering
    \includegraphics[width=\textwidth]{sections/figure/visual_examples_across_models1_new_new_new_new.pdf}
    \caption{
    \textbf{Qualitative comparison across models.}
    P/L/C/D denote the Gemini judge scores for \textit{presence}, \textit{layout}, \textit{connectivity}, and \textit{details}.
    }
\label{fig:visual_examples}
\vspace{-5mm}
\end{figure*}

\section{Conclusion and Limitations}
\vspace{-5mm}
\label{sec:conclusion}

 We introduce \dataset{} and \bench{} to advance data and evaluation for complex diagram-centric figure-to-SVG generation. \dataset{} provides a large-scale 66K figure–SVG dataset that combines real-world diagrams with programmatically generated ones, while \bench{} offers a structure-aware evaluation suite that measures layout, connectivity, editability, and rendering quality beyond pixel similarity. Our data and benchmark focus on diagram-centric figures rather than natural images, dense plots, tables, math-heavy pages, or broader graphic-design domains such as websites. Some limitations remain: current models still struggle with fine-grained local structures such as connectors and arrowheads, and because \dataset{}'s real-world targets are Gemini-3-Pro renders, the data inherits a stylistic prior. We mitigate this through programmatic supervision in \progds{}, a cross-family GPT-5.2 judge, and human Elo on \benchood{}, though additional judge families and human-authored ground truth would further close this gap. We have publicly released the dataset, benchmark, models, and code to support future research on editable diagram-centric figure generation.



\bibliographystyle{plainnat}
\bibliography{main}

@String(CVPR  = {IEEE Conf. Comput. Vis. Pattern Recog.})

@String(ICCV  = {Int. Conf. Comput. Vis.})

@String(ECCV  = {Eur. Conf. Comput. Vis.})

@String(NeurIPS = {Adv. Neural Inform. Process. Syst.})

@String(TOG   = {ACM Trans. Graph.})

@String(TIP   = {IEEE Trans. Image Process.})

@String(CVPR  = {CVPR})

@String(ICCV  = {ICCV})

@String(ECCV  = {ECCV})

@String(NeurIPS = {NeurIPS})

@String(TOG   = {ACM TOG})

@String(TIP   = {IEEE TIP})

@article{selinger2003potrace,
  title={Potrace: A polygon-based tracing algorithm},
  author={Selinger, Peter},
  year={2003},
  journal={Unpublished manuscript}
}

@misc{vtracer,
  author = {{Vision Cortex}},
  title = {VTracer},
  year = {2023},
  url = {https://www.visioncortex.org/vtracer-docs}
}

@misc{deepsvg,
      title={DeepSVG: A Hierarchical Generative Network for Vector Graphics Animation}, 
      author={Alexandre Carlier and Martin Danelljan and Alexandre Alahi and Radu Timofte},
      year={2020},
      eprint={2007.11301},
      archivePrefix={arXiv},
      primaryClass={cs.CV},
      url={https://arxiv.org/abs/2007.11301}, 
}

@inproceedings{diffvg,
  title={Differentiable Vector Graphics Rasterization for Editing and Learning},
  author={Li, Tzu-Mao and Luk{\'a}{\v{c}}, Michal and Gharbi, Micha{\"e}l and Ragan-Kelley, Jonathan},
  booktitle={SIGGRAPH},
  year={2020}
}

@misc{im2vec,
      title={Im2Vec: Synthesizing Vector Graphics without Vector Supervision}, 
      author={Pradyumna Reddy and Michael Gharbi and Michal Lukac and Niloy J. Mitra},
      year={2021},
      eprint={2102.02798},
      archivePrefix={arXiv},
      primaryClass={cs.CV},
      url={https://arxiv.org/abs/2102.02798}, 
}

@misc{starvector,
      title={StarVector: Generating Scalable Vector Graphics Code from Images and Text}, 
      author={Juan A. Rodriguez and Abhay Puri and Shubham Agarwal and Issam H. Laradji and Pau Rodriguez and Sai Rajeswar and David Vazquez and Christopher Pal and Marco Pedersoli},
      year={2025},
      eprint={2312.11556},
      archivePrefix={arXiv},
      primaryClass={cs.CV},
      url={https://arxiv.org/abs/2312.11556}, 
}

@misc{llm4svg,
      title={Empowering LLMs to Understand and Generate Complex Vector Graphics}, 
      author={Ximing Xing and Juncheng Hu and Guotao Liang and Jing Zhang and Dong Xu and Qian Yu},
      year={2025},
      eprint={2412.11102},
      archivePrefix={arXiv},
      primaryClass={cs.CV},
      url={https://arxiv.org/abs/2412.11102}, 
}

@inproceedings{zhang2018unreasonable,
  title={The unreasonable effectiveness of deep features as a perceptual metric},
  author={Zhang, Richard and Isola, Phillip and Efros, Alexei A and Shechtman, Eli and Wang, Oliver},
  booktitle={CVPR},
  year={2018}
}

@misc{omnisvg,
      title={OmniSVG: A Unified Scalable Vector Graphics Generation Model}, 
      author={Yiying Yang and Wei Cheng and Sijin Chen and Xianfang Zeng and Fukun Yin and Jiaxu Zhang and Liao Wang and Gang Yu and Xingjun Ma and Yu-Gang Jiang},
      year={2025},
      eprint={2504.06263},
      archivePrefix={arXiv},
      primaryClass={cs.CV},
      url={https://arxiv.org/abs/2504.06263}, 
}

@article{ocr-vqgan,
  title={Ocr-vqgan: Taming text-within-image generation},
  author={Rodriguez, Juan A and Vazquez, David and Laradji, Issam and Pedersoli, Marco and Rodriguez, Pau},
  journal={WACV},
  year={2023}
}

@misc{bai2025qwen3vltechnicalreport,
      title={Qwen3-VL Technical Report}, 
      author={Shuai Bai and Yuxuan Cai and Ruizhe Chen and Keqin Chen and Xionghui Chen and Zesen Cheng and Lianghao Deng and Wei Ding and Chang Gao and Chunjiang Ge and Wenbin Ge and Zhifang Guo and Qidong Huang and Jie Huang and Fei Huang and Binyuan Hui and Shutong Jiang and Zhaohai Li and Mingsheng Li and Mei Li and Kaixin Li and Zicheng Lin and Junyang Lin and Xuejing Liu and Jiawei Liu and Chenglong Liu and Yang Liu and Dayiheng Liu and Shixuan Liu and Dunjie Lu and Ruilin Luo and Chenxu Lv and Rui Men and Lingchen Meng and Xuancheng Ren and Xingzhang Ren and Sibo Song and Yuchong Sun and Jun Tang and Jianhong Tu and Jianqiang Wan and Peng Wang and Pengfei Wang and Qiuyue Wang and Yuxuan Wang and Tianbao Xie and Yiheng Xu and Haiyang Xu and Jin Xu and Zhibo Yang and Mingkun Yang and Jianxin Yang and An Yang and Bowen Yu and Fei Zhang and Hang Zhang and Xi Zhang and Bo Zheng and Humen Zhong and Jingren Zhou and Fan Zhou and Jing Zhou and Yuanzhi Zhu and Ke Zhu},
      year={2025},
      eprint={2511.21631},
      archivePrefix={arXiv},
      primaryClass={cs.CV},
      url={https://arxiv.org/abs/2511.21631}, 
}

@misc{bai2025qwen25vltechnicalreport,
      title={Qwen2.5-VL Technical Report}, 
      author={Shuai Bai and Keqin Chen and Xuejing Liu and Jialin Wang and Wenbin Ge and Sibo Song and Kai Dang and Peng Wang and Shijie Wang and Jun Tang and Humen Zhong and Yuanzhi Zhu and Mingkun Yang and Zhaohai Li and Jianqiang Wan and Pengfei Wang and Wei Ding and Zheren Fu and Yiheng Xu and Jiabo Ye and Xi Zhang and Tianbao Xie and Zesen Cheng and Hang Zhang and Zhibo Yang and Haiyang Xu and Junyang Lin},
      year={2025},
      eprint={2502.13923},
      archivePrefix={arXiv},
      primaryClass={cs.CV},
      url={https://arxiv.org/abs/2502.13923}, 
}

@misc{zhu2025internvl3exploringadvancedtraining,
      title={InternVL3: Exploring Advanced Training and Test-Time Recipes for Open-Source Multimodal Models}, 
      author={Jinguo Zhu and Weiyun Wang and Zhe Chen and Zhaoyang Liu and Shenglong Ye and Lixin Gu and Hao Tian and Yuchen Duan and Weijie Su and Jie Shao and Zhangwei Gao and Erfei Cui and Xuehui Wang and Yue Cao and Yangzhou Liu and Xingguang Wei and Hongjie Zhang and Haomin Wang and Weiye Xu and Hao Li and Jiahao Wang and Nianchen Deng and Songze Li and Yinan He and Tan Jiang and Jiapeng Luo and Yi Wang and Conghui He and Botian Shi and Xingcheng Zhang and Wenqi Shao and Junjun He and Yingtong Xiong and Wenwen Qu and Peng Sun and Penglong Jiao and Han Lv and Lijun Wu and Kaipeng Zhang and Huipeng Deng and Jiaye Ge and Kai Chen and Limin Wang and Min Dou and Lewei Lu and Xizhou Zhu and Tong Lu and Dahua Lin and Yu Qiao and Jifeng Dai and Wenhai Wang},
      year={2025},
      eprint={2504.10479},
      archivePrefix={arXiv},
      primaryClass={cs.CV},
      url={https://arxiv.org/abs/2504.10479}, 
}

@misc{hu2021loralowrankadaptationlarge,
      title={LoRA: Low-Rank Adaptation of Large Language Models}, 
      author={Edward J. Hu and Yelong Shen and Phillip Wallis and Zeyuan Allen-Zhu and Yuanzhi Li and Shean Wang and Lu Wang and Weizhu Chen},
      year={2021},
      eprint={2106.09685},
      archivePrefix={arXiv},
      primaryClass={cs.CL},
      url={https://arxiv.org/abs/2106.09685}, 
}

@misc{openai_gpt5_2_2025,
  author       = {{OpenAI}},
  title        = {Introducing GPT-5.2},
  year         = {2025},
  howpublished = {\url{https://openai.com/index/introducing-gpt-5-2/}}
  }

@misc{google_gemini3_2025,
  author       = {{Google DeepMind}},
  title        = {Gemini 3},
  year         = {2025},
  howpublished = {\url{https://deepmind.google/technologies/gemini/}}
  }

@article{ssim,
  title={Image quality assessment: from error visibility to structural similarity},
  author={Wang, Zhou and Bovik, Alan C and Sheikh, Hamid R and Simoncelli, Eero P},
  journal={IEEE TIP},
  year={2004},
}

@misc{dino,
      title={DINOv2: Learning Robust Visual Features without Supervision}, 
      author={Maxime Oquab and Timothée Darcet and Théo Moutakanni and Huy Vo and Marc Szafraniec and Vasil Khalidov and Pierre Fernandez and Daniel Haziza and Francisco Massa and Alaaeldin El-Nouby and Mahmoud Assran and Nicolas Ballas and Wojciech Galuba and Russell Howes and Po-Yao Huang and Shang-Wen Li and Ishan Misra and Michael Rabbat and Vasu Sharma and Gabriel Synnaeve and Hu Xu and Hervé Jegou and Julien Mairal and Patrick Labatut and Armand Joulin and Piotr Bojanowski},
      year={2024},
      eprint={2304.07193},
      archivePrefix={arXiv},
      primaryClass={cs.CV},
      url={https://arxiv.org/abs/2304.07193}, 
}

@misc{siglip,
      title={Sigmoid Loss for Language Image Pre-Training}, 
      author={Xiaohua Zhai and Basil Mustafa and Alexander Kolesnikov and Lucas Beyer},
      year={2023},
      eprint={2303.15343},
      archivePrefix={arXiv},
      primaryClass={cs.CV},
      url={https://arxiv.org/abs/2303.15343}, 
}

@misc{clip,
      title={Learning Transferable Visual Models From Natural Language Supervision}, 
      author={Alec Radford and Jong Wook Kim and Chris Hallacy and Aditya Ramesh and Gabriel Goh and Sandhini Agarwal and Girish Sastry and Amanda Askell and Pamela Mishkin and Jack Clark and Gretchen Krueger and Ilya Sutskever},
      year={2021},
      eprint={2103.00020},
      archivePrefix={arXiv},
      primaryClass={cs.CV},
      url={https://arxiv.org/abs/2103.00020}, 
}

@misc{internsvg,
      title={InternSVG: Towards Unified SVG Tasks with Multimodal Large Language Models}, 
      author={Haomin Wang and Jinhui Yin and Qi Wei and Wenguang Zeng and Lixin Gu and Shenglong Ye and Zhangwei Gao and Yaohui Wang and Yanting Zhang and Yuanqi Li and Yanwen Guo and Wenhai Wang and Kai Chen and Yu Qiao and Hongjie Zhang},
      year={2026},
      eprint={2510.11341},
      archivePrefix={arXiv},
      primaryClass={cs.CV},
      url={https://arxiv.org/abs/2510.11341}, 
}

@misc{reason-svg,
      title={Reason-SVG: Hybrid Reward RL for Aha-Moments in Vector Graphics Generation}, 
      author={Ximing Xing and Yandong Guan and Jing Zhang and Dong Xu and Qian Yu},
      year={2025},
      eprint={2505.24499},
      archivePrefix={arXiv},
      primaryClass={cs.CV},
      url={https://arxiv.org/abs/2505.24499}, 
}

@misc{rlrf,
      title={Rendering-Aware Reinforcement Learning for Vector Graphics Generation}, 
      author={Juan A. Rodriguez and Haotian Zhang and Abhay Puri and Aarash Feizi and Rishav Pramanik and Pascal Wichmann and Arnab Mondal and Mohammad Reza Samsami and Rabiul Awal and Perouz Taslakian and Spandana Gella and Sai Rajeswar and David Vazquez and Christopher Pal and Marco Pedersoli},
      year={2025},
      eprint={2505.20793},
      archivePrefix={arXiv},
      primaryClass={cs.CV},
      url={https://arxiv.org/abs/2505.20793}, 
}

@misc{Molmo2,
      title={Molmo2: Open Weights and Data for Vision-Language Models with Video Understanding and Grounding}, 
      author={Christopher Clark and Jieyu Zhang and Zixian Ma and Jae Sung Park and Mohammadreza Salehi and Rohun Tripathi and Sangho Lee and Zhongzheng Ren and Chris Dongjoo Kim and Yinuo Yang and Vincent Shao and Yue Yang and Weikai Huang and Ziqi Gao and Taira Anderson and Jianrui Zhang and Jitesh Jain and George Stoica and Winson Han and Ali Farhadi and Ranjay Krishna},
      year={2026},
      eprint={2601.10611},
      archivePrefix={arXiv},
      primaryClass={cs.CV},
      url={https://arxiv.org/abs/2601.10611}, 
}

@misc{deepseekmath,
      title={DeepSeekMath: Pushing the Limits of Mathematical Reasoning in Open Language Models}, 
      author={Zhihong Shao and Peiyi Wang and Qihao Zhu and Runxin Xu and Junxiao Song and Xiao Bi and Haowei Zhang and Mingchuan Zhang and Y. K. Li and Y. Wu and Daya Guo},
      year={2024},
      eprint={2402.03300},
      archivePrefix={arXiv},
      primaryClass={cs.CL},
      url={https://arxiv.org/abs/2402.03300}, 
}

@inproceedings{he2016resnet,
  title={Identity mappings in deep residual networks},
  author={He, Kaiming and Zhang, Xiangyu and Ren, Shaoqing and Sun, Jian},
  booktitle={ECCV},
  year={2016},
}

@article{vaswani2017attention,
  title={Attention is all you need},
  author={Vaswani, Ashish and Shazeer, Noam and Parmar, Niki and Uszkoreit, Jakob and Jones, Llion and Gomez, Aidan N and Kaiser, {\L}ukasz and Polosukhin, Illia},
  journal={NeurIPS},
  year={2017}
}

@phdthesis{diebel2008bayesian,
  title        = {Bayesian Image Vectorization: the probabilistic inversion of vector image rasterization},
  author       = {Diebel, James Richard},
  school       = {Stanford University},
  year         = {2008}
}

@article{xia2009patch,
  title        = {Patch-Based Image Vectorization with Automatic Curvilinear Feature Alignment},
  author       = {Xia, Tian and Liao, Binbin and Yu, Yizhou},
  journal      = {ACM TOG},
  year         = {2009},
}

@article{liao2012subdivision,
  title        = {A Subdivision-Based Representation for Vector Image Editing},
  author       = {Liao, Zicheng and Hoppe, Hugues and Forsyth, David and Yu, Yizhou},
  journal      = {IEEE Transactions on Visualization and Computer Graphics},
  year         = {2012},
}

@InProceedings{Lopes_2019_ICCV,
  author    = {Lopes, Raphael Gontijo and Ha, David and Eck, Douglas and Shlens, Jonathon},
  title     = {A Learned Representation for Scalable Vector Graphics},
  booktitle = {ICCV},
  year      = {2019},
}

@misc{ma_2022_cvpr,
      title={Towards Layer-wise Image Vectorization}, 
      author={Xu Ma and Yuqian Zhou and Xingqian Xu and Bin Sun and Valerii Filev and Nikita Orlov and Yun Fu and Humphrey Shi},
      year={2022},
      eprint={2206.04655},
      archivePrefix={arXiv},
      primaryClass={cs.CV},
      url={https://arxiv.org/abs/2206.04655}, 
}

@InProceedings{Jain_2023_CVPR,
  author    = {Jain, Ajay and Xie, Amber and Abbeel, Pieter},
  title     = {VectorFusion: Text-to-SVG by Abstracting Pixel-Based Diffusion Models},
  booktitle = {CVPR},
  year      = {2023},
}

@InProceedings{Xing_2024_CVPR,
  author    = {Xing, Ximing and Zhou, Haitao and Wang, Chuang and Zhang, Jing and Xu, Dong and Yu, Qian},
  title     = {SVGDreamer: Text Guided SVG Generation with Diffusion Model},
  booktitle = {CVPR},
  year      = {2024},
}

@InProceedings{Wu_2025_CVPR,
  author    = {Wu, Ronghuan and Su, Wanchao and Liao, Jing},
  title     = {Chat2SVG: Vector Graphics Generation with Large Language Models and Image Diffusion Models},
  booktitle = {CVPR},
  year      = {2025},
}

@article{zhang2023beyondpixels,
  title        = {Beyond Pixels: Exploring Human-Readable SVG Generation for Simple Images with Vision Language Models},
  author       = {Zhang, Tong and Liu, Haoyang and Zhang, Peiyan and Cheng, Yuxuan and Wang, Haohan},
  journal      = {arXiv preprint arXiv:2311.15543},
  year         = {2023}
}

@article{wu2023iconshop,
  title        = {IconShop: Text-Guided Vector Icon Synthesis with Autoregressive Transformers},
  author       = {Wu, Ronghuan and Su, Wanchao and Ma, Kede and Liao, Jing},
  journal      = {ACM TOG},
  year         = {2023},
}

@inproceedings{hu2024supersvg,
  title     = {SuperSVG: Superpixel-based Scalable Vector Graphics Synthesis},
  author    = {Hu, Teng and Yi, Ran and Qian, Baihong and Zhang, Jiangning and Rosin, Paul L and Lai, Yu-Kun},
  booktitle = {CVPR},
  year      = {2024}
}

@misc{wang2021deepvecfont,
      title={DeepVecFont: Synthesizing High-quality Vector Fonts via Dual-modality Learning}, 
      author={Yizhi Wang and Zhouhui Lian},
      year={2021},
      eprint={2110.06688},
      archivePrefix={arXiv},
      primaryClass={cs.CV},
      url={https://arxiv.org/abs/2110.06688}, 
}

@misc{ha2017neuralrepresentationsketchdrawings,
      title={A Neural Representation of Sketch Drawings}, 
      author={David Ha and Douglas Eck},
      year={2017},
      eprint={1704.03477},
      archivePrefix={arXiv},
      primaryClass={cs.NE},
      url={https://arxiv.org/abs/1704.03477}, 
}

@inproceedings{zou2024vgbench,
  title     = {VGBench: A Comprehensive Benchmark of Vector Graphics Understanding and Generation for Large Language Models},
  author    = {Zou, Bocheng and Cai, Mu and Zhang, Jianrui and Lee, Yong Jae},
  booktitle = {EMNLP},
  year      = {2024}
}

@article{clouatre2019figr,
  title   = {FIGR: Few-Shot Image Generation with Reptile},
  author  = {Clou{\^a}tre, Louis and Demers, Marc},
  journal = {arXiv preprint arXiv:1901.02199},
  year    = {2019}
}

@inproceedings{chiang2024chatbot,
  title={Chatbot arena: An open platform for evaluating llms by human preference},
  author={Chiang, Wei-Lin and Zheng, Lianmin and Sheng, Ying and Angelopoulos, Anastasios Nikolas and Li, Tianle and Li, Dacheng and Zhu, Banghua and Zhang, Hao and Jordan, Michael and Gonzalez, Joseph E and others},
  booktitle={Forty-first International Conference on Machine Learning},
  year={2024}
}

@misc{wang2026internsvgunifiedsvgtasks,
      title={InternSVG: Towards Unified SVG Tasks with Multimodal Large Language Models}, 
      author={Haomin Wang and Jinhui Yin and Qi Wei and Wenguang Zeng and Lixin Gu and Shenglong Ye and Zhangwei Gao and Yaohui Wang and Yanting Zhang and Yuanqi Li and Yanwen Guo and Wenhai Wang and Kai Chen and Yu Qiao and Hongjie Zhang},
      year={2026},
      eprint={2510.11341},
      archivePrefix={arXiv},
      primaryClass={cs.CV},
      url={https://arxiv.org/abs/2510.11341}, 
}

@misc{greisinger2026tikzillascalingtexttotikzhighquality,
      title={TikZilla: Scaling Text-to-TikZ with High-Quality Data and Reinforcement Learning}, 
      author={Christian Greisinger and Steffen Eger},
      year={2026},
      eprint={2603.03072},
      archivePrefix={arXiv},
      primaryClass={cs.AI},
      url={https://arxiv.org/abs/2603.03072}, 
}

@misc{hazimeh2025semanticdocumentderenderingsvg,
      title={Semantic Document Derendering: SVG Reconstruction via Vision-Language Modeling}, 
      author={Adam Hazimeh and Ke Wang and Mark Collier and Gilles Baechler and Efi Kokiopoulou and Pascal Frossard},
      year={2025},
      eprint={2511.13478},
      archivePrefix={arXiv},
      primaryClass={cs.CV},
      url={https://arxiv.org/abs/2511.13478}, 
}

@misc{liang2026renderintheloopvectorgraphicsgeneration,
      title={Render-in-the-Loop: Vector Graphics Generation via Visual Self-Feedback}, 
      author={Guotao Liang and Zhangcheng Wang and Juncheng Hu and Haitao Zhou and Ziteng Xue and Jing Zhang and Dong Xu and Qian Yu},
      year={2026},
      eprint={2604.20730},
      archivePrefix={arXiv},
      primaryClass={cs.CV},
      url={https://arxiv.org/abs/2604.20730}, 
}

@misc{zhang2025duetsvgunifiedmultimodalsvg,
      title={DuetSVG: Unified Multimodal SVG Generation with Internal Visual Guidance}, 
      author={Peiying Zhang and Nanxuan Zhao and Matthew Fisher and Yiran Xu and Jing Liao and Difan Liu},
      year={2025},
      eprint={2512.10894},
      archivePrefix={arXiv},
      primaryClass={cs.CV},
      url={https://arxiv.org/abs/2512.10894}, 
}

@inproceedings{
    zeng2026davinci,
    title={DaVinci: Reinforcing Visual-Structural Syntax in {MLLM}s for Generalized Scientific Diagram Parsing},
    author={Xingchen ZENG and Zhewei Su and Hengming Zhang and Juyong Jiang and Jiazhi Xia and Wei Zeng},
    booktitle={The Fourteenth International Conference on Learning Representations},
    year={2026},
    url={https://openreview.net/forum?id=OAXECnLxuk}
}

@misc{anthropic2026claudesonnet46,
  title        = {Introducing Claude Sonnet 4.6},
  author       = {Anthropic},
  year         = {2026},
  howpublished = {\url{https://www.anthropic.com/news/claude-sonnet-4-6}}
}

\newpage
\appendix
{\Large \bf Appendix}\\

This appendix provides additional details and results supporting the main paper. 
\begin{itemize}
    \item We first provide additional details on dataset construction, including image filtering, data composition, structural metrics for characterizing SVG complexity and semantic organization, as well as the rule-based evaluation protocol for programmatically generated SVG diagrams (Sec.~\ref{appendix:data}).
    \item We then describe the experimental setup, including training and inference details (Sec.~\ref{appendix:setup}).
    \item We further present additional ablation studies for both supervised fine-tuning and reinforcement learning to analyze the effects of model design choices (Sec.~\ref{appendix: ablation}).
    \item We also present a human evaluation of figure-to-SVG generation quality to complement the automatic benchmark results in the main paper (Sec.~\ref{appendix:human_evaluations}).
    \item We include additional qualitative results across multiple benchmarks (Sec.~\ref{appendix:results}).
    \item We discuss representative failure cases and summarize the current limitations of our approach (Sec.~\ref{appendix: limitation}).
    \item Finally, we review earlier non-LLM SVG generation approaches, including tracing-based vectorization, differentiable rendering, and sequential decoder models (Sec.~\ref{appendix:extended_related_work}).
\end{itemize}

\section{\dataset}
\label{appendix:data}
This section provides additional details on the construction of the \model{} dataset, including: 
(i) figure collection from arXiv documents and programmatically generated diagrams; 
(ii) the image filtering pipeline applied to curate high-quality samples;  
(iii) the definition of the SVG Structural Complexity Metrics used to characterize structural properties of the dataset; and
(iv) the rule-based evaluation for \progds dataset.
\subsection{Arxiv Data} 

To complement our synthetic and model-generated sources with realistic scientific figures, we curate an \emph{in-the-wild} corpus from two pipelines: (i) figures adapted from the Paper2Fig dataset \cite{ocr-vqgan}, and (ii) a large-scale crawl of recent arXiv papers.

For the large-scale acquisition of recent research, we developed an automated pipeline to curate scientific figures from \text{arXiv} papers published after January~2025. By leveraging the \text{arXiv} API, we systematically queried those papers and processed a total of 259{,}073 figure candidates. To ensure the collected diagrams align with the target distribution of our benchmark, we restricted the crawl to specific computer science and systems-oriented disciplines. 

For each selected paper, we extracted referenced figure assets from the \LaTeX{} source by identifying \texttt{\textbackslash includegraphics} occurrences. We prioritized high-quality formats (PDF, PNG, JPG/JPEG) and discarded unsupported or missing references. For figures provided as embedded PDFs, we rasterized them using \texttt{PyMuPDF} to ensure a unified image representation. Following the image-level filtering procedure described in Sec.~\ref{sec:data_filtering}, we retained \textbf{52k} high-quality figures. Overall, plot-style figures dominate the raw crawl, accounting for 153{,}579 out of 259{,}073 candidates, or approximately 59.3\%. Among the non-plot candidates, 52{,}698 out of 105{,}494 were retained as high-quality target diagrams, corresponding to approximately 50.0\% of the non-plot pool.

\subsection{\compds Generation}
\label{sec:compds-generation}
\begin{figure}[h]
    \centering
    \includegraphics[width=\linewidth]{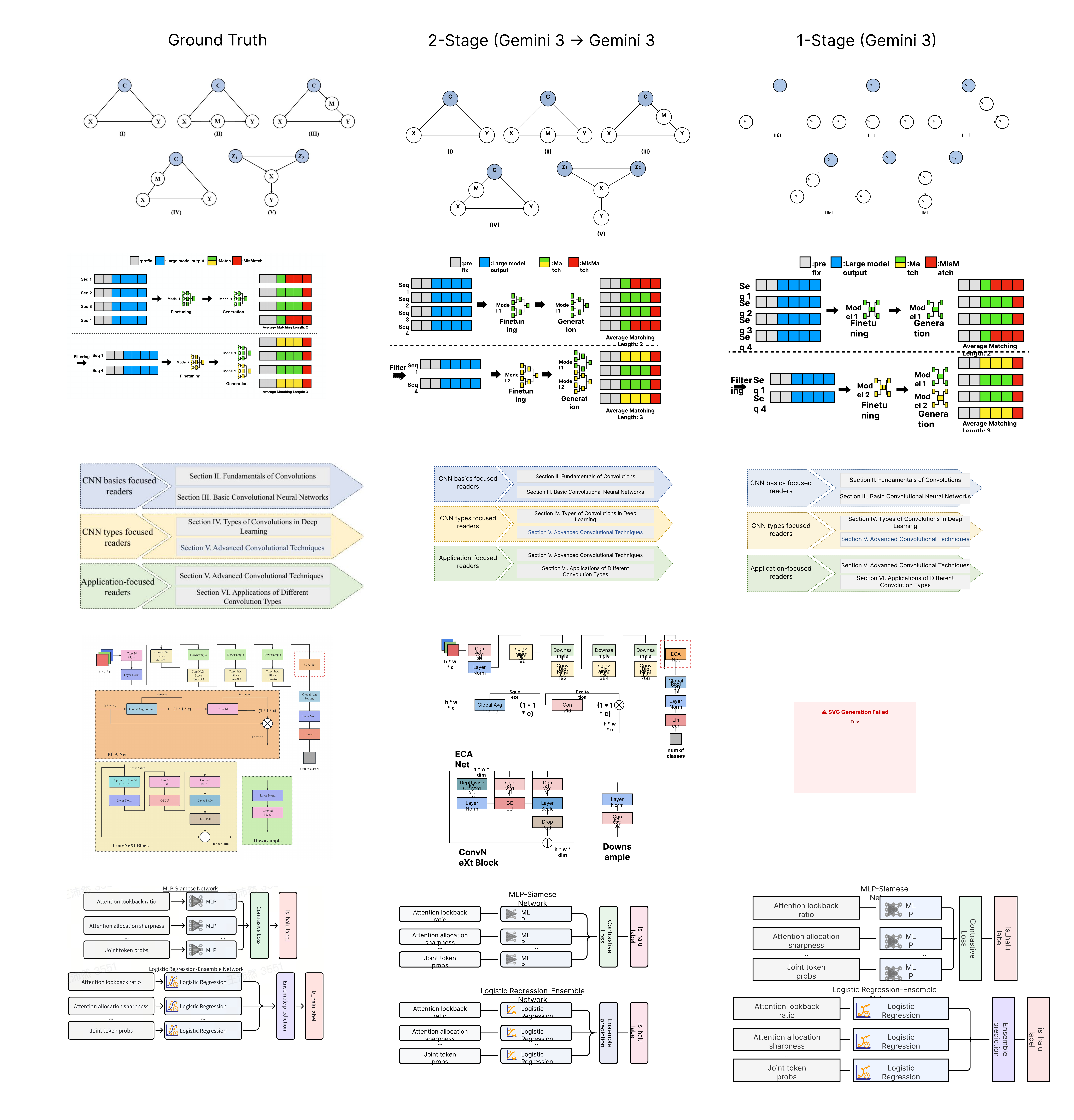}
    \caption{Samples annotated in favor of the 2-stage pipeline based on human evaluation.}
    \label{fig:2v1-stage-examples}
\end{figure}

We arrived at a two-stage describe-and-generate pipeline after evaluating it head-to-head against a single-stage approach in which the model is prompted to produce SVG code directly from the input image. All metrics reported here are computed on the \texttt{arxiv-famous-figures} dataset. While the margins are modest, the two-stage pipeline wins consistently across every benchmark we measured: SSIM scores improve by roughly 0.02 (from a base of $\sim$0.65), render success rate is 99\% versus 98\% for the single-stage baseline, GPT-based evaluation favors the two-stage outputs 91\% to 90\%, and small-scale human preference surveys likewise tilted toward the two-stage results. Qualitatively, the gap is most visible in high fidelity figures with intricate notation, dense arrows, and unique spatial layouts. Some side-by-side comparisons are shown in Figure~\ref{fig:2v1-stage-examples}. 

\begin{figure}[h]
    \centering
    \includegraphics[width=\linewidth]{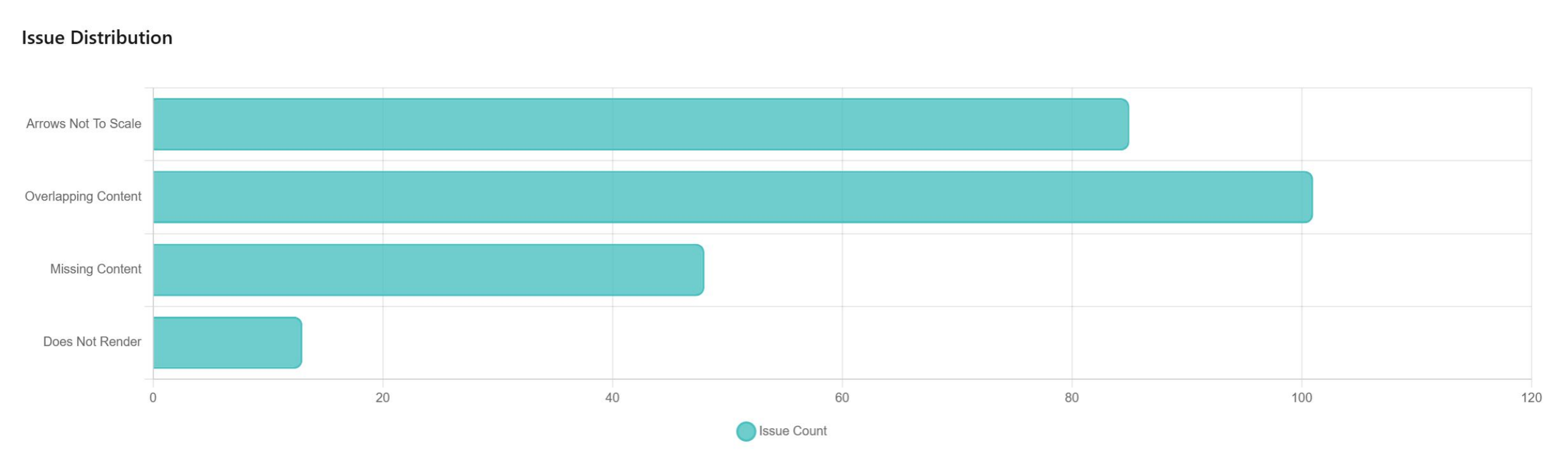}
    \caption{Distribution of annotated issues across 332 evaluated samples.}
    \label{fig:issue_dist}
\end{figure}

\begin{figure}[h]
    \centering
    \includegraphics[width=\linewidth]{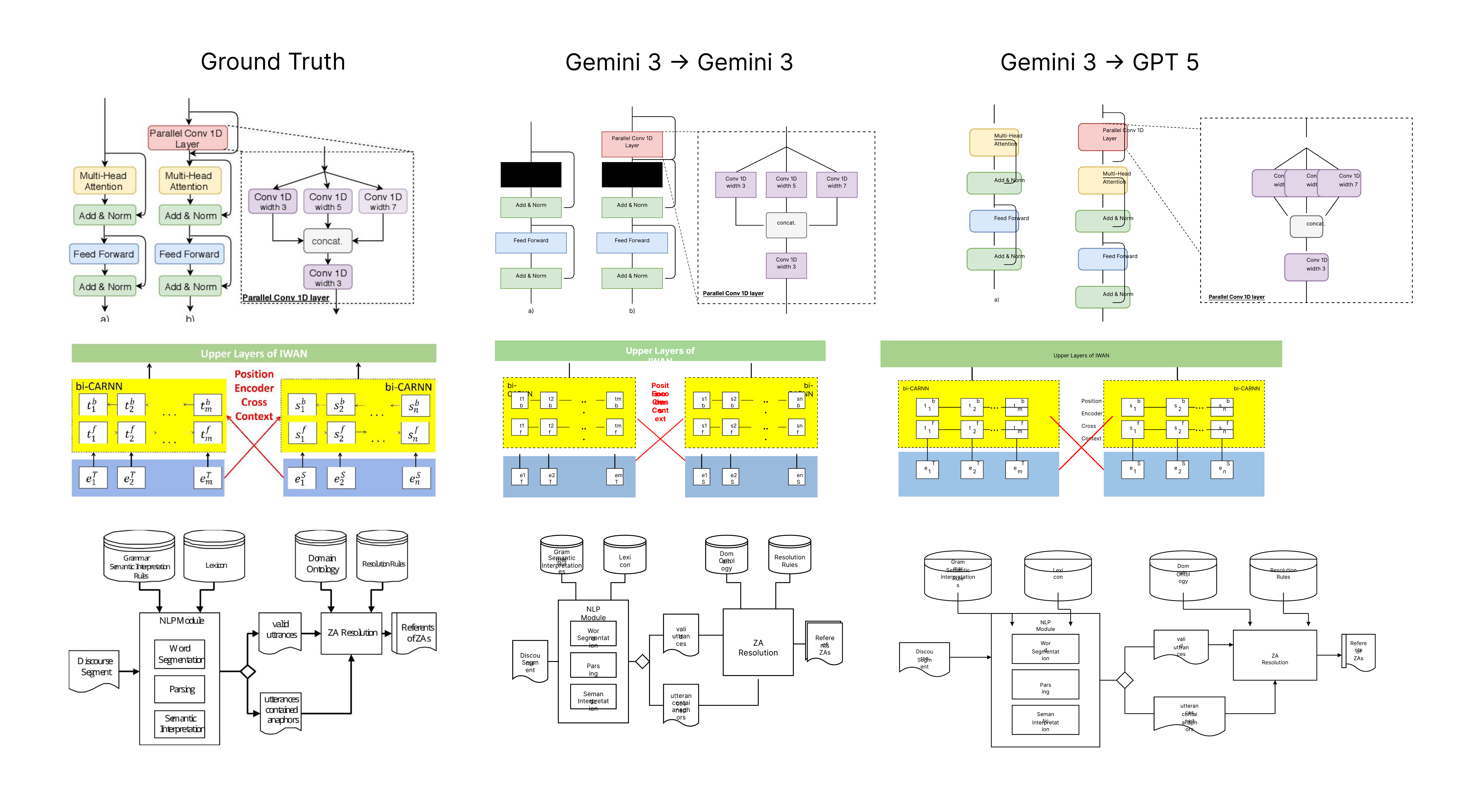}
    \caption{Examples where the Gemini-3-Pro $\rightarrow$ Gemini-3-Pro pipeline produced superior SVG outputs.}
    \label{fig:gemini_win}
\end{figure}

\begin{figure}[h]
    \centering
    \includegraphics[width=\linewidth]{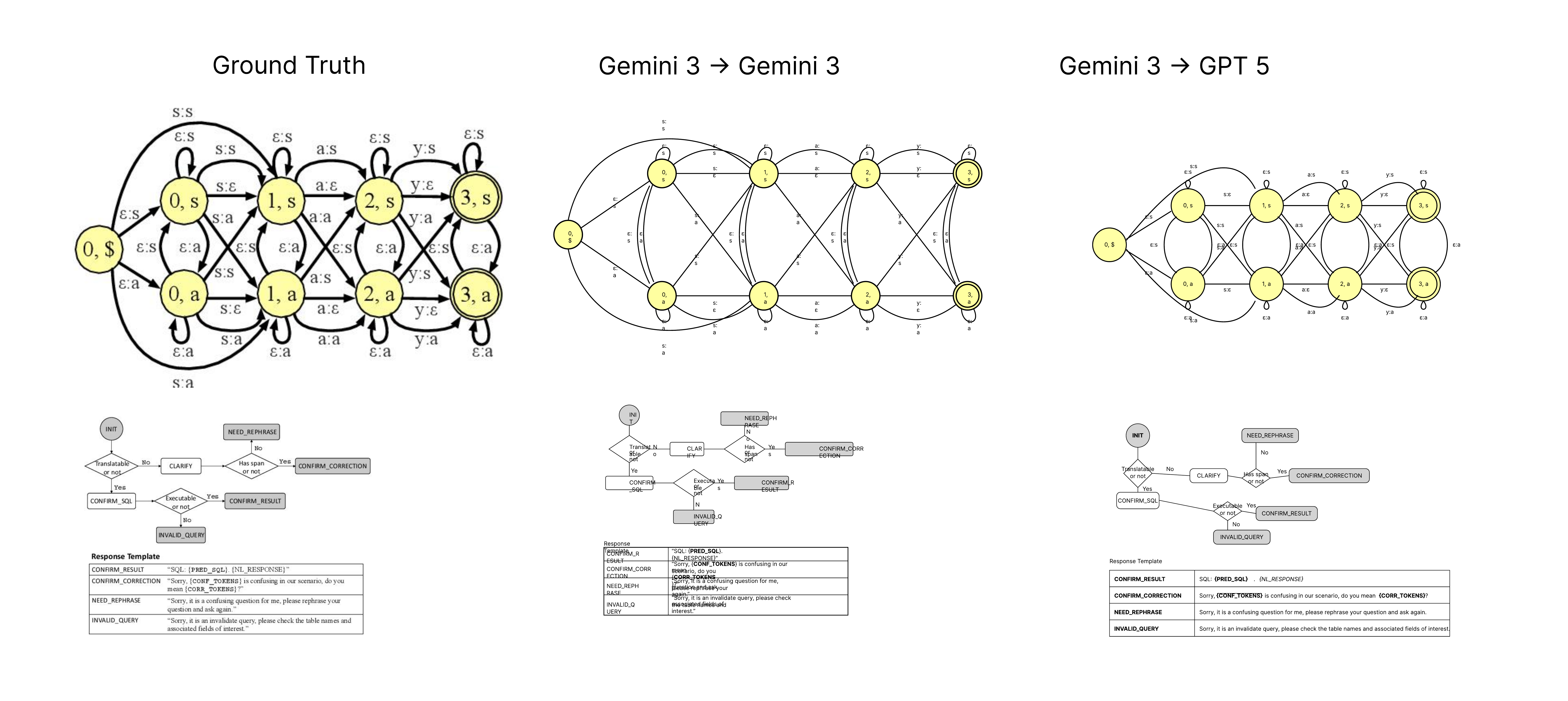}
    \caption{Examples where the Gemini-3-Pro $\rightarrow$ GPT-5.1 pipeline produced superior SVG outputs.}
    \label{fig:gpt_win}
\end{figure}

To obtain the best model selection for this SVG generation pipeline, we developed a web sandbox\footnote{\url{https://llmsvg.netlify.app/}} for qualitative comparison for all models from the Gemini, GPT, and Anthropic families. From this initial screening, we identified two competitive describe-and-generate candidates for deeper evaluation: Gemini-3-Pro $\rightarrow$ Gemini-3-Pro, and Gemini-3-Pro $\rightarrow$ GPT-5.1. In both pipelines, the \emph{describe} stage produces a detailed textual description of the input image, and the \emph{generate} stage converts that description into SVG code.
To compare these, we conducted an internal human evaluation survey across 332 tasks, collecting 1088 responses from 5 annotators. Annotators identified the superior pipeline for a given sample or chose "both good" or "both bad" and flagged issues across four categories: arrows not to scale, overlapping content, missing content, and rendering failures for the winning inference(s). As shown in Figure~\ref{fig:issue_dist}, the most common failure modes were overlapping content and arrows not to scale, while outright rendering failures were rare. Based on this evaluation, the Gemini-3-Pro $\rightarrow$ Gemini-3-Pro pipeline was preferred in 88.7\% of decisive responses, compared to 11.3\% for Gemini-3-Pro $\rightarrow$ GPT-5.1, and was adopted as our generation backbone. All evaluated examples and results are publicly accessible.\footnote{\url{https://geminivsgpt.netlify.app/}} Representative examples of cases where each pipeline produced superior outputs are shown below (Figure~\ref{fig:gemini_win} and \ref{fig:gpt_win}).

The prompts used for each stage are as follows:

\begin{tcolorbox}[
breakable,
colback=gray!5,
colframe=gray!60,
title={Prompt: Description Stage},
fonttitle=\bfseries
]
\footnotesize
Please provide an extremely detailed description of this image. Describe every visual element, including shapes, colors, positions, sizes, text content, styles, patterns, gradients, shadows, borders, and any other visual characteristics. Be as thorough as possible, as this description will be used to recreate the image as an SVG.
\end{tcolorbox}

\begin{tcolorbox}[
breakable,
colback=gray!5,
colframe=gray!60,
title={Prompt: SVG Generation Stage},
fonttitle=\bfseries
]
You are a technical SVG code generator. Your task is to convert the attached image into SVG markup that reconstructs all visual elements from the original image. Return only the SVG code, with no explanations, markdown, or additional text. The output should start with \texttt{<svg>} and end with \texttt{</svg>}. Use appropriate SVG elements such as \texttt{path}, \texttt{rect}, \texttt{circle}, and \texttt{text}, and specify a proper \texttt{viewBox}. Ensure that text does not overflow its container, lines do not overlap other elements, and the layout preserves sufficient spacing for readability. Preserve the original colors, structure, and text content, and ensure that arrows are drawn to scale with those in the input image.
\end{tcolorbox}

\subsection{\progds Generation}
\label{app:progds}
Each sample is generated in four sequential steps: selecting a layout, placing shapes,
assigning visual styles, and routing connections. The completed SVG is then rasterized
into a PNG, and all 10 attributes per shape and 6 per arrow are recorded in a paired
JSON metadata file for our rule-based evaluation.

\vspace{0.3em}
\begin{list}{}{
  \setlength{\leftmargin}{0em}
  \setlength{\itemsep}{0.4em}
  \setlength{\parsep}{0pt}
  \setlength{\topsep}{0pt}
}
\item \textbf{Layout Selection.}
Shape positions are seeded by one of 19 layout templates designed to mimic human-made diagrams and avoid outputs with no structure. Each template encodes a unique set of shape positions along with connection hints specifying which shapes should be linked. Randomized jitter is applied to all shape positions to avoid rigid and similar layouts and additional arrows may be generated as discussed below. For additional variance, with probability 0.3, two templates are combined by splitting the canvas in half and applying one to each side, with a small number of cross-connections added between them.  Each layout template also specifies which shapes should be connected by returning a list of template hints or directed index pairs $(i \to j)$. These hints encode the template's intended flow structure, e.g.\ a linear chain, hub-to-spoke edges, or row- and column-adjacency edges. The number of hints varies with the number of placed shapes: templates with fixed role groups (start, middle, end) produce a bounded number, while chain-like templates produce one hint per consecutive shape pair.

\item \textbf{Shape Placement.}
The generator supports 18 shape types: 12 flat shapes (circle, rectangle, square, ellipse,
diamond, hexagon, parallelogram, trapezoid, blob, wave-rect, cloud, text-label) and 6
pseudo-3D shapes (cylinder, prism, cube, 3d-diamond, 3d-hexagon, 3d-trapezoid).
Each diagram draws from a palette of 2--3 randomly selected types, with a bias against
placing the same type consecutively to mimic human-made diagrams. Shape sizes are randomized within a fixed range. Candidate positions are validated using Axis-Aligned Bounding Box (AABB)
collision detection based on bounding box to prevent shape overlap, while enforcing a minimum
center-to-center separation between shapes. If a position fails, alternatives are tried at increasing
radii; shapes that cannot be placed are skipped. With probability 0.15, a flat shape
is rendered as a stack of 2--4 layered copies, each offset and progressively darkened
to suggest depth.

\item \textbf{Visual Style Assignment.}
Each shape is assigned one of 7 fill styles---solid, hatching, dots, crosshatch,
horizontal-lines, linear gradient, or radial gradient---implemented with SVG pattern or
gradient definitions, with parameters randomized per shape. Each diagram also samples a
single border style (solid, dashed, dotted, or dash-dot), stroke width, and with
probability 0.6, rounded corners with a randomized radius. Fill colors are drawn from a
palette of light colors; strokes from dark colors. Arrow colors are selected per diagram
(1--3 distinct colors). Labels come from a domain-specific word bank with word-level
uniqueness enforced per diagram, rendered in 1--2 font families selected from a set of 8.

\item \textbf{Connection Routing.}
Given $n$ placed shapes, the number of directed edges $c$ is drawn from a randomized
range $[n\,r_\ell,\; n\,r_h]$, where $r_\ell$ and $r_h$ are the lower and upper
connection density ratios, each sampled per diagram:
\[
  c \;\sim\; \mathcal{U}\!\left(\lfloor n\, r_\ell \rfloor,\; \lfloor n\, r_h \rfloor\right),
  \qquad r_\ell \sim \mathcal{U}(0.4, 0.6), \quad r_h \sim \mathcal{U}(0.6, 0.8).
\]
Template hints are consumed first; random pairs fill the remainder. Each ordered pair
appears at most once; bidirectional pairs are offset sideways so both arrowheads remain
visible. Each arrow independently samples a stroke width, arrowhead size, and line
pattern (solid, dashed, or dotted), and is drawn as a straight line (60\%) or quadratic
Bezier curve (40\%). Arrow endpoints attach at the exact shape boundary via ray-casting
from the shape center: analytically for circles and ellipses, and via parametric
ray segment intersection for polygons and 3D shapes. Curved arrows use a quadratic
Bezier with a perpendicularly offset control point, we ensure paths don't interesect an intermediate shape to avoid confusion when running inference.
\end{list}

\subsection{Image Filtering} \label{app: img_filter}
We provide additional details for the image filtering stage used to construct the \model{} dataset. 
Specifically, we employ a classifier using Gemini-3-Flash to automatically filter figures extracted from scientific documents, retaining only structured diagram-style figures suitable for SVG generation. The filtering prompt is given below.
\begin{tcolorbox}[
breakable,
colback=gray!5,
colframe=gray!60,
title={Prompt: Figure Filtering Classifier},
fonttitle=\bfseries
]
\footnotesize

You are a highly precise automated figure-filtering classifier.

Return exactly one word from: IMAGE, PLOT, MATH, KEEP.  
Do not output any additional text.

\textbf{Goal.}  
Keep only figures that are structured SVG-style diagrams used for
systems, pipelines, architectures, networks, or flowcharts.
Filter out all other figure types (plots, tables, screenshots,
photos, simulations, textures, memes, anatomy charts, etc.).

Apply the following rules in order. The first matching rule determines
the label.

\textbf{1) PLOT}

Return PLOT if the figure primarily presents data or results, including:
\begin{itemize}
\item line, bar, or scatter plots; histograms; box or violin plots
\item heatmaps, confusion matrices, correlation matrices
\item charts with axes, ticks, legends, or color bars
\item geographic or data maps
\item tables or spreadsheet-style layouts
\end{itemize}

\textbf{2) IMAGE}

Return IMAGE if the figure is primarily illustrative or image-like content,
including:
\begin{itemize}
\item photographs or natural images
\item screenshots of software interfaces or terminals
\item simulation outputs or 3D-rendered scenes
\item qualitative grids of many image panels
\item textures, gradients, or raster patterns
\item memes, emojis, cartoons, or comics
\item infographic-style posters or anatomy diagrams
\end{itemize}

Small illustrative photo insets are allowed only if the overall figure
remains a structured diagram.

\textbf{3) MATH}

Return MATH if the figure is dominated by mathematical notation,
including multiple equations, integrals, summations, matrices, or
proof-like derivations.

If the figure is a structured diagram containing only small
mathematical annotations, return KEEP instead.

\textbf{4) KEEP}

Return KEEP only if the figure is primarily a structured diagram
typically drawn using SVG primitives in research papers, including:
\begin{itemize}
\item flowcharts, pipelines, block diagrams, architecture diagrams
\item diagrams composed mainly of text, boxes, arrows, and icons
\item coordinate schematics with axes or reference grids
\item diagrams containing small mathematical annotations
\item diagrams with a few small illustrative photo insets
\end{itemize}

\end{tcolorbox}

\subsection{Code Filtering}
\label{app:code_filtering}
We provide the full thresholds and cleaning rules for the SVG code filtering stage summarized in section~\ref{sec:data_filtering}. We group SVG elements into three categories: basic shapes (\texttt{rect}, \texttt{circle}, \texttt{ellipse}), connectors (\texttt{line}, \texttt{polyline}), and complex shapes (\texttt{path}, \texttt{polygon}). Let $N$ be the total count of geometric elements (sum of the three groups). We enforce two rules to retain primitive-dominated diagrams:
\begin{enumerate}
  \item the proportion of basic shapes and connectors must be at least $40\%$ of $N$;
  \item the absolute number of complex shapes must not exceed $50$.
\end{enumerate}
These rules filter out tracing-style SVGs dominated by long \texttt{path} sequences while preserving diagram-style figures composed of simple geometric primitives. We further apply light cleaning to reduce syntactic noise: removing redundant metadata, standardizing canvas settings, and normalizing coordinate precision without affecting visual fidelity. Corrupted samples with abnormal repeated numeric or character patterns are discarded.

\subsection{SVG Structural Complexity Metrics}
\label{app: svg complexity metrics}
To quantitatively characterize the structural properties of SVG datasets, we define a set of interpretable metrics that measure geometric complexity and semantic organization. These metrics are designed to correlate with sequence modeling difficulty while remaining simple and reproducible.

Given an SVG file, elements are counted in a case-insensitive manner and grouped into four categories: 
(i) basic primitives (\texttt{rect}, \texttt{circle}, \texttt{ellipse}), 
(ii) connectors (\texttt{line}, \texttt{polyline}), 
(iii) complex shapes (\texttt{path}, \texttt{polygon}), and 
(iv) text (\texttt{text}). 

We denote
\begin{align}
B &= \#(\text{basic primitives}), \\
K &= \#(\text{connectors}), \\
C &= \#(\text{complex shapes}), \\
T &= \#(\text{text elements}), \\
N &= B + K + C,
\end{align}
where $N$ counts geometric elements only (excluding text). A small constant $\epsilon$ (e.g., $10^{-9}$) is used to avoid division by zero when necessary.

\noindent\textbf{Structural Complexity (SC).}
Structural Complexity is the raw count of geometric elements per SVG, serving as a direct proxy for the size of the structural element set the model must generate:
\begin{equation}
\mathrm{SC} = N = B + K + C.
\end{equation}
SC is reported on a linear scale (without logarithmic transformation), making it interpretable as a literal element count but sensitive to long-tailed figures.

\noindent\textbf{Element Complexity (EC).}
Element Complexity measures the combined geometric and labeling burden of a figure on a log scale:
\begin{equation}
\mathrm{EC} = \log\left(1 + N + T\right).
\end{equation}
Compared to SC, EC additionally accounts for text annotations ($T$) and applies a logarithmic transform that stabilizes the metric for large figures and reduces sensitivity to extreme outliers.

\noindent\textbf{Semantic Cleanliness (Clean).}
Semantic Cleanliness measures the proportion of semantic primitives and connectors relative to path-based tracing:
\begin{equation}
\mathrm{Clean} = \frac{B + K}{N }
= 1 - \frac{C}{N }.
\end{equation}
Values close to 1 indicate structured, editable SVGs composed primarily of semantic primitives and connectors, while lower values indicate path-dominated representations typical of tracing-based SVGs.

\noindent\textbf{Path Dominance (PD).}
For completeness, we also report Path Dominance:
\begin{equation}
\mathrm{PD} = \frac{C}{N }
= 1 - \mathrm{Clean}.
\end{equation}
PD captures the extent to which an SVG relies on complex tracing-style elements.

\begin{figure}[t]
    \centering
    \includegraphics[width=1\linewidth]{sections//figure/structural_complexity_metric.pdf}
    \caption{Examples illustrating the structural complexity metrics. 
For each SVG diagram we report element counts ($B,K,C,T$) and the derived metrics EC, Clean, and PD. 
Structured diagrams typically exhibit higher cleanliness and lower path dominance.}
    \label{fig:structural_complexity_metric}
\end{figure}

Tracing-based SVGs typically exhibit higher element complexity and higher path dominance, which increases sequence length and reduces structural interpretability. In contrast, semantically structured SVG representations emphasize primitives and connectors, yielding higher cleanliness and improved controllability for sequence modeling and editing tasks.

\subsection{Rule-Based Benchmark for \progds.} 
\label{app: rule based eval}

Since our \progds dataset includes structured metadata describing element attributes and relationships (Section~\ref{sec:programmatic_figure_gen}), we evaluate generated SVGs directly against this ground-truth specification. For each sample we compute $R = (R_{\text{S}} + R_{\text{A}})/2$, where

$R_{\text{S}}$: Shapes are matched to ground-truth metadata by label, then scored across nine visual attributes: shape type, fill and stroke color, fill and border style, font, aspect ratio, and relative spatial position.

$R_{\text{A}}$: Arrows are matched by endpoint proximity to their expected source and destination shapes, then scored across seven attributes: connectivity, arrowhead presence and size, curvature, and color.

\begin{figure}[h]
\centering
{\small Side-by-side comparison: ground truth (left) vs.\ inference (right)}\\[4pt]
\begin{minipage}[c]{0.48\textwidth}
    \centering
    \includegraphics[width=\textwidth]{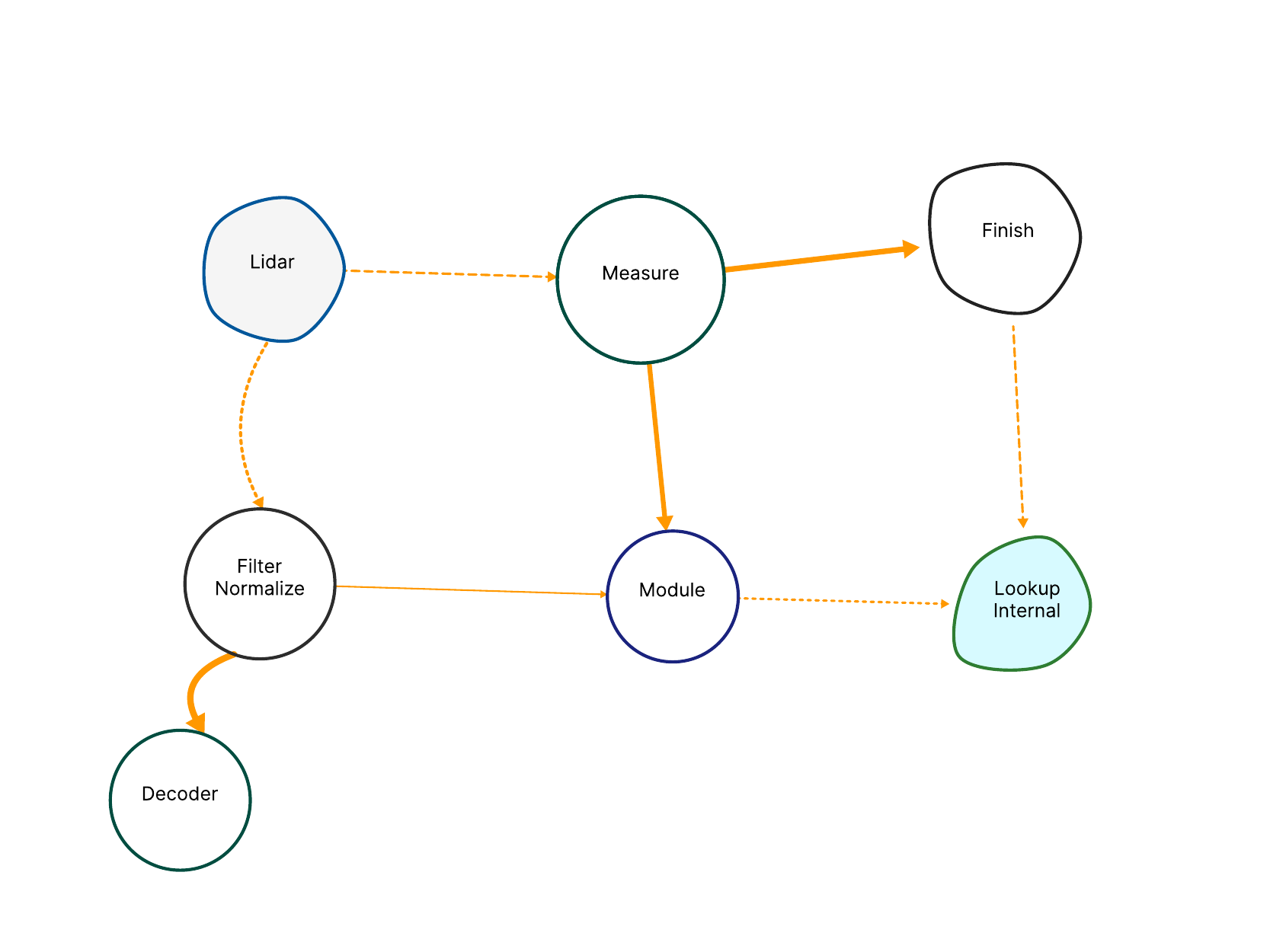}
    \footnotesize Ground Truth
\end{minipage}
\hfill
\begin{minipage}[c]{0.48\textwidth}
    \centering
    \includegraphics[width=\textwidth]{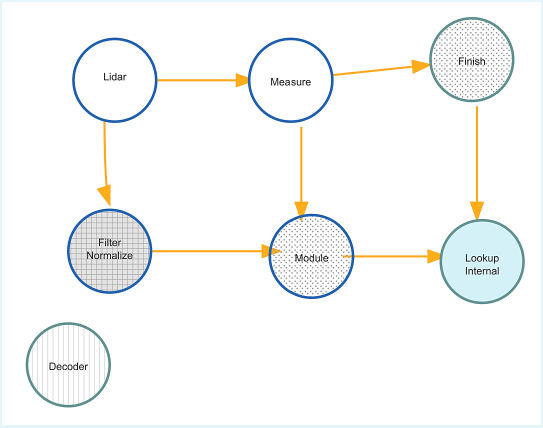}
    \footnotesize Inference (generated by VFig)
\end{minipage}
\vspace{10pt}

\begin{minipage}[c]{0.42\textwidth}
    \vspace{0pt}
    \includegraphics[width=\textwidth]{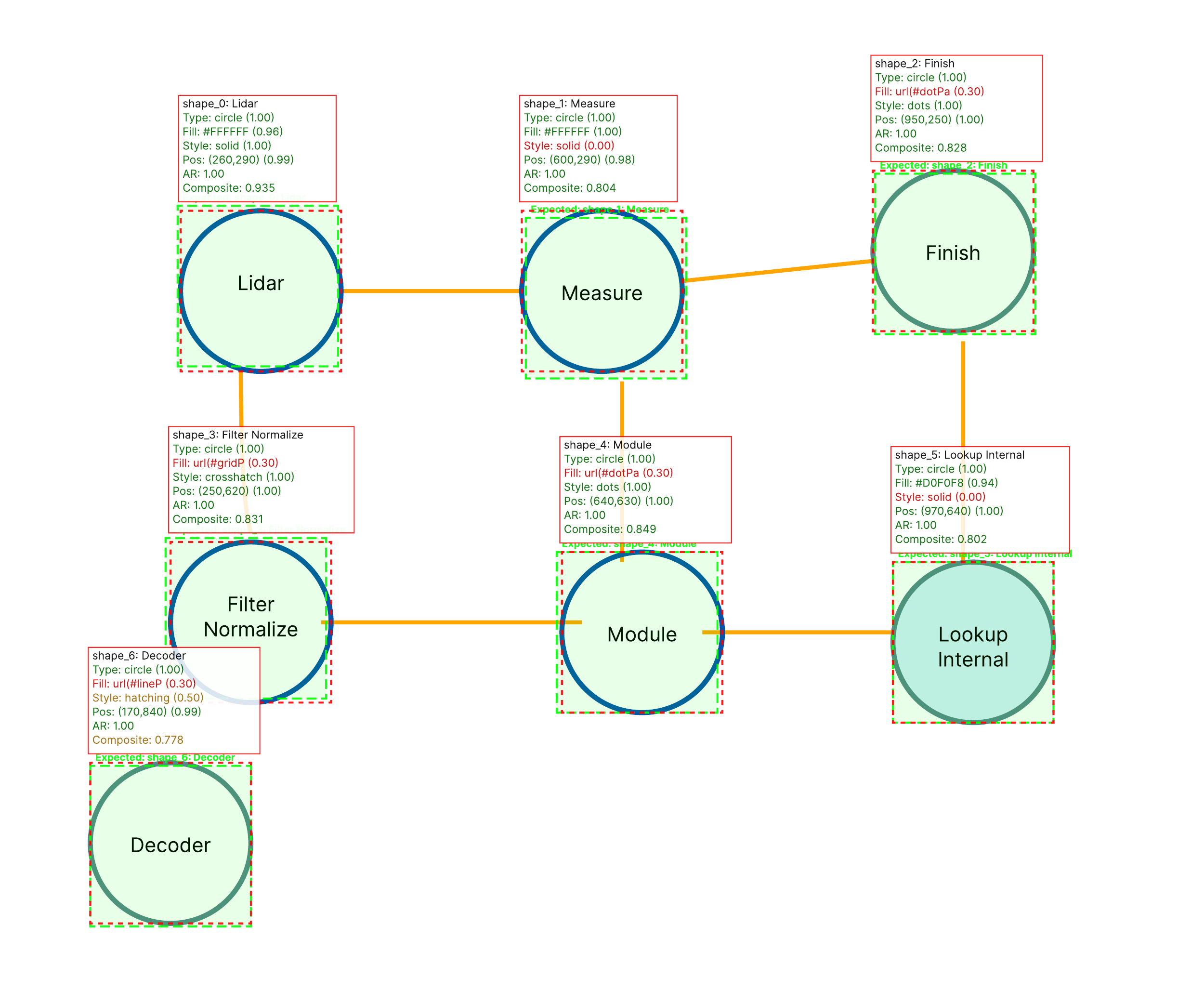}
\end{minipage}
\hfill
\begin{minipage}[c]{0.55\textwidth}
\centering
{\small Shape attributes}\\[2pt]
\resizebox{\textwidth}{!}{%
\footnotesize
\begin{tabular}{llccccccc}
\toprule
\# & Label & Fill & Style & Stroke & Font & Pos & AR & Comp \\
\midrule
0 & Lidar          & \scc{0.961} & \scc{1.000} & \scc{0.966} & \scc{0.500} & \scc{0.985} & \scc{1.000} & \scc{0.935} \\
1 & Measure        & \scc{0.996} & \scc{0.000} & \scc{0.759} & \scc{0.500} & \scc{0.977} & \scc{1.000} & \scc{0.804} \\
2 & Finish         & \scc{0.300} & \scc{1.000} & \scc{0.650} & \scc{0.500} & \scc{0.999} & \scc{1.000} & \scc{0.828} \\
3 & Filter Norm.   & \scc{0.300} & \scc{1.000} & \scc{0.684} & \scc{0.500} & \scc{1.000} & \scc{1.000} & \scc{0.831} \\
4 & Module         & \scc{0.300} & \scc{1.000} & \scc{0.838} & \scc{0.500} & \scc{0.999} & \scc{1.000} & \scc{0.849} \\
5 & Lookup Int.    & \scc{0.937} & \scc{0.000} & \scc{0.785} & \scc{0.500} & \scc{1.000} & \scc{1.000} & \scc{0.802} \\
6 & Decoder        & \scc{0.300} & \scc{0.500} & \scc{0.715} & \scc{0.500} & \scc{0.992} & \scc{1.000} & \scc{0.778} \\
\midrule
\multicolumn{2}{l}{\textit{Avg.}} & \scc{0.585} & \scc{0.643} & \scc{0.771} & \scc{0.500} & \scc{0.993} & \scc{1.000} & \scc{0.832} \\
\bottomrule
\end{tabular}}
\end{minipage}
\vspace{10pt}

\begin{minipage}[c]{0.42\textwidth}
    \vspace{0pt}
    \includegraphics[width=\textwidth]{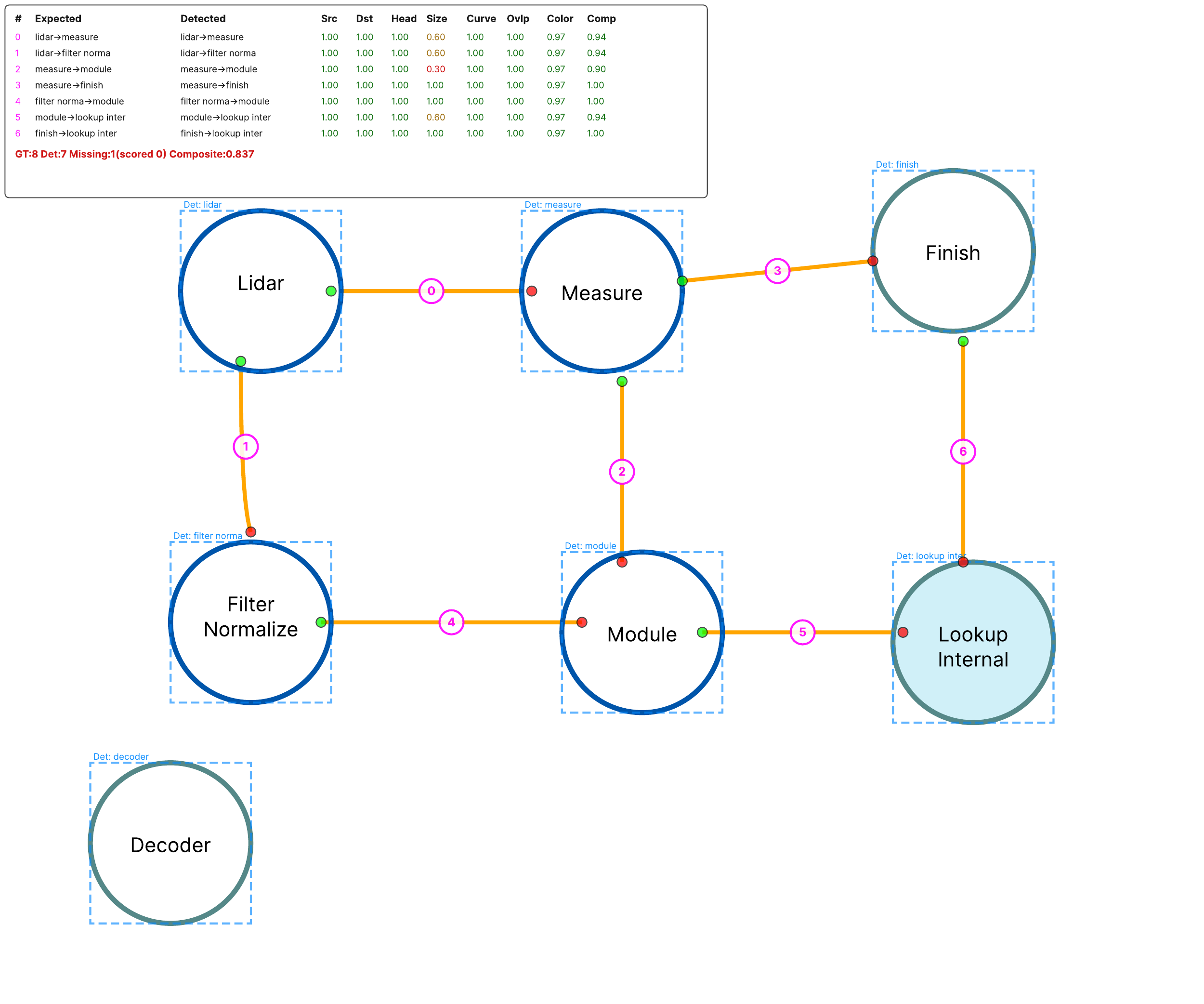}
\end{minipage}
\hfill
\begin{minipage}[c]{0.55\textwidth}
\centering
{\small Arrow attributes}\\[2pt]
\resizebox{\textwidth}{!}{%
\footnotesize
\begin{tabular}{lllcccccccc}
\toprule
\# & Expected & Detected & Src & Dst & Head & Size & Curve & Ovlp & Color & Comp \\
\midrule
0 & emit$\to$obs.       & emit$\to$obs.       & \scc{1.00} & \scc{1.00} & \scc{1.00} & \scc{0.10} & \scc{1.00} & \scc{0.40} & \scc{0.84} & \scc{0.76} \\
1 & hist.$\to$dist.     & hist.$\to$dist.     & \scc{1.00} & \scc{1.00} & \scc{1.00} & \scc{0.10} & \scc{1.00} & \scc{1.00} & \scc{0.84} & \scc{0.85} \\
2 & obs.$\to$hist.      & obs.$\to$hist.      & \scc{1.00} & \scc{1.00} & \scc{1.00} & \scc{0.30} & \scc{1.00} & \scc{1.00} & \scc{0.84} & \scc{0.88} \\
3 & proj.$\to$train.    & proj.$\to$train.    & \scc{1.00} & \scc{1.00} & \scc{1.00} & \scc{0.10} & \scc{1.00} & \scc{1.00} & \scc{0.84} & \scc{0.85} \\
4 & dist.$\to$proj.     & dist.$\to$proj.     & \scc{1.00} & \scc{1.00} & \scc{1.00} & \scc{0.60} & \scc{1.00} & \scc{1.00} & \scc{0.84} & \scc{0.92} \\
5 & hist.$\to$proj.     & hist.$\to$proj.     & \scc{1.00} & \scc{1.00} & \scc{1.00} & \scc{0.30} & \scc{1.00} & \scc{1.00} & \scc{0.84} & \scc{0.88} \\
\bottomrule
\end{tabular}}
\end{minipage}
\caption{Rule-based evaluation for shapes (top) and arrows (bottom) on sample 000024.}
\end{figure}


Each metric produces a score in $[0,1]$, where higher is better. All color comparisons use the
$\ell_2$ distance in RGB space normalized by the diameter of the RGB
cube, mapping any color pair to $[0,1]$. 
\begin{table*}[t]
\centering
\caption{Rule-based evaluation on \progds\ (500 samples). All scores in
  $[0,1]$; higher is better. $R_S$ and $R_A$ are shape and arrow
  composites after extra-element penalty; $R$ is their mean.}
\label{tab:arrow_result}
\setlength{\tabcolsep}{1.1pt}
\resizebox{\textwidth}{!}{%
\scriptsize
\begin{tabular}{l ccc cc c ccccccccc c ccccccc}
\toprule
& \multicolumn{3}{c}{Composite} & \multicolumn{2}{c}{Coverage} & & \multicolumn{9}{c}{Shape attributes} & & \multicolumn{7}{c}{Arrow attributes} \\
\cmidrule(lr){2-4}\cmidrule(lr){5-6}\cmidrule(lr){8-16}\cmidrule(lr){18-24}
Model
  & $R$ & $R_S$ & $R_A$
  & Miss. & Err.
  & & Lbl & Typ & FC & FS & SC & BS & Pos & Fnt & AR
  & & Src & Dst & Hd & Sz & Cv & Col & Ovl \\
\midrule
\multicolumn{24}{l}{\textit{Open-source}} \\
\midrule
\model (SFT + RL)
  & \textbf{0.638} & \textbf{0.675} & \textbf{0.602}
  & 0 & 15
  & & \textbf{0.870} & 0.650 & 0.750 & 0.370 & \textbf{0.750} & \textbf{0.790} & \textbf{0.860} & \textbf{0.260} & \textbf{0.800}
  & & \textbf{0.650} & \textbf{0.670} & \textbf{0.790} & \textbf{0.680} & 0.650 & \textbf{0.710} & \textbf{0.650} \\
\model (SFT)
  & 0.618 & 0.654 & 0.583
  & 0 & 24
  & & 0.850 & \textbf{0.670} & \textbf{0.770} & 0.270 & 0.720 & 0.770 & 0.840 & 0.230 & 0.780
  & & 0.620 & 0.640 & 0.750 & 0.650 & 0.620 & 0.670 & 0.620 \\
Qwen3-VL
  & 0.274 & 0.257 & 0.291
  & 0 & 235
  & & 0.430 & 0.330 & 0.110 & 0.160 & 0.100 & 0.390 & 0.420 & 0.000 & 0.390
  & & 0.310 & 0.340 & 0.400 & 0.330 & 0.270 & 0.330 & 0.330 \\
OmniSVG$^{\ddag}$
  & 0.000 & 0.000 & 0.000
  & 142 & 0
  & & 0.000 & 0.000 & 0.000 & 0.000 & 0.000 & 0.000 & 0.000 & 0.000 & 0.000
  & & 0.000 & 0.000 & 0.000 & 0.000 & 0.000 & 0.000 & 0.000 \\
StarVector$^{\mathsection}$
  & 0.000 & 0.000 & 0.000
  & 0 & 497
  & & 0.000 & 0.000 & 0.000 & 0.000 & 0.000 & 0.000 & 0.000 & 0.000 & 0.000
  & & 0.000 & 0.000 & 0.000 & 0.000 & 0.000 & 0.000 & 0.000 \\
\midrule
\multicolumn{24}{l}{\textit{Closed-source}} \\
\midrule
Gemini-3-Pro
  & \textbf{0.611} & \textbf{0.595} & \textbf{0.627}
  & 0 & 2
  & & \textbf{0.730} & \textbf{0.590} & \textbf{0.530} & \textbf{0.510} & \textbf{0.610} & \textbf{0.710} & \textbf{0.720} & \textbf{0.250} & \textbf{0.720}
  & & \textbf{0.640} & \textbf{0.650} & \textbf{0.730} & \textbf{0.570} & \textbf{0.660} & \textbf{0.700} & \textbf{0.690} \\
GPT-5.2
  & 0.409 & 0.459 & 0.358
  & 0 & 1
  & & 0.620 & 0.490 & 0.370 & 0.340 & 0.400 & 0.590 & 0.600 & 0.160 & 0.590
  & & 0.400 & 0.410 & 0.450 & 0.360 & 0.260 & 0.420 & 0.370 \\
\bottomrule
\end{tabular}}
\vspace{2pt}
\begin{minipage}{\textwidth}
\scriptsize
$^{\ddag}$OmniSVG is evaluated on the original transparent-background SVG inputs as in the main paper. 
Here the inputs use white backgrounds, causing many outputs to be rejected as \texttt{empty\_image}, so no file is saved.
\par
$^{\mathsection}$ 497 of 500 StarVector outputs cannot be parsed as valid SVG; the model is therefore excluded from the main comparison.
\end{minipage}
\end{table*}
We provide detailed definitions for each metric below.

\emph{Shape attributes:}

\emph{Label (Lbl).}
Measures text-label correctness. A score of 1.0 is assigned for an exact
match, with partial credit proportional to word overlap and 0.0 for no
overlap.

\emph{Type (Typ).}
Evaluates whether the generated SVG element type (e.g.,
\texttt{rect}, \texttt{ellipse}, \texttt{path}, \texttt{g})
matches the expected type. Accepted tags receive 1.0, plausible
substitutes receive 0.3--0.4, and incorrect elements receive 0.0.

\emph{Fill color (FC).}
Measures fill color accuracy using normalized RGB distance.
Ground-truth colors are read from the JSON metadata rather than the SVG
directly to avoid ambiguities from blended or patterned fills.

\emph{Fill style (FS).}
Evaluates the fill pattern (solid, dots, crosshatch, hatching, or
gradient). The generated pattern is inferred from the SVG
\texttt{<defs>} section. Exact matches score 1.0; mismatched patterned
styles score 0.5; solid vs.\ patterned mismatches score 0.0.

\emph{Stroke color (SC).}
Border color accuracy measured using the same normalized RGB distance
as fill color.

\emph{Border style (BS).}
Evaluates the stroke dash pattern (solid, dashed, dotted, or dash-dot)
using the \texttt{stroke-dasharray} attribute. Exact matches score 1.0;
different non-solid patterns score 0.5; solid vs.\ non-solid mismatches
score 0.0.

\emph{Position (Pos).}
Measures spatial layout using anchor-relative offsets, making the score
invariant to global translation and scale differences between the
generated and ground-truth canvases.

\emph{Font (Fnt).}
Checks the font family, mapped to a class (serif, sans-serif,
monospace). Exact name matches score 1.0, same-class matches score 0.5,
and mismatches score 0.0.

\emph{Aspect ratio (AR).}
Measures width-to-height ratio accuracy. Ratios within 20\% of the
ground truth score 1.0, with the score decaying linearly to 0.0 at a
$3\times$ deviation.

\emph{Arrow attributes:}

\emph{Source / destination (Src, Dst).}
Evaluates whether arrow endpoints connect to the correct shapes.
Endpoints are matched to the nearest shape bounding box within
100\,px, and the matched label is compared to the ground-truth
connection. The reversed orientation is also considered and the better
match kept. Binary scoring: 1.0 for correct, 0.0 otherwise.

\emph{Head (Hd).}
Checks for the presence of an arrowhead using the
\texttt{marker-end} attribute or a polygon sibling in the same group.
If the ground-truth arrowhead is visually occluded by another shape,
missing arrowheads are not penalized.

\emph{Head size (Sz).}
Evaluates arrowhead size relative to stroke width. Both dimensions are
normalized by stroke width before comparison. Scores are assigned in
four tiers, ranging from 1.0 (within 30\%) to 0.1 (more than
$2.5\times$ deviation).

\emph{Curve (Cv).}
Determines whether arrow curvature matches the ground truth (straight
or curved) by detecting Bézier or arc commands in the SVG path. Binary
scoring: 1.0 for a match, 0.0 otherwise.

\emph{Color (Col).}
Measures arrow stroke color accuracy using normalized RGB distance,
with CSS inheritance resolved from ancestor elements and style
definitions.

\emph{Overlap (Ovl).}
Penalizes arrows whose endpoints fall inside shapes they are not
supposed to connect to. Scores are 1.0 for no violations, 0.4 for one
incorrect endpoint, and 0.0 for two or more violations.

\subsection{Standalone Rule-Based Evaluation and Benchmarks}
\label{app:standalone_results}

The evaluation in Table~\ref{tab:arrow_result} relies on structured JSON
metadata bundled with \progds, which provides ground-truth labels, colors,
and layout information directly. For real-world SVGs where no such metadata
exists, we instead extract equivalent ground-truth attributes directly from
the ground truth SVG file. This standalone approach is necessarily noisier,
however validation on \progds shows negligible differences from the
JSON-based evaluation across all models and metrics, with mean absolute gaps
of at most $0.003$ on $R_S$, $0.001$ on $R_A$, and $0.002$ on $R$ overall,
confirming that standalone attribute extraction is a reliable proxy for the
JSON ground truth.

We then use this standalone rule-based evaluation to benchmark models on external datasets: arXiv
figures (\textbf{arXiv}) and diagrams generated by Molmo
(\textbf{Molmo}). The actual metrics and their weighting is unchanged, however we are using new methods for deriving the values from the ground truth SVG instead of using the JSON metadata generated with the dataset.
Table~\ref{tab:standalone_results} reports composite scores for all models
evaluated on these benchmarks. 

It is worth noting that this rule-based approach depends on accurately extracting visual attributes from generated SVG markup, which can both under- and over-estimate true visual quality. Models using unconventional but valid element structures may be penalized unfairly, while models producing invisible or malformed elements may receive undeserved credit. Scores are therefore best interpreted as an approximation of fidelity rather than an exact measure. By using supersets of valid SVG generations across all models, a variety of evaluation datasets, and fine-tuning detection attribute extraction algorithms to input formats, we mitigate these drawbacks.

\begin{table}[!ht]
\centering
\caption{Standalone rule-based evaluation on external benchmarks.
  $R_S$ and $R_A$ are shape and arrow composites; $R$ is their mean.
  Bold indicates the best score within each group (open- and
  closed-source) per column. $^\star$Qwen3-VL excluded from arXiv
  due to failure rate $>50\%$.}
\label{tab:standalone_results}
\setlength{\tabcolsep}{3.5pt}
\small
\begin{tabular}{l ccc c ccc c ccc}
\toprule
& \multicolumn{3}{c}{arXiv} & & \multicolumn{3}{c}{Molmo} & & \multicolumn{3}{c}{VFIG-Data-Shapes...} \\
\cmidrule(lr){2-4}\cmidrule(lr){6-8}\cmidrule(lr){10-12}
Model & $R$ & $R_S$ & $R_A$ & & $R$ & $R_S$ & $R_A$ & & $R$ & $R_S$ & $R_A$ \\
\midrule
\multicolumn{12}{l}{Open-source} \\
\midrule
\model (SFT + RL)
  & \textbf{0.464} & \textbf{0.312} & \textbf{0.615}
  & & \textbf{0.580} & \textbf{0.353} & \textbf{0.806}
  & & \textbf{0.612} & \textbf{0.657} & \textbf{0.566} \\
Qwen3-VL
  & $^\star${} & $^\star${} & $^\star${}
  & & 0.359 & 0.219 & 0.499
  & & 0.307 & 0.337 & 0.277 \\
OmniSVG
  & 0.244 & 0.000 & 0.488
  & & 0.354 & 0.000 & 0.707
  & & 0.000 & 0.000 & 0.000 \\
\midrule
\multicolumn{12}{l}{Closed-source} \\
\midrule
GPT-5.2
  & \textbf{0.469} & \textbf{0.315} & \textbf{0.623}
  & & \textbf{0.608} & \textbf{0.379} & \textbf{0.837}
  & & \textbf{0.649} & \textbf{0.697} & 0.601 \\
Gemini-3-Pro
  & 0.436 & 0.292 & 0.581
  & & 0.555 & 0.332 & 0.777
  & & 0.618 & 0.633 & \textbf{0.602} \\
\bottomrule
\end{tabular}

\smallskip
\raggedright\footnotesize
Scores are computed over the shared superset of samples for which all
non-excluded models produced a valid generation, so that comparisons
are made on identical inputs.
\end{table}
Across all three external benchmarks, \model (SFT + RL) is the strongest open-source
model, leading on $R$, $R_S$, and $R_A$ on all evaluation datasets, and
coming close to the closed-source models on Molmo. Gemini-3-Pro and
GPT-5.2 perform on par overall. OmniSVG scores zero on all shape metrics due to output being rendered with raw paths instead of shape SVG elements. Qwen3-VL has high parse failure rates and poor fill and stroke scores across benchmarks but achieves reasonable arrow scores.

In per-metric tables we see shape fidelity ($R_S$) is the harder sub-task for all models. Font is a consistent weak point, with scores rarely exceeding $0.27$, as models default to generic system fonts regardless of the ground-truth family. Fill style is similarly low for fine-tuned models, suggesting that training on structural accuracy comes at the cost of visual fidelity.

Arrow scores ($R_A$) are generally strong, particularly on Molmo where
the top models exceed $0.80$. Source and destination matching is
reliable across models, indicating that connectivity is well learned.
Head size is a consistent weak point, with models frequently
over- or under-scaling arrowheads relative to stroke width.

More detailed per-metric breakdowns of model performance on each dataset using this standalone evaluation are given in
  Tables~\ref{tab:permetric_arxiv}, \ref{tab:permetric_molmo},
  and~\ref{tab:permetric_figdata}.

\begin{table}[!ht]
\centering
\caption{Per-metric evaluation on \textbf{arXiv} (superset of $321$ samples).
  Bold indicates the best score within each group (open- and closed-source)
  per row. Qwen3-VL excluded due to failure rate $>50\%$.}
\label{tab:permetric_arxiv}
\setlength{\tabcolsep}{3.5pt}
\small
\begin{tabular}{l cc c cc}
\toprule
& \multicolumn{2}{c}{Closed-source} & & \multicolumn{2}{c}{Open-source} \\
\cmidrule(lr){2-3}\cmidrule(lr){5-6}
Metric & GPT-5.2 & Gem.-3-Pro & & RL & OmniSVG \\
\midrule
\multicolumn{6}{l}{\itshape Shape attributes} \\
\midrule
Label        & 0.463 & \textbf{0.469} & & \textbf{0.466} & 0.000 \\
Type         & 0.419 & \textbf{0.432} & & \textbf{0.431} & 0.000 \\
Fill Color   & \textbf{0.317} & \textbf{0.317} & & \textbf{0.318} & 0.000 \\
Fill Style   & 0.211 & \textbf{0.214} & & \textbf{0.208} & 0.000 \\
Stroke Color & 0.288 & \textbf{0.289} & & \textbf{0.281} & 0.000 \\
Border Style & 0.462 & \textbf{0.473} & & \textbf{0.465} & 0.000 \\
Position     & 0.470 & \textbf{0.478} & & \textbf{0.473} & 0.000 \\
Font         & 0.171 & \textbf{0.202} & & \textbf{0.227} & 0.000 \\
Aspect Ratio & 0.414 & \textbf{0.428} & & \textbf{0.423} & 0.000 \\
$R_S$        & 0.323 & \textbf{0.331} & & \textbf{0.332} & 0.000 \\
\midrule
\multicolumn{6}{l}{\itshape Arrow attributes} \\
\midrule
Source       & 0.676 & \textbf{0.682} & & \textbf{0.683} & 0.533 \\
Dest         & \textbf{0.665} & 0.663      & & \textbf{0.664} & 0.533 \\
Head         & \textbf{0.702} & 0.700      & & \textbf{0.714} & 0.533 \\
Head Size    & 0.595 & \textbf{0.654} & & \textbf{0.649} & 0.533 \\
Curve        & \textbf{0.730} & 0.727      & & \textbf{0.730} & 0.533 \\
Color        & \textbf{0.721} & 0.718      & & \textbf{0.723} & 0.533 \\
Overlap      & \textbf{0.737} & 0.732      & & \textbf{0.729} & 0.533 \\
$R_A$        & 0.637 & \textbf{0.654} & & \textbf{0.646} & 0.533 \\
\midrule
$R$          & 0.480 & \textbf{0.492} & & \textbf{0.489} & 0.266 \\
\bottomrule
\end{tabular}
\end{table}

\begin{table}[!ht]
\centering
\caption{Per-metric evaluation on \textbf{Molmo} (superset of $281$ samples).
  Bold indicates the best score within each group (open- and closed-source)
  per row.}
\label{tab:permetric_molmo}
\setlength{\tabcolsep}{3.5pt}
\small
\begin{tabular}{l cc c ccc}
\toprule
& \multicolumn{2}{c}{Closed-source} & & \multicolumn{3}{c}{Open-source} \\
\cmidrule(lr){2-3}\cmidrule(lr){5-7}
Metric & GPT-5.2 & Gem.-3-Pro & & RL & Qwen3-VL & OmniSVG \\
\midrule
\multicolumn{7}{l}{\itshape Shape attributes} \\
\midrule
Label        & \textbf{0.526} & 0.488 & & \textbf{0.507} & 0.475 & 0.000 \\
Type         & \textbf{0.489} & 0.465 & & \textbf{0.469} & 0.439 & 0.000 \\
Fill Color   & \textbf{0.428} & 0.397 & & \textbf{0.400} & 0.360 & 0.000 \\
Fill Style   & \textbf{0.413} & 0.379 & & \textbf{0.396} & 0.364 & 0.000 \\
Stroke Color & \textbf{0.404} & 0.364 & & \textbf{0.352} & 0.278 & 0.000 \\
Border Style & \textbf{0.533} & 0.502 & & \textbf{0.518} & 0.490 & 0.000 \\
Position     & \textbf{0.530} & 0.496 & & \textbf{0.514} & 0.473 & 0.000 \\
Font         & 0.169 & \textbf{0.207} & & \textbf{0.258} & 0.123 & 0.000 \\
Aspect Ratio & \textbf{0.492} & 0.464 & & \textbf{0.475} & 0.422 & 0.000 \\
$R_S$        & \textbf{0.416} & 0.393 & & \textbf{0.395} & 0.347 & 0.000 \\
\midrule
\multicolumn{7}{l}{\itshape Arrow attributes} \\
\midrule
Source       & \textbf{0.874} & 0.858 & & \textbf{0.842} & 0.817 & 0.737 \\
Dest         & \textbf{0.860} & 0.849 & & \textbf{0.826} & 0.814 & 0.737 \\
Head         & \textbf{0.862} & 0.854 & & \textbf{0.834} & 0.829 & 0.737 \\
Head Size    & 0.785 & \textbf{0.824} & & \textbf{0.808} & 0.775 & 0.737 \\
Curve        & \textbf{0.884} & 0.872 & & \textbf{0.854} & 0.818 & 0.737 \\
Color        & \textbf{0.880} & 0.863 & & \textbf{0.851} & 0.820 & 0.737 \\
Overlap      & \textbf{0.894} & 0.882 & & \textbf{0.860} & 0.830 & 0.737 \\
$R_A$        & \textbf{0.839} & 0.838 & & \textbf{0.819} & 0.795 & 0.737 \\
\midrule
$R$          & \textbf{0.627} & 0.616 & & \textbf{0.607} & 0.571 & 0.368 \\
\bottomrule
\end{tabular}
\end{table}

\begin{table}[!ht]
\centering
\caption{Per-metric evaluation on \textbf{\progds} (superset of $182$ samples).
  Bold indicates the best score within each group (open- and closed-source)
  per row.}
\label{tab:permetric_figdata}
\setlength{\tabcolsep}{3.5pt}
\small
\begin{tabular}{l cc c ccc}
\toprule
& \multicolumn{2}{c}{Closed-source} & & \multicolumn{3}{c}{Open-source} \\
\cmidrule(lr){2-3}\cmidrule(lr){5-7}
Metric & GPT-5.2 & Gem.-3-Pro & & RL & Qwen3-VL & OmniSVG \\
\midrule
\multicolumn{7}{l}{\itshape Shape attributes} \\
\midrule
Label        & \textbf{0.919} & 0.807 & & 0.881 & \textbf{0.908} & 0.000 \\
Type         & \textbf{0.787} & 0.710 & & 0.695 & \textbf{0.718} & 0.000 \\
Fill Color   & 0.594 & \textbf{0.626} & & 0.812 & \textbf{0.815} & 0.000 \\
Fill Style   & 0.515 & \textbf{0.569} & & \textbf{0.377} & 0.136 & 0.000 \\
Stroke Color & \textbf{0.714} & 0.692 & & \textbf{0.808} & 0.711 & 0.000 \\
Border Style & \textbf{0.884} & 0.816 & & \textbf{0.853} & 0.811 & 0.000 \\
Position     & \textbf{0.906} & 0.837 & & \textbf{0.917} & 0.878 & 0.000 \\
Font         & \textbf{0.266} & 0.241 & & \textbf{0.263} & 0.034 & 0.000 \\
Aspect Ratio & \textbf{0.864} & 0.784 & & \textbf{0.847} & 0.824 & 0.000 \\
$R_S$        & \textbf{0.708} & 0.661 & & \textbf{0.691} & 0.645 & 0.000 \\
\midrule
\multicolumn{7}{l}{\itshape Arrow attributes} \\
\midrule
Source       & \textbf{0.749} & 0.676 & & \textbf{0.652} & 0.642 & 0.000 \\
Dest         & \textbf{0.781} & 0.702 & & 0.666 & \textbf{0.685} & 0.000 \\
Head         & \textbf{0.785} & 0.742 & & \textbf{0.805} & 0.664 & 0.000 \\
Head Size    & 0.322 & \textbf{0.330} & & \textbf{0.519} & 0.271 & 0.000 \\
Curve        & 0.486 & \textbf{0.724} & & \textbf{0.711} & 0.526 & 0.000 \\
Color        & \textbf{0.791} & 0.769 & & \textbf{0.761} & 0.660 & 0.000 \\
Overlap      & 0.738 & \textbf{0.769} & & \textbf{0.686} & 0.664 & 0.000 \\
$R_A$        & 0.624 & \textbf{0.628} & & \textbf{0.600} & 0.524 & 0.000 \\
\midrule
$R$          & \textbf{0.666} & 0.645 & & \textbf{0.646} & 0.585 & 0.000 \\
\bottomrule
\end{tabular}
\end{table}

\section{Experiment Setup}
\label{appendix:setup}
This subsection provides implementation details for training and inference. 
We describe the training setup for both supervised fine-tuning (SFT) and reinforcement learning (RL), including model initialization, optimization settings, and compute resources. 
We also present the inference configuration and evaluation protocols used in our experiments.
\subsection{Training Setup}
We describe the training configurations for both supervised fine-tuning (SFT) and reinforcement learning (RL). 

\noindent\textbf{SFT Training Configuration.}
Table~\ref{tab:sft_config} summarizes the main hyperparameters used in SFT. 
We adopt Qwen3-VL-4B-Instruct as the backbone and perform parameter-efficient fine-tuning using LoRA with rank 64 applied to language model layers only, while keeping the vision encoder and projector frozen. The maximum sequence length is set to 8192 tokens, and input images are resized to at most $262{,}144$ pixels ($512\times512$). 
Training is conducted with a learning rate of $2\times10^{-5}$ using the cosine scheduler with a warmup ratio of 0.1. We use a per-device batch size of 1 with gradient accumulation of 16 and train for 3 epochs using bf16 precision.

Across different backbones, the wall-clock training time varies depending on model size and training strategy. For the single-stage SFT setup, training takes approximately 3d20h for Qwen3-VL-8B, 1d03h for Qwen2.5-VL-3B, and 1d14h for InternVL3.5-4B, while the Qwen3-VL-4B run was trained for approximately 5 days. For the simple-to-complex SFT strategy, the simple and complex stages are trained sequentially. Specifically, Qwen3-VL-8B requires 2d9h for the simple stage and 2d10h for the complex stage, Qwen2.5-VL-3B requires 23h and 24h respectively, and InternVL3.5-4B requires 1d07h and 22h. The Qwen3-VL-4B simple stage was trained for approximately 5 days, and the complex stage requires about 2d20h. All experiments are conducted on 5$\times$L40S GPUs unless otherwise specified; the Qwen3-VL-4B complex stage uses 8$\times$L40S GPUs.

\noindent\textbf{RL Training Configuration.}
We initialize RL training from the simple-to-complex SFT checkpoint of Qwen3-VL-4B-Instruct and optimize the policy using GRPO. Due to compute and time constraints, we conduct RL only on the Qwen3-VL-4B backbone. Similar to SFT, we apply LoRA with rank 64 to the language model layers while keeping the vision tower frozen. Table~\ref{tab:rl_config} summarizes the main hyperparameters used in RL training. We train on the combined RL dataset with a train batch size of 64 and validation batch size of 64, using up to 50 validation samples for periodic evaluation. The maximum prompt length and response length are set to 9000 and 8500 tokens, respectively, and the rollout engine uses a maximum model length of 17{,}500 tokens. We optimize the policy with a learning rate of $9 \times 10^{-6}$, cosine learning-rate decay, and a warmup ratio of 0.03 in bf16 precision. For policy optimization, we use a mini-batch size of 16 and a micro-batch size of 1 per GPU. During rollout, we sample $n = 8$ candidate responses for each prompt to compute group-relative advantages. We further apply a KL regularization term with coefficient 0.01 and an entropy bonus with coefficient 0.001 to stabilize optimization and encourage exploration. RL training is performed on 4$\times$L40S GPUs for approximately 30 hours. The reward is computed using our custom rubric-based reward function described in the main text.

\begin{table}[t]
\centering
\small
\caption{SFT training configuration}
\label{tab:sft_config}
\setlength{\tabcolsep}{5pt}
\resizebox{0.6\linewidth}{!}{
\begin{tabular}{lll}
\toprule
\textbf{Component} & \textbf{Parameter} & \textbf{Value} \\
\midrule
Backbone & Base model & Qwen3-VL-4B-Instruct \\
 & Finetuning method & LoRA \\
\midrule
LoRA & Rank & 64 \\
 & Target modules & LM only \\
\midrule
Input & Max sequence length & 8192 \\
 & Image max pixels & 262{,}144 \\
\midrule
Training & Learning rate & 2e-5 \\
 & Per-device batch size & 1 \\
 & Gradient accumulation & 16 \\
 & Epochs & 3 \\
 & LR scheduler & cosine \\
 & Warmup ratio & 0.1 \\
 & Precision & bf16 \\
\bottomrule
\end{tabular}
}
\end{table}

\begin{table}[t]
\centering
\small
\caption{RL training configuration}
\label{tab:rl_config}
\setlength{\tabcolsep}{5pt}
\resizebox{0.8\linewidth}{!}{
\begin{tabular}{lll}
\toprule
\textbf{Component} & \textbf{Parameter} & \textbf{Value} \\
\midrule
Backbone & Base model & Qwen3-VL-4B-Instruct \\
 & Initialization & Simple-to-complex SFT checkpoint \\
 & RL algorithm & GRPO \\
\midrule
LoRA & Rank & 64 \\
 & Alpha & 16 \\
 & Target modules & all-linear (excluding visual modules) \\
 & Vision tower & frozen \\
\midrule
Input / Rollout & Max prompt length & 9000 \\
 & Max response length & 8500 \\
 & Max model length & 17{,}500 \\
 & Rollout samples per prompt & 8 \\
\midrule
Training & Learning rate & 9e-6 \\
 & Train batch size & 64 \\
 & Validation batch size & 64 \\
 & Optimization mini-batch size & 16 \\
 & Micro-batch size / GPU & 1 \\
 & LR scheduler & cosine \\
 & Warmup ratio & 0.03 \\
 & KL loss coefficient & 0.01 \\
 & Entropy coefficient & 0.001 \\
 & Precision & bf16 \\
\midrule
System & GPUs & 4$\times$L40S \\
 & Training time & $\sim$30 hours \\
\bottomrule
\end{tabular}
}
\end{table}

\subsection{Inference Setup}
During inference, we generate SVG code with a maximum generation length of 8192 tokens. Given an input figure, the model receives the prompt \emph{“Convert this figure into valid SVG code.”} together with the image, and produces the SVG program autoregressively. For our model and baselines without prescribed decoding settings, we use deterministic greedy decoding (\texttt{do\_sample=False}). For OmniSVG-4B and Starvector-8B, we follow the default decoding parameters recommended in their official implementations. We use greedy decoding by default because SVG generation is a structured program synthesis task where small token-level errors can easily break the XML syntax or produce invalid code. Selecting the most probable token at each step generally improves structural stability compared to stochastic sampling. To ensure valid outputs, we further extract the content between the \texttt{<svg>} and \texttt{</svg>} tags from the generated text.

\subsection{Training Objectives and Reward Design}
\label{app:training_objectives}
This subsection provides the formal training objectives and full rubric reward details summarized in \cref{sec: method}.

\noindent\textbf{SFT objective.}
Given paired figure--SVG examples $\mathcal{D}=\{(x_i,y_i)\}_{i=1}^{N}$ where each input figure is rendered directly from its ground-truth SVG to ensure exact visual--structural alignment, we fine-tune a VLM with parameters $\theta$ to model the conditional distribution $p_\theta(y \mid x)$ by maximizing the data likelihood:
\begin{equation}
\mathcal{L}_{\text{SFT}}
=
-\mathbb{E}_{(x,y)\sim\mathcal{D}}
\left[\log p_\theta(y \mid x)\right].
\label{eq:sft}
\end{equation}
At training time we condition on a simple prompt (\eg, ``\texttt{Generate the SVG for this figure}'').

\noindent\textbf{GRPO details.}
The full GRPO objective with group-normalized advantages and KL regularization is given in Eq.~\ref{eq:grpo} of \cref{sec:rl_reward}. KL regularization against the frozen SFT reference $\pi_{\text{ref}}$ prevents reward hacking and maintains generation stability; programs that fail to produce valid renderings (\eg, due to syntax errors or rendering timeouts) receive zero reward.

\noindent\textbf{Rubric reward axes.}
The rubric VLM judge (Gemini-3-Flash) compares the rendered prediction against the ground-truth image and outputs four scores in $[0,1]$, which are averaged into the scalar reward $R$ (Eq.~\ref{eq:grpo}):
\begin{itemize}
  \item \textbf{Presence ($r_{\text{pres}}$).} Whether all required visual elements (shapes, arrows, labels) are present. Penalizes missing components and discourages incomplete diagram generation.
  \item \textbf{Layout ($r_{\text{layout}}$).} Spatial arrangement, alignment, and relative positioning. Diagram-centric figures require precise structural placement; this term ensures geometric correctness beyond mere element existence.
  \item \textbf{Connectivity ($r_{\text{conn}}$).} Whether arrows and lines connect the correct endpoints. Critical for preserving relational semantics (\eg, source--destination correctness), which pixel similarity metrics often fail to capture.
  \item \textbf{Details ($r_{\text{det}}$).} Text accuracy and fine styling attributes such as font, stroke, and color. Encourages preservation of fine-grained visual fidelity important for diagram readability.
\end{itemize}

\noindent\textbf{Pixel-level reward variant.}
For the Gemini+Pixel ablation, the scalar reward uniformly averages four signals: the full Gemini reward, image-level L2 similarity, Canny-edge-map L2 similarity, and SSIM. The full Gemini reward is the uniform average of presence, layout, connectivity, and details. For the Canny-edge-map L2 term, we first extract Canny edges from grayscale renderings, then dilate and Gaussian-blur the edge maps before computing L2 similarity, making the term less sensitive to one-pixel misalignment than raw image L2.

\noindent\textbf{Why a rubric judge over pixel metrics.}
Pixel-level metrics (\eg, L2, SSIM, LPIPS) can assign high similarity to outputs with structurally incorrect arrows or misplaced labels because the affected pixel area is small. A rubric-based VLM judge captures these structural and semantic differences and aligns better with human judgment. On 100 annotated examples, Gemini judge scores correlate with human ratings at Pearson $r{=}0.89$ overall, with per-axis correlations of $0.79$ (presence), $0.63$ (layout), $0.83$ (connectivity), and $0.87$ (details).

\section{Additional Ablations}
\label{appendix: ablation}
This section presents additional ablation studies for both SFT and RL training. 
Specifically, we analyze: 
(i) SFT design choices, including backbone selection, training curricula, and parameter-efficient fine-tuning configurations; and 
(ii) RL design choices, including reward decomposition and SFT initialization.
\subsection{SFT Ablation}

Tables~\ref{tab:sft_ablation_backbone}, \ref{tab:sft_ablation_curriculum}, \ref{tab:sft_ablation_all}, and~\ref{tab:sft_lora_lm} analyze the effect of backbone choice, training curriculum, and parameter-efficient fine-tuning configurations across diagram benchmarks. Tables~\ref{tab:sft_ablation_backbone} and~\ref{tab:sft_ablation_curriculum} report aggregated averages across the three benchmarks, complementing the focused 3-row ablation in the main paper (Table~\ref{tab:sft_ablation}); Table~\ref{tab:sft_ablation_all} provides the per-dataset breakdown.

\begin{table*}[t]
\centering

\begin{minipage}[t]{0.49\textwidth}
\centering
\small
\caption{\textbf{Backbone}. Qwen3-VL consistently outperforms the other backbones by large margins on visual metrics, with the 8B model slightly better than 4B across all metrics.}
\setlength{\tabcolsep}{4pt}
\renewcommand{\arraystretch}{1.05}

\resizebox{\linewidth}{!}{%
\begin{tabular}{lcccccc}
\multicolumn{7}{c}{\textbf{Average across \bench + Molmo2-Diagram + SVG-Diagram}} \\
\midrule
\textbf{Models} &
\textbf{SSIM}$\uparrow$ &
\textbf{LPIPS}$\downarrow$ &
\textbf{VisualSim}$\uparrow$ &
\textbf{VLM-Judge}$\uparrow$ &
\textbf{Clean}$\uparrow$ &
\textbf{Render}$\uparrow$ \\
\midrule
InternVL3.5-4B & 70.0 & 39.3 & 90.1 & 62.9 & \underline{76.9} & 77.8 \\
Qwen2.5-VL-4B & 69.3 & 44.1 & 88.3 & 52.7 & 76.0 & \underline{82.3} \\
Qwen3-VL-4B & \underline{74.9} & \underline{27.0} & \underline{92.5} & \underline{71.2} & 76.1 & 74.9 \\
Qwen3-VL-8B & {\textbf{75.0}} & {\textbf{26.1}} & {\textbf{93.8}} & {\textbf{74.6}} & {\textbf{77.6}} &{\textbf{85.9}} \\
\end{tabular}%
}

\label{tab:sft_ablation_backbone}
\end{minipage}%
\hfill
\begin{minipage}[t]{0.49\textwidth}
\centering
\caption{\textbf{Curriculum}. We note that the two-stage curriculum improves the render success rate by a lot over the one-stage one and slightly increases the VLM-judge score.}
\setlength{\tabcolsep}{4pt}
\renewcommand{\arraystretch}{1.05}

\resizebox{\linewidth}{!}{%
\begin{tabular}{llcccccc}
\multicolumn{8}{c}{\textbf{Average across \bench + Molmo2-Diagram + SVG-Diagram}} \\
\midrule
\textbf{Model} & \textbf{Curriculum} &
\textbf{SSIM}$\uparrow$ &
\textbf{LPIPS}$\downarrow$ &
\textbf{VisualSim}$\uparrow$ &
\textbf{VLM-Judge}$\uparrow$ &
\textbf{Clean}$\uparrow$ &
\textbf{Render}$\uparrow$ \\
\midrule
\multirow{2}{*}{Qwen3-VL-4B}
& one-stage & {\textbf{74.9}} & 27.0 & 92.5 & 71.2 & 76.1 & 74.9 \\
& two-stage & 72.7 & {\textbf{26.5}} & {\textbf{93.1}} & {\textbf{73.7}} & {\textbf{77.6}} & {\textbf{93.3}} \\
\midrule
\multirow{2}{*}{Qwen3-VL-8B}
& one-stage & {\textbf{75.0}} & 26.1 & {\textbf{93.8}} & 74.6 & {\textbf{77.6}} & 85.9 \\
& two-stage & 72.7 & {\textbf{25.8}} & 93.1 & {\textbf{75.0}} & 75.9 & {\textbf{94.5}} \\
\end{tabular}%
}

\label{tab:sft_ablation_curriculum}
\end{minipage}

\end{table*}

Table~\ref{tab:sft_ablation_all} provides the full per-dataset results for different SFT backbones and training curricula. While the main paper reports aggregated results, this table reveals several dataset-specific behaviors. From Table~\ref{tab:sft_ablation_all}, we observe a consistent trend that newer Qwen3-VL backbones outperform earlier VLMs across all datasets. In particular, Qwen3-VL achieves the strongest results on semantic metrics such as VisualSim and VLM-Judge, indicating improved alignment between generated SVG programs and diagram semantics. Increasing the model size from 4B to 8B provides additional gains, although the improvement is moderate across most metrics. Simple-to-complex SFT further improves structural reliability and semantic alignment on the more compositionally complex datasets, suggesting that separating primitive-heavy diagram pretraining from realistic figure fine-tuning helps the model better capture hierarchical diagram structure. Based on these observations, we adopt \textbf{Qwen3-VL-4B with simple-to-complex SFT} as the default backbone for subsequent RL experiments.

Table~\ref{tab:sft_lora_lm} examines parameter-efficient fine-tuning configurations on Qwen2.5-VL-3B. Increasing the LoRA rank generally improves semantic alignment metrics, with rank 64 achieving the best overall performance on \model{}-Bench and Molmo2-Diagram, particularly in VisualSim and VLM-judge. However, lower ranks remain competitive on certain datasets, indicating that moderate ranks already capture much of the task adaptation.

We further compare different SFT target modules while fixing the LoRA rank to 64. Adapting the language model alone consistently yields the strongest or most balanced results across datasets, outperforming or matching configurations that additionally tune the projector or vision encoder. These results suggest that most task-specific adaptation occurs in the language generation component responsible for producing structured SVG programs, while modifying the vision backbone or multimodal projector provides limited additional benefit in this setting.

\begin{table*}[t]
\centering
\small
\setlength{\tabcolsep}{3pt}
\renewcommand{\arraystretch}{1.05}
\caption{\textbf{Ablation study for SFT across three datasets.} VisualSim denotes the average cosine similarity of DINO, CLIP, and SigLIP embeddings. VLM-judge denotes the mean of Gemini and GPT judge score used in evaluation.}
\label{tab:sft_ablation_all}
\resizebox{\textwidth}{!}{
\begin{tabular}{lcccccc|cccccc|cccccc}
\toprule

& \multicolumn{6}{c|}{\textbf{\model{}-Bench}}
& \multicolumn{6}{c|}{\textbf{Molmo2-Diagram}}
& \multicolumn{6}{c}{\textbf{SVG-Diagram}} \\

\cmidrule(lr){2-7}
\cmidrule(lr){8-13}
\cmidrule(lr){14-19}

\textbf{Model}

& \rotatebox{60}{SSIM$\uparrow$}
& \rotatebox{60}{LPIPS$\downarrow$}
& \rotatebox{60}{VisualSim$\uparrow$}
& \rotatebox{60}{VLM-judge$\uparrow$}
& \rotatebox{60}{Clean$\uparrow$}
& \rotatebox{60}{Render$\uparrow$}

& \rotatebox{60}{SSIM$\uparrow$}
& \rotatebox{60}{LPIPS$\downarrow$}
& \rotatebox{60}{VisualSim$\uparrow$}
& \rotatebox{60}{VLM-judge$\uparrow$}
& \rotatebox{60}{Clean$\uparrow$}
& \rotatebox{60}{Render$\uparrow$}

& \rotatebox{60}{SSIM$\uparrow$}
& \rotatebox{60}{LPIPS$\downarrow$}
& \rotatebox{60}{VisualSim$\uparrow$}
& \rotatebox{60}{VLM-judge$\uparrow$}
& \rotatebox{60}{Clean$\uparrow$}
& \rotatebox{60}{Render$\uparrow$}

\\
\midrule

\multicolumn{19}{l}{\textit{\textcolor{gray}{Open-source VLMs}}} \\

InternVL3.5-4B (1-stage) & 0.707 & 0.438 & 0.910 & 0.609 & 0.755 & 0.821 & 0.731 & 0.361 & 0.911 & 0.685 & 0.828 & 0.944 & 0.629 & 0.396 & 0.870 & 0.545 & 0.704 & 0.550 \\
InternVL3.5-4B (2-stage) & 0.707 & 0.433 & 0.908 & 0.594 & 0.783 & 0.718 & 0.711 & 0.404 & 0.900 & 0.648 & 0.820 & 0.902 & 0.624 & 0.392 & 0.867 & 0.549 & 0.731 & 0.750 \\
Qwen2.5-VL-3B (1-stage) & 0.703 & 0.492 & 0.901 & 0.537 & 0.794 & 0.879 & 0.752 & 0.368 & 0.908 & 0.611 & 0.847 & 0.962 & 0.577 & 0.503 & 0.815 & 0.365 & 0.615 & 0.616 \\
Qwen2.5-VL-3B (2-stage) & 0.715 & 0.466 & 0.910 & 0.580 & \cellcolor{green!20}{\textbf{0.811}} & \cellcolor{green!20}{\textbf{0.922}} & 0.735 & 0.409 & 0.891 & 0.578 & 0.832 & 0.940 & 0.576 & 0.485 & 0.834 & 0.429 & \cellcolor{green!20}{\textbf{0.736}} & 0.914 \\
Qwen3-VL-4B (1-stage) & 0.746 & 0.296 & 0.939 & 0.738 & 0.747 & 0.831 & 0.795 & 0.216 & 0.937 & 0.770 & 0.836 & 0.940 & \cellcolor{green!20}{\textbf{0.653}} & 0.352 & 0.876 & 0.534 & 0.655 & 0.457 \\
Qwen3-VL-4B (2-stage) & \cellcolor{green!20}{\textbf{0.763}} & 0.264 & 0.951 & 0.781 & 0.784 & 0.884 & 0.783 & 0.226 & 0.937 & 0.776 & 0.828 & 0.966 & 0.633 & 0.311 & \cellcolor{green!20}{\textbf{0.907}} & 0.653 & 0.709 & 0.939 \\
Qwen3-VL-8B (1-stage) & 0.748 & 0.302 & 0.946 & 0.760 & 0.787 & 0.884 & \cellcolor{green!20}{\textbf{0.815}} & \cellcolor{green!20}{\textbf{0.187}} & \cellcolor{green!20}{\textbf{0.952}} & \cellcolor{green!20}{\textbf{0.817}} & \cellcolor{green!20}{\textbf{0.850}} & \cellcolor{green!20}{\textbf{0.978}} & 0.649 & 0.331 & 0.906 & 0.617 & 0.670 & 0.702 \\
Qwen3-VL-8B (2-stage) & \cellcolor{green!20}{\textbf{0.763}} & \cellcolor{green!20}{\textbf{0.260}} & \cellcolor{green!20}{\textbf{0.954}} & \cellcolor{green!20}{\textbf{0.806}} & 0.758 & 0.907 & 0.783 & 0.220 & 0.941 & 0.790 & 0.816 & 0.964 & 0.633 & \cellcolor{green!20}{\textbf{0.301}} & 0.901 & \cellcolor{green!20}{\textbf{0.657}} & 0.696 & \cellcolor{green!20}{\textbf{0.959}} \\
\bottomrule
\end{tabular}
}
\end{table*}

\begin{table*}[t]
\centering
\small
\setlength{\tabcolsep}{3pt}
\renewcommand{\arraystretch}{1.05}
\caption{\textbf{Ablation of LoRA rank and SFT target modules on Qwen2.5-VL-3B (1-stage).}}
\label{tab:sft_lora_lm}
\resizebox{\textwidth}{!}{
\begin{tabular}{lcccccc|cccccc|cccccc}
\toprule

& \multicolumn{6}{c|}{\textbf{\model{}-Bench}}
& \multicolumn{6}{c|}{\textbf{Molmo2-Diagram}}
& \multicolumn{6}{c}{\textbf{SVG-Diagram}} \\

\cmidrule(lr){2-7}
\cmidrule(lr){8-13}
\cmidrule(lr){14-19}

\textbf{Model}

& \rotatebox{60}{SSIM$\uparrow$}
& \rotatebox{60}{LPIPS$\downarrow$}
& \rotatebox{60}{VisualSim$\uparrow$}
& \rotatebox{60}{VLM-judge$\uparrow$}
& \rotatebox{60}{Clean$\uparrow$}
& \rotatebox{60}{Render$\uparrow$}

& \rotatebox{60}{SSIM$\uparrow$}
& \rotatebox{60}{LPIPS$\downarrow$}
& \rotatebox{60}{VisualSim$\uparrow$}
& \rotatebox{60}{VLM-judge$\uparrow$}
& \rotatebox{60}{Clean$\uparrow$}
& \rotatebox{60}{Render$\uparrow$}

& \rotatebox{60}{SSIM$\uparrow$}
& \rotatebox{60}{LPIPS$\downarrow$}
& \rotatebox{60}{VisualSim$\uparrow$}
& \rotatebox{60}{VLM-judge$\uparrow$}
& \rotatebox{60}{Clean$\uparrow$}
& \rotatebox{60}{Render$\uparrow$}

\\
\midrule

\multicolumn{19}{l}{\textit{\textcolor{gray}{LoRA Rank Ablation (Qwen2.5-VL-3B, 1-stage, LM-only SFT)}}} \\
LoRA rank = 16
& \cellcolor{green!20}{\textbf{0.708}} & 0.500 & 0.891 & 0.486 & 0.764 & 0.811
& 0.745 & 0.401 & 0.893 & 0.556 & 0.845 & 0.904
& \cellcolor{green!20}{\textbf{0.586}} & 0.512 & 0.813 & \cellcolor{green!20}{\textbf{0.367}} & \cellcolor{green!20}{\textbf{0.694}} & 0.720 \\
LoRA rank = 32
& 0.705 & 0.500 & 0.888 & 0.502 & \cellcolor{green!20}{\textbf{0.811}} & 0.864
& 0.750 & 0.389 & 0.896 & 0.572 & \cellcolor{green!20}{\textbf{0.861}} & 0.950
& \cellcolor{green!20}{\textbf{0.586}} & 0.509 & \cellcolor{green!20}{\textbf{0.816}} & 0.360 & 0.685 & \cellcolor{green!20}{\textbf{0.727}} \\
LoRA rank = 64
& 0.703 & \cellcolor{green!20}{\textbf{0.492}} & \cellcolor{green!20}{\textbf{0.901}} & \cellcolor{green!20}{\textbf{0.537}} & 0.794 & \cellcolor{green!20}{\textbf{0.879}}
& \cellcolor{green!20}{\textbf{0.752}} & \cellcolor{green!20}{\textbf{0.368}} & \cellcolor{green!20}{\textbf{0.908}} & \cellcolor{green!20}{\textbf{0.611}} & 0.847 & \cellcolor{green!20}{\textbf{0.962}}
& 0.577 & \cellcolor{green!20}{\textbf{0.503}} & 0.815 & 0.365 & 0.615 & 0.616 \\

\midrule

\multicolumn{19}{l}{\textit{\textcolor{gray}{SFT Target Module Ablation (Qwen2.5-VL-3B, 1-stage, LoRA rank = 64)}}} \\
LM + projector + vision encoder
& 0.702 & \cellcolor{green!20}{\textbf{0.483}} & 0.884 & 0.478 & 0.765 & 0.700
& \cellcolor{green!20}{\textbf{0.765}} & \cellcolor{green!20}{\textbf{0.354}} & 0.902 & 0.579 & 0.850 & 0.826
& 0.586 & \cellcolor{green!20}{\textbf{0.501}} & 0.815 & 0.365 & \cellcolor{green!20}{\textbf{0.643}} & 0.564 \\
LM + projector
& \cellcolor{green!20}{\textbf{0.704}} & 0.508 & 0.881 & 0.478 & 0.774 & 0.849
& 0.745 & 0.399 & 0.894 & 0.562 & \cellcolor{green!20}{\textbf{0.854}} & 0.934
& \cellcolor{green!20}{\textbf{0.592}} & 0.530 & 0.801 & 0.339 & 0.594 & \cellcolor{green!20}{\textbf{0.714}} \\
LM only
& 0.703 & 0.492 & \cellcolor{green!20}{\textbf{0.901}} & \cellcolor{green!20}{\textbf{0.537}} & \cellcolor{green!20}{\textbf{0.794}} & \cellcolor{green!20}{\textbf{0.879}}
& 0.752 & 0.368 & \cellcolor{green!20}{\textbf{0.908}} & \cellcolor{green!20}{\textbf{0.611}} & 0.847 & \cellcolor{green!20}{\textbf{0.962}}
& 0.577 & 0.503 & \cellcolor{green!20}{\textbf{0.815}} & \cellcolor{green!20}{\textbf{0.365}} & 0.615 & 0.616 \\
\bottomrule
\end{tabular}
}
\end{table*}

\subsection{RL Ablation}
Tables~\ref{tab:rl_ablation_all} and~\ref{tab:rl_init_ablation_all} provide additional ablations for RL, covering both reward design and SFT initialization. Table~\ref{tab:rl_ablation_all} reports the full per-dataset breakdown of the reward ablation results. Consistent with the main paper, the full reward achieves the strongest judge-based performance across datasets, confirming the benefit of jointly optimizing \emph{presence}, \emph{layout}, \emph{connectivity}, and \emph{details}. The per-dataset view further shows that the relative importance of reward components varies across benchmarks. On \model{}-Bench and SVG-Diagram, removing \emph{layout} or \emph{details} leads to clearer drops in VLM-Judge, supporting the intuition that these datasets require stronger spatial reasoning and fine-grained structural fidelity. Removing \emph{connectivity} also hurts performance, although the degradation is generally smaller than removing \emph{layout} or \emph{details}. On Molmo2-Diagram, the differences across reward variants are more moderate, suggesting that this benchmark is comparatively more regular and less sensitive to reward decomposition. We also observe the same trade-off as in the main paper: adding pixel-level objectives improves SSIM/LPIPS on some datasets, but does not translate into better judge-based scores, indicating that pixel reconstruction is not always aligned with structural correctness. The \emph{Hard Data} row (Tab.~\ref{tab:rl_ablation_all}) restricts RL to a curated subset of difficult samples; this does not improve performance over the Full reward across any of the three datasets, suggesting that diverse training data benefits RL stability more than focusing on hard examples.

Table~\ref{tab:rl_init_ablation_all} further analyzes two factors under the same RL framework: the choice of SFT initialization and the backbone model size. In the initialization ablation, we compare RL starting from single-stage and simple-to-complex SFT checkpoints of Qwen3-VL-4B. Overall, both initializations benefit from RL, while the simple-to-complex SFT initialization yields stronger semantic and structural performance on \model{}-Bench and SVG-Diagram, achieving higher SSIM, VisualSim, and VLM-Judge, together with lower LPIPS. This suggests that the simple-to-complex training curriculum provides a better starting point for subsequent reward optimization, especially on benchmarks with more complex layout and element relationships. On Molmo2-Diagram, the pattern is more mixed: the simple-to-complex initialization still achieves slightly stronger VisualSim and VLM-Judge, whereas the single-stage initialization attains better SSIM, LPIPS, Clean, and Render. This indicates that RL initialized from the single-stage checkpoint can still produce competitive visual fidelity and code quality on Molmo2-Diagram, but the simple-to-complex initialization remains preferable when prioritizing semantic alignment and judge-based structural quality.

The model size ablation compares Qwen3-VL-4B and Qwen3-VL-8B under the same simple-to-complex SFT initialization. Increasing the backbone size to 8B consistently improves VisualSim and VLM-Judge across all three datasets, and also gives the best SSIM and LPIPS on Molmo2-Diagram and SVG-Diagram. These gains suggest that a larger backbone improves semantic fidelity and structural alignment under RL. At the same time, the 4B model remains competitive on several code-level metrics: it attains higher Clean scores on all three datasets, matches the 8B model in Render on \model{}-Bench, and achieves a slightly higher Render score on SVG-Diagram. On \model{}-Bench, the 4B model also retains slightly better SSIM and LPIPS. Taken together, these results indicate a trade-off between perceptual or semantic quality and code cleanliness: the 8B model is generally stronger when optimizing for visual and judge-based quality, while the 4B model remains a more efficient choice and can produce cleaner SVG code. This supports our use of \textbf{Qwen3-VL-4B with simple-to-complex SFT} as the default RL setting in the main paper, as it provides a favorable balance between structural fidelity, code quality, and computational efficiency.

\begin{table*}[t]
\centering
\small
\setlength{\tabcolsep}{3pt}
\renewcommand{\arraystretch}{1.05}
\caption{\textbf{Per-dataset RL reward ablation across three datasets.} While the main paper reports results averaged across all benchmarks, this table provides the full per-dataset breakdown.}
\label{tab:rl_ablation_all}
\resizebox{\textwidth}{!}{
\begin{tabular}{lcccccc|cccccc|cccccc}
\toprule

& \multicolumn{6}{c|}{\textbf{\model{}-Bench}}
& \multicolumn{6}{c|}{\textbf{Molmo2-Diagram}}
& \multicolumn{6}{c}{\textbf{SVG-Diagram}} \\

\cmidrule(lr){2-7}
\cmidrule(lr){8-13}
\cmidrule(lr){14-19}

\textbf{Reward}

& \rotatebox{60}{SSIM$\uparrow$}
& \rotatebox{60}{LPIPS$\downarrow$}
& \rotatebox{60}{VisualSim$\uparrow$}
& \rotatebox{60}{VLM-judge$\uparrow$}
& \rotatebox{60}{Clean$\uparrow$}
& \rotatebox{60}{Render$\uparrow$}

& \rotatebox{60}{SSIM$\uparrow$}
& \rotatebox{60}{LPIPS$\downarrow$}
& \rotatebox{60}{VisualSim$\uparrow$}
& \rotatebox{60}{VLM-judge$\uparrow$}
& \rotatebox{60}{Clean$\uparrow$}
& \rotatebox{60}{Render$\uparrow$}

& \rotatebox{60}{SSIM$\uparrow$}
& \rotatebox{60}{LPIPS$\downarrow$}
& \rotatebox{60}{VisualSim$\uparrow$}
& \rotatebox{60}{VLM-judge$\uparrow$}
& \rotatebox{60}{Clean$\uparrow$}
& \rotatebox{60}{Render$\uparrow$}
\\
\midrule

Full Reward
& 0.776 & 0.216 & 0.956 & \cellcolor{green!20}{\textbf{0.828}} & \cellcolor{green!20}{\textbf{0.856}} & \cellcolor{green!20}{\textbf{0.952}}
& 0.806 & 0.163 & \cellcolor{green!20}{\textbf{0.953}} & \cellcolor{green!20}{\textbf{0.850}} & \cellcolor{green!20}{\textbf{0.865}} & \cellcolor{green!20}{\textbf{0.996}}
& \cellcolor{green!20}{\textbf{0.654}} & \cellcolor{green!20}{\textbf{0.253}} & \cellcolor{green!20}{\textbf{0.921}} & \cellcolor{green!20}{\textbf{0.728}} & \cellcolor{green!20}{\textbf{0.805}} & \cellcolor{green!20}{\textbf{0.980}} \\
No Presence
& 0.781 & 0.214 & \cellcolor{green!20}{\textbf{0.957}} & 0.824 & 0.840 & 0.892
& 0.804 & 0.169 & 0.952 & 0.844 & 0.850 & 0.978
& 0.644 & 0.266 & 0.919 & 0.704 & 0.797 & 0.975 \\
No Connectivity
& 0.780 & 0.217 & \cellcolor{green!20}{\textbf{0.957}} & 0.825 & 0.856 & 0.937
& 0.804 & 0.168 & 0.952 & 0.844 & 0.857 & 0.988
& 0.648 & 0.265 & \cellcolor{green!20}{\textbf{0.921}} & 0.712 & 0.803 & 0.973 \\
No Layout
& 0.775 & 0.223 & 0.955 & 0.820 & 0.848 & 0.907
& 0.803 & 0.175 & 0.950 & 0.839 & 0.852 & 0.978
& 0.645 & 0.270 & 0.917 & 0.697 & 0.794 & 0.977 \\
No Details
& 0.777 & 0.223 & 0.954 & 0.813 & 0.845 & 0.937
& 0.804 & 0.171 & 0.951 & 0.844 & 0.856 & 0.976
& 0.647 & 0.276 & 0.918 & 0.703 & 0.797 & 0.966 \\
Gemini + Pixel
& \cellcolor{green!20}{\textbf{0.785}} & \cellcolor{green!20}{\textbf{0.211}} & 0.955 & 0.802 & 0.835 & 0.917
& \cellcolor{green!20}{\textbf{0.812}} & \cellcolor{green!20}{\textbf{0.161}} & 0.952 & 0.836 & 0.842 & 0.982
& 0.653 & 0.261 & 0.920 & 0.692 & 0.771 & 0.966 \\
Hard Data
& 0.777 & 0.226 & 0.954 & 0.804 & 0.844 & 0.929
& 0.802 & 0.176 & 0.949 & 0.834 & 0.856 & 0.990
& 0.640 & 0.283 & 0.915 & 0.682 & 0.778 & 0.968 \\
\bottomrule
\end{tabular}
}
\end{table*}

\begin{table*}[t]
\centering
\small
\setlength{\tabcolsep}{3pt}
\renewcommand{\arraystretch}{1.05}
\caption{\textbf{Additional ablation study for RL across three datasets.} We analyze two factors under the same RL training framework: the effect of SFT initialization and the effect of backbone model size.}
\label{tab:rl_init_ablation_all}
\resizebox{\textwidth}{!}{
\begin{tabular}{lcccccc|cccccc|cccccc}
\toprule

& \multicolumn{6}{c|}{\textbf{\model{}-Bench}}
& \multicolumn{6}{c|}{\textbf{Molmo2-Diagram}}
& \multicolumn{6}{c}{\textbf{SVG-Diagram}} \\

\cmidrule(lr){2-7}
\cmidrule(lr){8-13}
\cmidrule(lr){14-19}

\textbf{Model}

& \rotatebox{60}{SSIM$\uparrow$}
& \rotatebox{60}{LPIPS$\downarrow$}
& \rotatebox{60}{VisualSim$\uparrow$}
& \rotatebox{60}{VLM-Judge$\uparrow$}
& \rotatebox{60}{Clean$\uparrow$}
& \rotatebox{60}{Render$\uparrow$}

& \rotatebox{60}{SSIM$\uparrow$}
& \rotatebox{60}{LPIPS$\downarrow$}
& \rotatebox{60}{VisualSim$\uparrow$}
& \rotatebox{60}{VLM-Judge$\uparrow$}
& \rotatebox{60}{Clean$\uparrow$}
& \rotatebox{60}{Render$\uparrow$}

& \rotatebox{60}{SSIM$\uparrow$}
& \rotatebox{60}{LPIPS$\downarrow$}
& \rotatebox{60}{VisualSim$\uparrow$}
& \rotatebox{60}{VLM-Judge$\uparrow$}
& \rotatebox{60}{Clean$\uparrow$}
& \rotatebox{60}{Render$\uparrow$}
\\
\midrule

\multicolumn{19}{l}{\textit{\textcolor{gray}{Initialization Ablation (Qwen3-VL-4B, RL from different SFT checkpoints)}}} \\

Qwen3-VL-4B (1-stage SFT) + RL
& 0.768 & 0.242 & 0.950 & 0.793 & 0.843 & 0.940
& \cellcolor{green!20}{\textbf{0.807}} & \cellcolor{green!20}{\textbf{0.175}} & 0.948 & 0.817 & \cellcolor{green!20}{\textbf{0.874}} & \cellcolor{green!20}{\textbf{0.994}}
& 0.644 & 0.275 & 0.915 & 0.690 & \cellcolor{green!20}{\textbf{0.796}} & 0.970 \\

Qwen3-VL-4B (2-stage SFT) + RL
& \cellcolor{green!20}{\textbf{0.778}} & \cellcolor{green!20}{\textbf{0.212}} & \cellcolor{green!20}{\textbf{0.957}} & \cellcolor{green!20}{\textbf{0.829}} & \cellcolor{green!20}{\textbf{0.853}} & \cellcolor{green!20}{\textbf{0.960}}
& 0.800 & 0.177 & \cellcolor{green!20}{\textbf{0.949}} & \cellcolor{green!20}{\textbf{0.834}} & 0.855 & 0.976
& \cellcolor{green!20}{\textbf{0.654}} & \cellcolor{green!20}{\textbf{0.267}} & \cellcolor{green!20}{\textbf{0.919}} & \cellcolor{green!20}{\textbf{0.705}} & 0.788 & \cellcolor{green!20}{\textbf{0.973}} \\

\midrule
\multicolumn{19}{l}{\textit{\textcolor{gray}{Model Size Ablation (2-stage SFT initialization + RL)}}} \\

Qwen3-VL-4B (2-stage SFT) + RL
& \cellcolor{green!20}{\textbf{0.778}} & \cellcolor{green!20}{\textbf{0.212}} & 0.957 & 0.829 & \cellcolor{green!20}{\textbf{0.853}} & \cellcolor{green!20}{\textbf{0.960}}
& 0.800 & 0.177 & 0.949 & 0.834 & \cellcolor{green!20}{\textbf{0.855}} & 0.976
& 0.654 & 0.267 & 0.919 & 0.705 & \cellcolor{green!20}{\textbf{0.788}} & \cellcolor{green!20}{\textbf{0.973}} \\

Qwen3-VL-8B (2-stage SFT) + RL
& 0.774 & 0.223 & \cellcolor{green!20}{\textbf{0.960}} & \cellcolor{green!20}{\textbf{0.845}} & 0.819 & \cellcolor{green!20}{\textbf{0.960}}
& \cellcolor{green!20}{\textbf{0.803}} & \cellcolor{green!20}{\textbf{0.173}} & \cellcolor{green!20}{\textbf{0.952}} & \cellcolor{green!20}{\textbf{0.851}} & 0.842 & \cellcolor{green!20}{\textbf{0.994}}
& \cellcolor{green!20}{\textbf{0.660}} & \cellcolor{green!20}{\textbf{0.251}} & \cellcolor{green!20}{\textbf{0.924}} & \cellcolor{green!20}{\textbf{0.729}} & 0.749 & 0.970 \\

\bottomrule
\end{tabular}
}
\end{table*}

\section{Human Evaluation}
\phantomsection
\label{appendix:human_evaluations}

To complement the \textbf{\bench} results in Sec.~\ref{sec: experiments}, we additionally conduct a human evaluation of figure-to-SVG generation quality. While the main paper reports automatic metrics, including rubric-based VLM judging, direct human comparison offers a valuable perspective on how generation quality is perceived by human evaluators.

\noindent\textbf{Evaluation setup.}
We compare seven models: Gemini 3 Pro, GPT-5.2, Claude Sonnet 4.6, \model-RL, \model-SFT, StarVector, and Qwen3-VL-4B. We run inference on \benchood{}. These images are excluded from the \dataset{} training set. Five authors served as human annotators. For each example, annotators were shown outputs from a pair of models and assigned one of four labels: first model better, second model better, both good, or both bad. In total, we collected 4{,}158 paired annotations across 21 model pairs, with 198 annotations per model pairs.

\begin{figure}[h]
    \centering
    \includegraphics[width=0.85\linewidth]{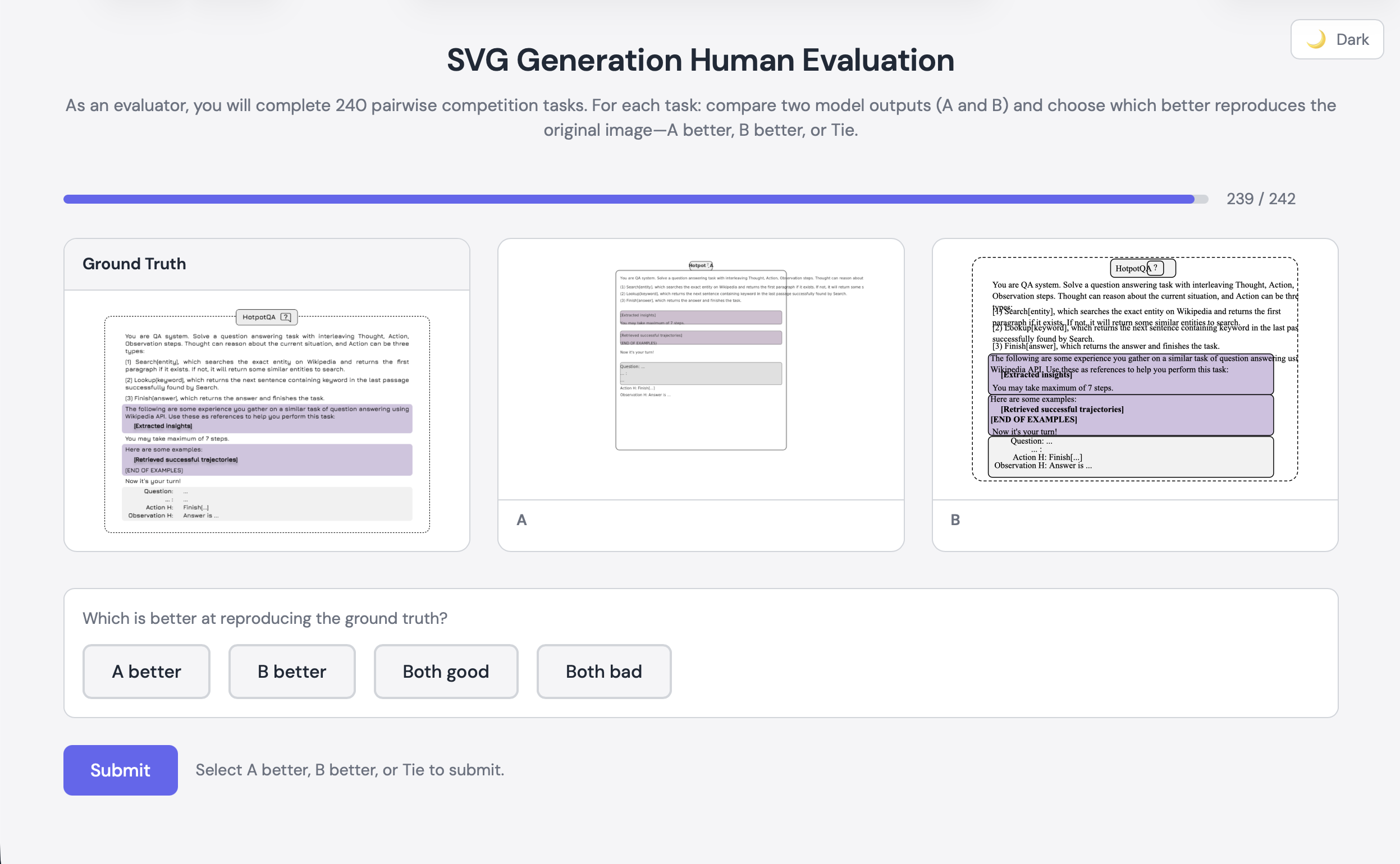}
    \caption{Screenshot of the human evaluation interface. Annotators are presented 
    with a ground-truth figure and two anonymous SVG reconstructions side by side, 
    and are asked to select which reconstruction more faithfully reproduces the original.}
    \label{fig:human_eval_interface}
\end{figure}

We evaluate model outputs via pairwise human comparison. For each trial, an annotator is presented with a ground-truth figure alongside two SVG reconstructions produced by two randomly sampled models. The annotator is asked to judge which reconstruction more faithfully reproduces the original figure. Figure~\ref{fig:human_eval_interface} illustrates the evaluation interface used in each trial. The available choices are:
\begin{itemize}
    \item \textbf{A is better}
    \item \textbf{B is better}
    \item \textbf{Both are good} (both reconstructions are acceptable)
    \item \textbf{Both are bad} (neither reconstruction is acceptable)
\end{itemize}
This protocol captures not only relative preference between the two outputs, but also cases where both outputs are acceptable or both fail to faithfully reproduce the target figure. Comparisons were conducted by the authors, and model identities were hidden from annotators during evaluation.

\noindent\textbf{Metrics.}
From the annotation results, we report three complementary summaries:
(1) \textbf{bootstrap-averaged Elo ratings}, where \textit{both good} and \textit{both bad} outcomes are treated as ties;
(2) a \textbf{quality-aware summary} reporting win rate, loss rate, both-good rate, both-bad rate, good rate (Win + Both Good), and decisive win rate (Win / (Win + Loss)); and
(3) a \textbf{pairwise comparison table} reporting head-to-head outcomes for each model pair.
Together, these metrics provide both a global ranking and a fine-grained view of human-perceived figure fidelity.

\noindent\textbf{Results.}
Results are shown in Tables~\ref{tab:human_eval_elo}, \ref{tab:human_eval_summary}, and~\ref{tab:human_eval_pairwise}.
Gemini-3-Pro achieves the strongest overall human preference, followed by GPT-5.2. Among the remaining models, \model (SFT+RL) ranks third in Elo, remains competitive with Claude-Sonnet-4.6, and clearly outperforms the other open-source baselines, including \model (SFT), Qwen3-VL-4B, and StarVector.

In the Elo ranking (Table~\ref{tab:human_eval_elo}), \model(SFT+RL) obtains an Elo score of 1577.5, slightly above Claude-Sonnet-4.6 at 1562.4 and above \model (SFT) at 1530.4. It substantially surpasses Qwen3-VL-4B and StarVector, which score 1212.8 and 942.3 respectively, though it remains below GPT-5.2 and Gemini-3-Pro.

In the quality-aware summary (Table~\ref{tab:human_eval_summary}), \model (SFT+RL) achieves a 46.0\% win rate and 55.5\% good rate. This improves over \model (SFT), which obtains a 42.3\% win rate and 50.7\% good rate, and is far above Qwen3-VL-4B and StarVector, which obtain good rates of 17.7\% and 3.1\%, respectively. Claude-Sonnet-4.6 achieves a slightly higher win rate than \model (SFT+RL), but a similar good rate, indicating that \model (SFT+RL) is competitive with this stronger proprietary baseline.

Pairwise comparisons (Table~\ref{tab:human_eval_pairwise}) further support this trend.
Against Qwen3-VL-4B, \model (SFT+RL) wins 81.6\% of trials versus 2.0\%, and against StarVector, it wins 91.4\% versus 2.5\%, indicating a clear human preference for \model (SFT+RL) over open-source baselines.
Compared with \model (SFT), \model (SFT+RL) wins 39.9\% of trials versus 25.3\%, with a 61.2\% decisive win rate after excluding ties, showing that reinforcement learning improves human-perceived reconstruction quality.
Against Claude-Sonnet-4.6, \model (SFT+RL) obtains an even 41.9\%--41.9\% split in raw wins, suggesting comparable human preference between the two models.
Against GPT-5.2, \model (SFT+RL) wins 17.4\% of trials versus 53.2\%, with a 16.9\% both-good rate, suggesting that \model (SFT+RL) produces competitive reconstructions on a non-trivial subset of examples.

\begin{figure}[h]
    \centering
    \includegraphics[width=0.72\linewidth]{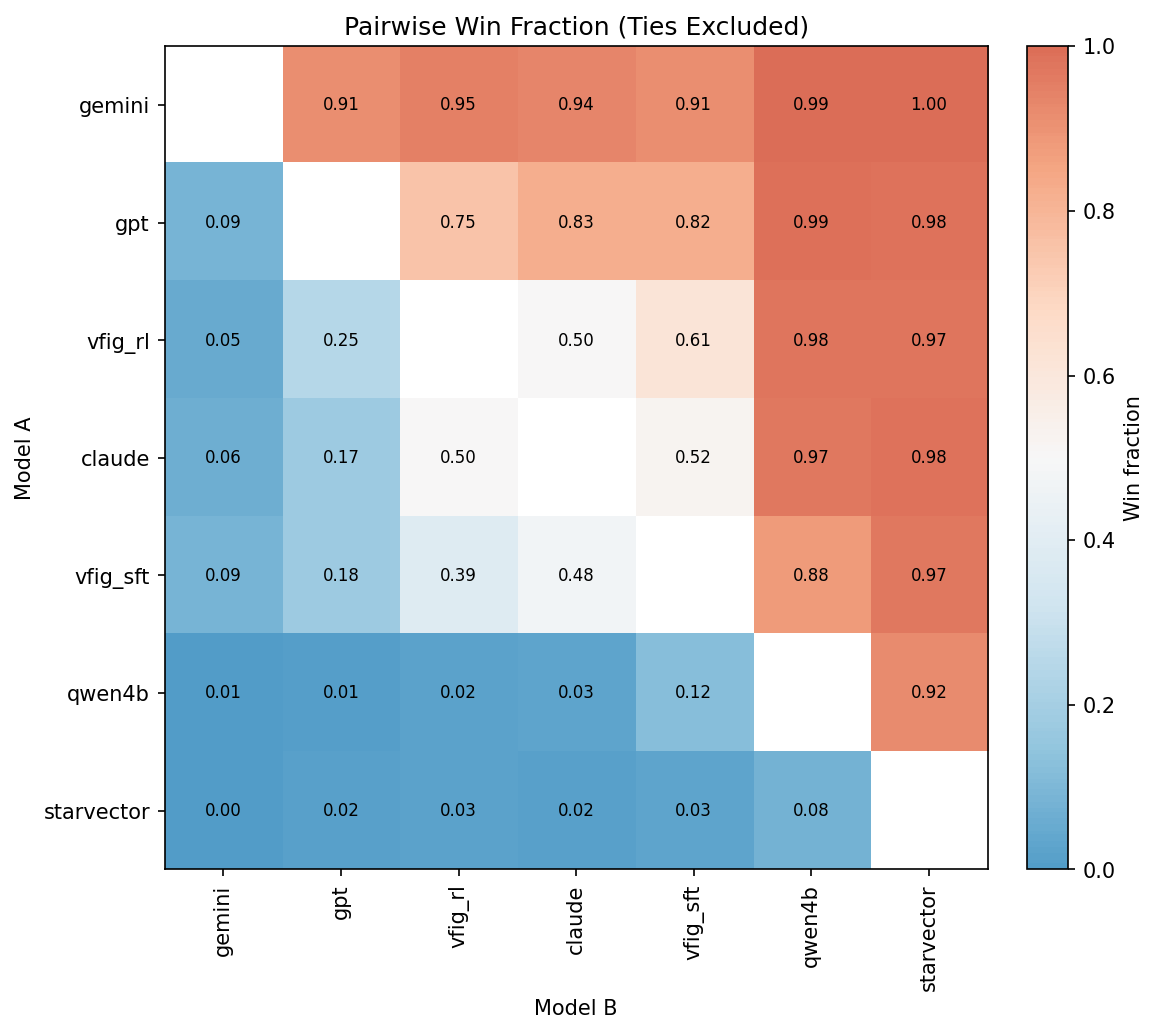}
    \caption{Pairwise win fractions in human evaluation with ties excluded. Each cell $(A, B)$ reports the fraction of decisive comparisons in which model $A$ is preferred over model $B$, excluding ``both good'' and ``both bad'' cases. Warmer colors indicate stronger pairwise preference. Gemini-3-Pro and GPT-5.2 achieve the strongest overall preferences, while \model (SFT+RL) consistently outperforms Qwen3-VL-4B and StarVector, improves over \model (SFT), and remains competitive with Claude-Sonnet-4.6.}
    \label{fig:human_eval_pairwise_heatmap}
\end{figure}

These trends are consistent with the benchmark results reported in the main paper on \bench.
Overall, the human evaluation confirms that \model (SFT+RL) improves over \model (SFT), substantially outperforms other open-source baselines, and reaches a level competitive with Claude-Sonnet-4.6, while leaving room for improvement compared with the strongest proprietary systems.
Figure~\ref{fig:human_eval_pairwise_heatmap} visualizes decisive pairwise win fractions with ties excluded.

\begin{table*}[t]
\centering
\caption{Elo ranking from human evaluation. We report the bootstrap mean, standard deviation, confidence interval, and median Elo score. Higher is better.}
\label{tab:human_eval_elo}
\resizebox{\textwidth}{!}{%
\begin{tabular}{lccccc}
\toprule
Model & Elo Mean & Elo Std & 95\% CI Lower & 95\% CI Upper & Elo Median \\
\midrule
Gemini-3-Pro        & \textbf{1942.0} & 39.7 & \textbf{1860.7} & \textbf{2018.8} & \textbf{1942.9} \\
GPT-5.2             & 1732.8 & 40.3 & 1654.6 & 1810.3 & 1733.6 \\
\model (SFT+RL)           & 1577.5 & 44.7 & 1492.1 & 1663.8 & 1577.2 \\
Claude-Sonnet-4.6   & 1562.4 & 44.2 & 1473.3 & 1654.3 & 1563.8 \\
\model (SFT)          & 1530.4 & 45.1 & 1444.4 & 1624.1 & 1531.1 \\
Qwen3-VL-4B         & 1212.8 & 37.2 & 1141.6 & 1285.4 & 1212.2 \\
StarVector          & 942.3  & 46.8 & 852.9  & 1039.6 & 941.6 \\
\bottomrule
\end{tabular}%
}
\end{table*}

\begin{table*}[t]
\centering
\caption{Quality-aware summary of human evaluation results. ``Good Rate'' is defined as 
\textit{Win + Both Good}. ``Decisive Win Rate'' is computed as 
$\text{Win} / (\text{Win} + \text{Loss})$. Higher is better for Win, Both Good, Good 
Rate, and Decisive Win Rate; lower is better for Loss and Both Bad.}
\label{tab:human_eval_summary}
\resizebox{\textwidth}{!}{%
\begin{tabular}{lccccccc}
\toprule
Model  & Win (\%) & Loss (\%) & Both Good (\%) & Both Bad (\%) & Good Rate (\%) & Decisive Win (\%) \\
\midrule
Gemini-3-Pro        & \textbf{85.0} & \textbf{4.1} & 10.6 & \textbf{0.3} & \textbf{95.6} & \textbf{95.4} \\
GPT-5.2             & 64.2 & 18.8 & \textbf{12.1} & 4.9 & 76.3 & 77.3 \\
\model (SFT+RL)           & 46.0 & 34.7 & 9.4 & 9.9 & 55.5 & 57.0 \\
Claude-Sonnet-4.6   & 48.4 & 39.5 & 6.3 & 5.8 & 54.7 & 55.1 \\
\model (SFT)          & 42.3 & 40.7 & 8.3 & 8.6 & 50.7 & 51.0 \\
Qwen3-VL-4B         & 17.5 & 75.7 & 0.2 & 6.6 & 17.7 & 18.8 \\
StarVector          & 2.7  & 93.3 & 0.4 & 3.6 & 3.1  & 2.8 \\
\bottomrule
\end{tabular}%
}
\end{table*}

\begin{table*}[t]
\centering
\caption{Pairwise human evaluation results between model pairs. Rates are reported over all pairwise annotations for each model pair. ``Decisive First Win'' excludes ties, i.e., both-good and both-bad cases.}
\label{tab:human_eval_pairwise}
\resizebox{\textwidth}{!}{%
\begin{tabular}{lccccc}
\toprule
Model Pair & First Win & Second Win & Both Good & Both Bad & Decisive First Win \\
\midrule
Gemini-3-Pro vs GPT-5.2             & 63.7 & 6.0  & 29.9 & 0.5  & 91.4 \\
\rowcolor{blue!8}
Gemini-3-Pro vs \model (SFT+RL)           & 82.6 & 4.5  & 12.4 & 0.5  & 94.9 \\
Gemini-3-Pro vs Claude-Sonnet-4.6   & 82.3 & 5.6  & 11.6 & 0.5  & 93.7 \\
Gemini-3-Pro vs \model (SFT)          & 82.8 & 8.1  & 9.1  & 0.0  & 91.1 \\
Gemini-3-Pro vs Qwen3-VL-4B         & 99.0 & 0.5  & 0.0  & 0.5  & 99.5 \\
Gemini-3-Pro vs StarVector          & 99.5 & 0.0  & 0.5  & 0.0  & 100.0 \\
\rowcolor{blue!8}
GPT-5.2 vs \model (SFT+RL)                & 53.2 & 17.4 & 16.9 & 12.4 & 75.4 \\
GPT-5.2 vs Claude-Sonnet-4.6        & 69.2 & 14.6 & 11.1 & 5.1  & 82.5 \\
GPT-5.2 vs \model (SFT)               & 65.2 & 14.1 & 14.1 & 6.6  & 82.2 \\
GPT-5.2 vs Qwen3-VL-4B              & 95.5 & 1.0  & 0.0  & 3.5  & 99.0 \\
GPT-5.2 vs StarVector               & 96.5 & 1.5  & 0.5  & 1.5  & 98.5 \\
\rowcolor{blue!8}
\model (SFT+RL) vs Claude-Sonnet-4.6      & 41.9 & 41.9 & 7.6  & 8.6  & 50.0 \\
\rowcolor{blue!8}
\model (SFT+RL) vs \model (SFT)             & 39.9 & 25.3 & 18.7 & 16.2 & 61.2 \\
\rowcolor{blue!8}
\model (SFT+RL) vs Qwen3-VL-4B            & 81.6 & 2.0  & 0.5  & 15.9 & 97.6 \\
\rowcolor{blue!8}
\model (SFT+RL) vs StarVector             & 91.4 & 2.5  & 0.5  & 5.6  & 97.3 \\
Claude-Sonnet-4.6 vs \model (SFT)     & 42.4 & 38.9 & 7.1  & 11.6 & 52.2 \\
Claude-Sonnet-4.6 vs Qwen3-VL-4B    & 89.4 & 3.0  & 0.0  & 7.6  & 96.7 \\
Claude-Sonnet-4.6 vs StarVector     & 96.5 & 1.5  & 0.5  & 1.5  & 98.5 \\
\model (SFT) vs Qwen3-VL-4B           & 80.3 & 11.1 & 0.5  & 8.1  & 87.8 \\
\model (SFT) vs StarVector            & 87.4 & 3.0  & 0.5  & 9.1  & 96.6 \\
Qwen3-VL-4B vs StarVector           & 88.4 & 7.6  & 0.0  & 4.0  & 92.1 \\
\bottomrule
\end{tabular}%
}
\end{table*}

\section{Additional Qualitative Results}
\label{appendix:results}
\begin{figure}[t]
    \centering
    \includegraphics[width=1\linewidth]{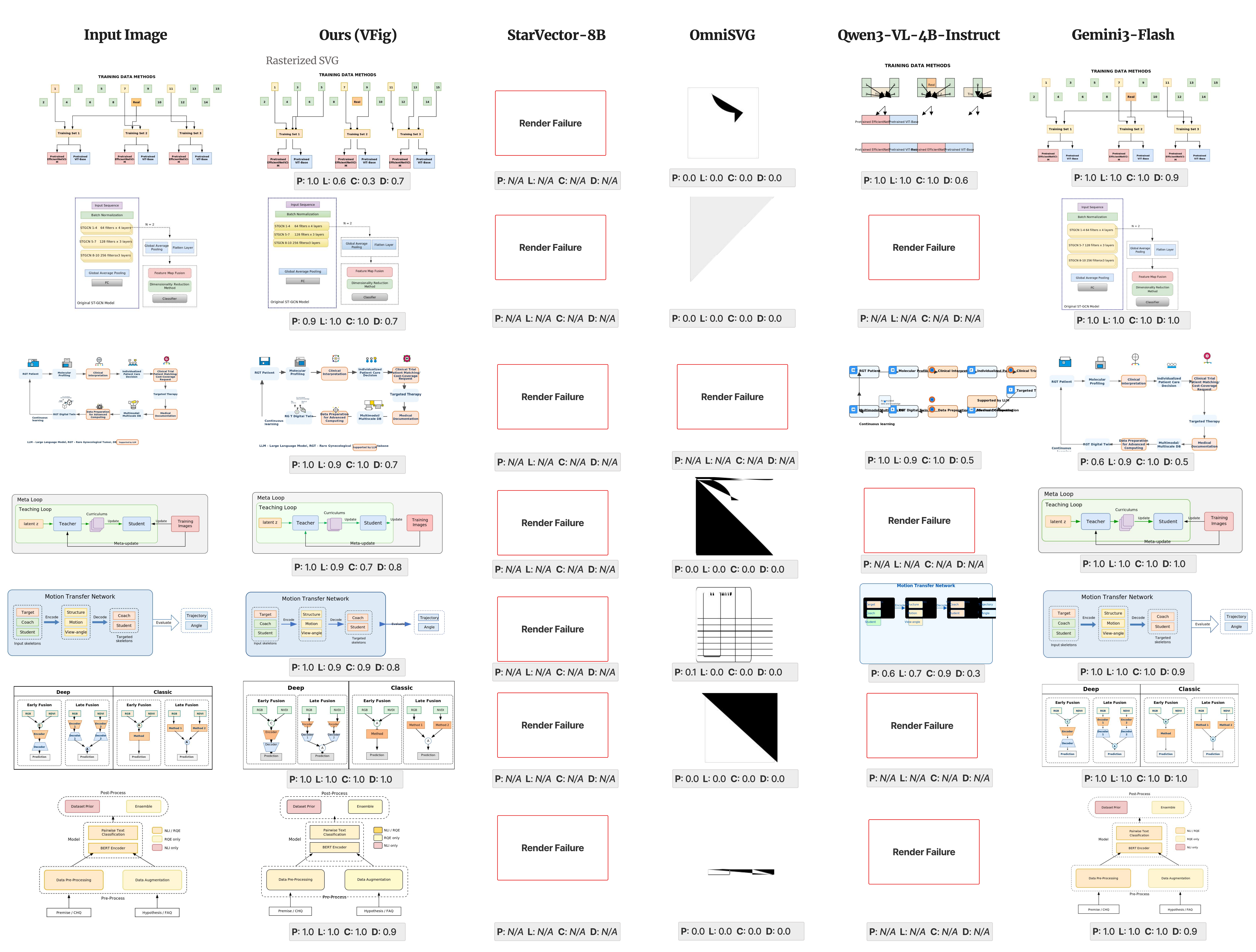}
    \caption{\textbf{Additional successful SVG generation examples on \model{}-Bench.} Given the same input raster image, we compare the rendered SVG outputs produced by different methods. Our model more faithfully preserves the structure of the input diagram. P/L/C/D denote the Gemini judge scores for presence, layout, connectivity, and details.}
    \label{fig:vfig_bench_example_appendix}
\end{figure}

We provide additional qualitative results on all three evaluation sources in \model{}-Bench, Molmo2-Diagram, and SVG-Diagram. Figures~\ref{fig:vfig_bench_example_appendix}, \ref{fig:molmo_general_example_appendix}, and \ref{fig:starvector_general_example_appendix} show representative examples generated by \model{} on the three benchmarks, respectively. All examples are drawn from the test sets.

Overall, these examples highlight both the strengths and limitations of our model. On the positive side, \model{} preserves the global diagram structure and layout well, so the rendered outputs often appear visually similar to the input figures. In particular, the model is generally effective at maintaining the high-level organization of objects and their spatial relationships. 

On the \model{}-Bench examples (Figure~\ref{fig:vfig_bench_example_appendix}), our model significantly improves rendering success compared with the base Qwen3-VL-4B-Instruct. The base model frequently produces invalid or non-renderable SVG outputs, whereas \model{} produces valid SVG programs for most inputs and generates diagrams that more closely resemble the original figures. In contrast, models such as StarVector-8B and OmniSVG struggle to handle these complex scientific diagrams, often producing repetition in SVG elements, noisy primitives, or outright rendering failures. \model{} therefore substantially improves robustness for complex figure reconstruction. At the same time, strong closed-source models such as Gemini3-Flash achieve very high visual fidelity on this benchmark, suggesting that further improvements are still possible. 

On Molmo2-Diagram (Figure~\ref{fig:molmo_general_example_appendix}) and SVG-Diagram (Figure~\ref{fig:starvector_general_example_appendix}), we observe a similar overall trend: compared with open-source baselines, \model{} more consistently produces valid SVG outputs and better preserves the global structure, layout, and major element relationships of the input images. These results are notable because both benchmarks differ from the complex scientific figures in \model{}-Bench. In particular, Molmo2-Diagram contains many stylized and more cartoon-like diagrams, while SVG-Diagram includes sparse geometric graphics and, in some cases, images that are not canonical diagrams. Despite this distribution shift, \model{} still reconstructs visually faithful outputs in many cases and remains robust across a broad range of diagram styles.

At the same time, the qualitative examples also highlight benchmark-specific behavior.On Molmo2-Diagram, \model{} generally preserves high-level layout, grouping, and major visual structure well across a diverse range of examples, including line plots, simple object illustrations, and table-like summary graphics. On SVG-Diagram, it is often effective at maintaining relative positions, edge directions, and simple topology, and it can also handle some non-diagram graphics reasonably well. Strong closed-source models such as Gemini3-Flash still recover finer text, styling, and local details more accurately. We also note that StarVector-8B performs worse in our qualitative comparison than what may be expected from its original paper. In our evaluation, we follow the default decoding parameters provided in the official StarVector repository, which differ from the inference configuration described in the paper. This difference in decoding setup, along with possible differences in prompting, may partially explain the discrepancy relative to the results reported in the original work.

\section{Failure Cases and Limitations}
\label{appendix: limitation}

\begin{figure}[t]
    \centering
    \includegraphics[width=1\linewidth]{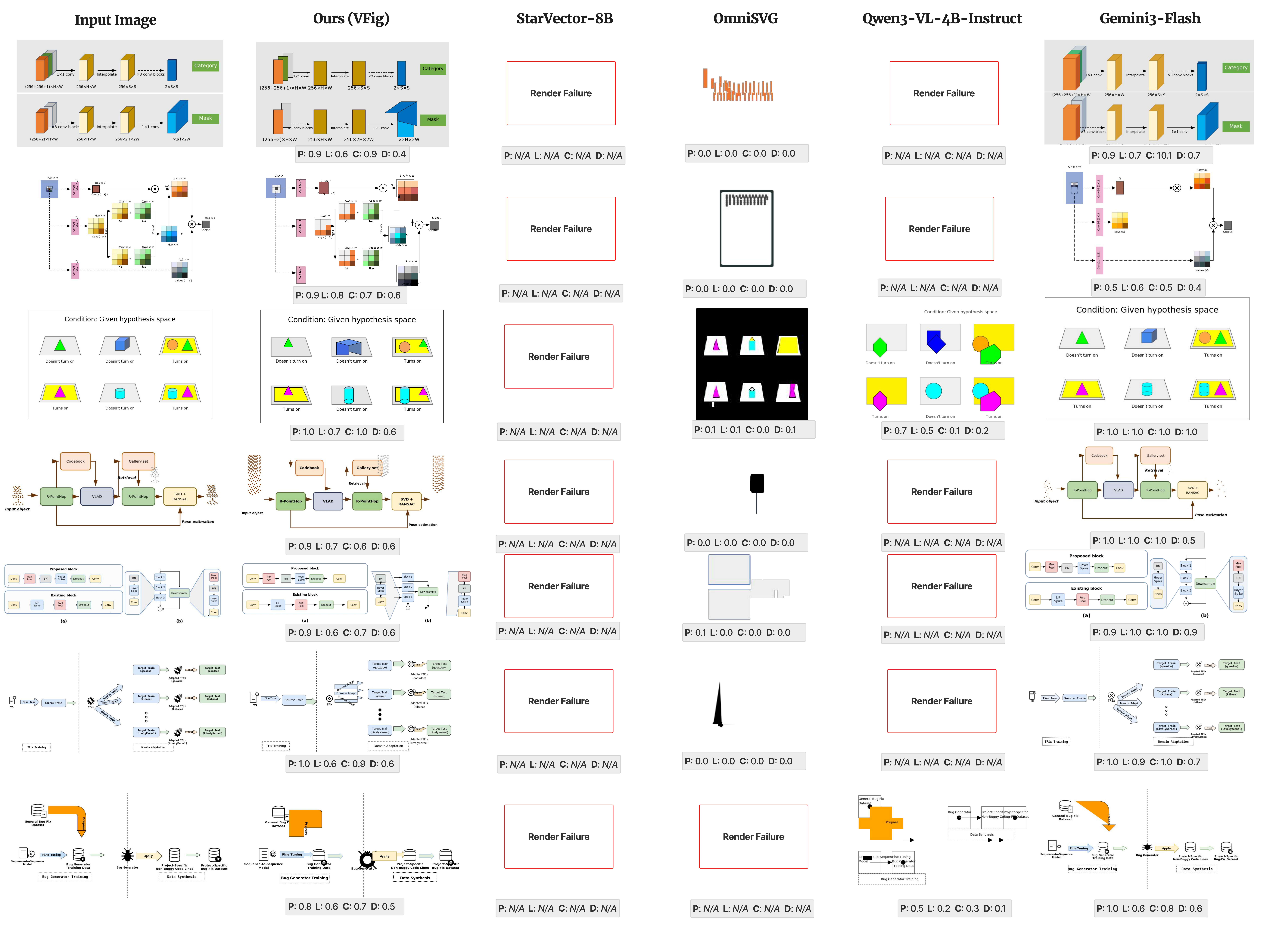}
    \caption{\textbf{Failure cases of \model{} on \model{}-Bench.}}
    \label{fig:vfig_bench_failure_examples}
\end{figure}

\begin{figure}[t]
    \centering
    \includegraphics[width=1\linewidth]{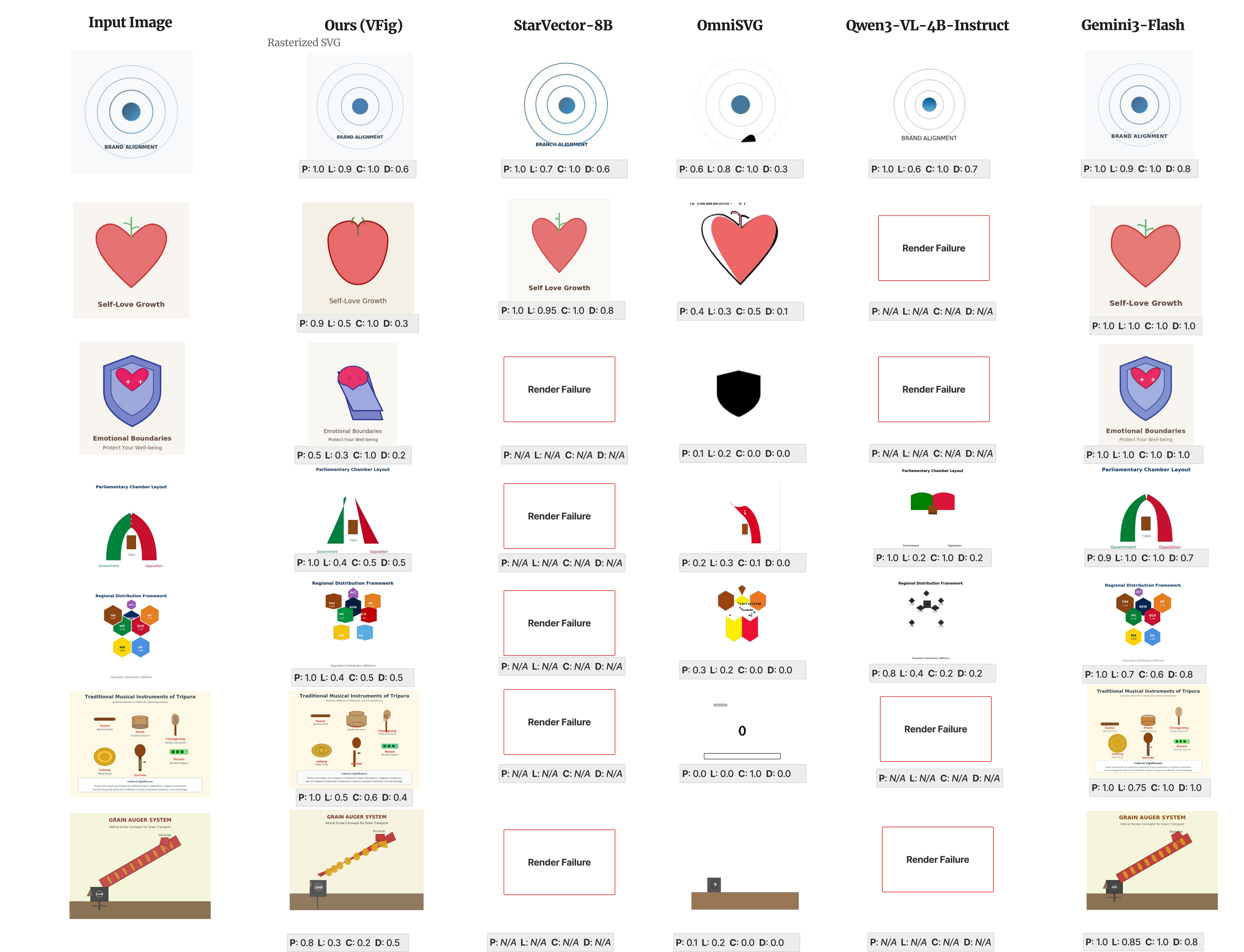}
    \caption{\textbf{Failure cases of \model{} on Molmo2-Diagram.}}
    \label{fig:molmo_failure_examples}
\end{figure}

\begin{figure}[t]
    \centering
    \includegraphics[width=1\linewidth]{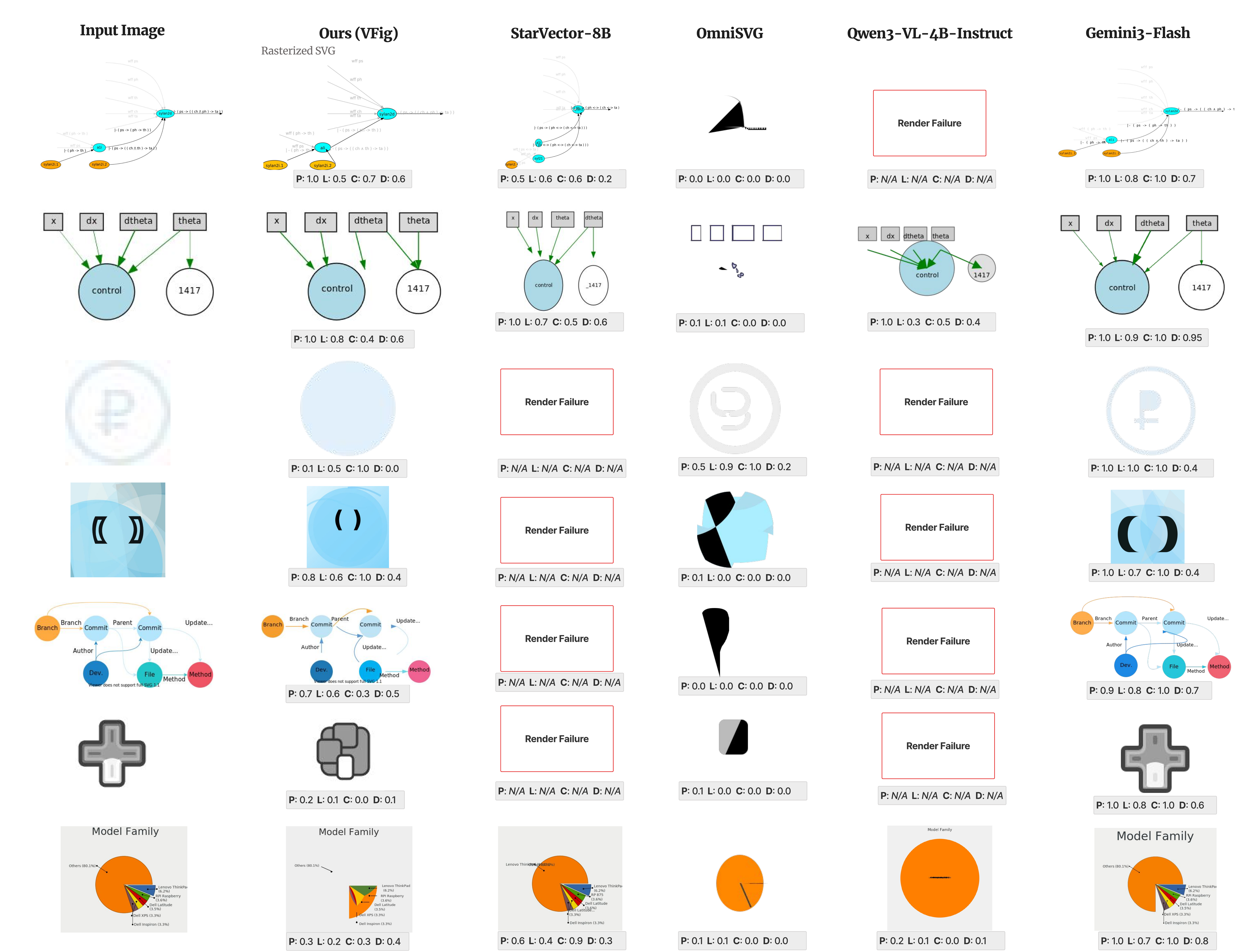}
    \caption{\textbf{Failure cases of \model{} on SVG-Diagram.}}
    \label{fig:starvector_failure_examples}
\end{figure}

Across the three benchmarks, several common failure patterns can be observed. Errors are often observed in thin lines, arrows, small text-like elements, and other precise local structures. The model may also fail to reproduce exact colors or subtle stylistic details. Figures~\ref{fig:vfig_bench_failure_examples}, \ref{fig:molmo_failure_examples}, and \ref{fig:starvector_failure_examples} present representative failure cases across the three benchmarks.

\noindent\textbf{Failure cases.} On the \model{}-Bench examples (Figure~\ref{fig:vfig_bench_failure_examples}), failures are primarily associated with fine-grained geometric details. In particular, the Gemini \textit{detail} metric is consistently the lowest among the four evaluation dimensions, indicating that local visual fidelity remains the main limitation of the model. Common errors include inaccurate arrowheads, distorted thin connectors, and incorrect rendering of small geometric components. Diagrams containing 3D shapes or perspective-like objects are especially challenging, often resulting in incorrect geometry or missing structural elements. These results suggest that while the model captures global layout and object relationships well, it still struggles to reproduce precise local visual structures.

On the Molmo2-Diagram and SVG-Diagram examples (Figure~\ref{fig:molmo_failure_examples} and Figure~\ref{fig:starvector_failure_examples}), we observe both shared and benchmark-specific failure patterns. Similar to \model{}-Bench, many errors remain concentrated in fine-grained local details, such as thin lines, arrowheads, small annotations, and other precise visual structures. Even when the overall layout and major object relationships are largely preserved, these local inaccuracies can noticeably reduce visual fidelity.

These two benchmarks also expose a broader range of challenging inputs beyond canonical diagrams. In Molmo2-Diagram, failure cases often appear on more stylized examples and visually simplified graphics. In SVG-Diagram, the model can struggle on sparse geometric graphics as well as inputs that resemble logos, icons, simple object drawings, or chart-like visuals. In such cases, \model{} may preserve only the coarse structure while simplifying shapes, missing small components, or distorting local geometry and styling. Overall, these examples suggest that the model's generalization across diverse graphic styles and element types remains limited, especially when the inputs deviate from the structured diagram distributions emphasized during training.

\noindent\textbf{Possible causes of failure.}
Several factors may contribute to the failure cases discussed above. First, our final reward design is based on the empirical observation in the main paper that the fully VLM-based reward performs better than the variants we tested with additional pixel-level objectives. However, this does not imply that pixel-level supervision is inherently unhelpful. In our current experiments, we only explored a limited set of pixel-level reward variants, namely equal-weight combinations of Gemini-based reward with L2, Canny-based L2, and SSIM-style image similarity terms. It remains possible that different types of pixel-level rewards, or different weighting ratios between structural and pixel-level objectives, could better improve fine-grained visual fidelity without sacrificing judge-based quality.

Second, our reward relies heavily on Gemini as a visual judge. Although Gemini-based judgment correlates well with human evaluation in our experiments, it is still an imperfect supervisory signal. Its preferences may not always align with human perception on every example, especially for subtle stylistic details, local geometry, or borderline rendering cases. Moreover, our current formulation assigns equal weights to the four rubric components---\emph{presence}, \emph{layout}, \emph{connectivity}, and \emph{details}. This equal-weight design is simple and effective in practice, but it is not necessarily the best approximation to human perceptual preferences: human judgments of figure quality may not correspond to a uniform average over these four aspects. Likewise, the specific judge prompt and rubric design may influence which errors are emphasized during training, and there may be additional dimensions of quality not fully captured by the current decomposition. Exploring alternative component weights, revised judge prompts, or richer reward dimensions could therefore further improve alignment between reward optimization and human judgments.

Third, our training data is primarily designed for complex diagram reconstruction rather than broad vector graphic generation. As a result, the data contains limited coverage of examples such as icons, sketches, simple object drawings, logos, and other non-canonical graphics. This mismatch is consistent with the qualitative failures observed on Molmo2-Diagram and SVG-Diagram, where some difficult examples fall outside the core distribution emphasized during training. Together, these observations suggest that further improvements may require both richer reward design and broader training data coverage beyond structured scientific diagrams.

\begin{figure}[t]
    \centering
    \includegraphics[width=1\linewidth]{sections//figure/molmo_general_examples_new.pdf}
    \caption{\textbf{Successful SVG generation examples on Molmo2-Diagram.}}
    \label{fig:molmo_general_example_appendix}
\end{figure}

\begin{figure}[t]
    \centering
    \includegraphics[width=1\linewidth]{sections//figure/starvector_general_new.pdf}
    \caption{\textbf{Successful SVG generation examples on SVG-Diagram.}}
    \label{fig:starvector_general_example_appendix}
\end{figure}

\section{Extended Related Work}
\label{appendix:extended_related_work}
This section summarizes earlier non-LLM approaches to vector graphics generation that are not the focus of the main related-work discussion (Sec.~\ref{sec: related_work}).

\noindent\textbf{Tracing-based Vectorization.}
Classical methods trace contours from raster to vector, from early systems \cite{diebel2008bayesian,xia2009patch,liao2012subdivision} to Potrace's Bézier fitting \cite{selinger2003potrace} and modern tools like VTracer with color handling and heuristic curve fitting \cite{vtracer}. They optimize contour accuracy but produce unconstrained Bézier paths rather than editable primitives (boxes, arrows, text), and cannot enforce alignment or other domain constraints required for diagram generation.

\noindent\textbf{Neural Vector Graphics Models.}
Non-LLM methods model vector graphics directly or via differentiable rendering: SVG-VAE~\cite{Lopes_2019_ICCV} uses sequential decoding, DeepSVG~\cite{deepsvg} learns a hierarchical generative model with latent-space editing, DiffVG~\cite{diffvg} propagates gradients from raster losses to vector parameters (which Im2Vec~\cite{im2vec} exploits to synthesize vectors without vector supervision), and LIVE~\cite{Ma_2022_CVPR} adds progressive layer-wise construction; related work spans font/sketch generation and learned super-vectorization~\cite{hu2024supersvg,wang2021deepvecfont,ha2017neuralrepresentationsketchdrawings}. These methods improve flexibility over tracers but struggle with semantic primitive selection and global structure in complex SVGs.

\newpage
\section*{NeurIPS Paper Checklist}

\begin{enumerate}

\item {\bf Claims}
    \item[] Question: Do the main claims made in the abstract and introduction accurately reflect the paper's contributions and scope?
    \item[] Answer: \answerYes{}
    \item[] Justification: The abstract and Sec.~\ref{sec:intro} state our contributions (\dataset, \bench, \benchood, and \model) and report concrete results that match the experiments in Sec.~\ref{sec: experiments}.

\item {\bf Limitations}
    \item[] Question: Does the paper discuss the limitations of the work performed by the authors?
    \item[] Answer: \answerYes{}
    \item[] Justification: The paper discusses limitations in section~\ref{sec:conclusion} and includes a dedicated ``Failure Cases and Limitations'' section in the Appendix that discusses remaining failure modes on out-of-distribution diagrams and the limited training-data coverage of non-canonical graphics.

\item {\bf Theory assumptions and proofs}
    \item[] Question: For each theoretical result, does the paper provide the full set of assumptions and a complete (and correct) proof?
    \item[] Answer: \answerNA{}
    \item[] Justification: The paper does not include theoretical results; the contributions are empirical (datasets, benchmark, and an open-source model trained with SFT and RL).

    \item {\bf Experimental result reproducibility}
    \item[] Question: Does the paper fully disclose all the information needed to reproduce the main experimental results of the paper to the extent that it affects the main claims and/or conclusions of the paper (regardless of whether the code and data are provided or not)?
    \item[] Answer: \answerYes{}
    \item[] Justification: Sec.~\ref{sec: method} and Sec.~\ref{sec:exp_setup} describe the SFT/GRPO recipe and rubric reward, Sec.~\ref{sec: dataset} describes the data construction pipeline, and Appendix~\ref{appendix:setup} provides full hyperparameters; we have publicly released \dataset, \bench, \benchood, and the \model{} model checkpoints.

\item {\bf Open access to data and code}
    \item[] Question: Does the paper provide open access to the data and code, with sufficient instructions to faithfully reproduce the main experimental results, as described in supplemental material?
    \item[] Answer: \answerYes{}
    \item[] Justification: \dataset, \bench, \benchood{}, and \model{} model checkpoints have been released publicly (HuggingFace) with a Croissant metadata record; the conclusion (Sec.~\ref{sec:conclusion}) commits to releasing models, data, and evaluation tools.

\item {\bf Experimental setting/details}
    \item[] Question: Does the paper specify all the training and test details (e.g., data splits, hyperparameters, how they were chosen, type of optimizer) necessary to understand the results?
    \item[] Answer: \answerYes{}
    \item[] Justification: Sec.~\ref{sec:exp_setup} summarizes backbones, LoRA configuration, GRPO setup, baselines, benchmarks, and metrics; full hyperparameters (LR, sequence lengths, KL/entropy coefficients, batch sizes, schedulers, GPU counts, wall-clock times) are tabulated in Appendix~\ref{appendix:setup}.

\item {\bf Experiment statistical significance}
    \item[] Question: Does the paper report error bars suitably and correctly defined or other appropriate information about the statistical significance of the experiments?
    \item[] Answer: \answerYes{}
    \item[] Justification: Each training configuration is run once due to the cost of full SFT and GRPO on multi-billion-parameter VLMs; the only stochastic estimate, the human-eval Elo, is reported with bootstrap-averaged uncertainty (Table~\ref{tab:human_eval_elo} in Appendix~\ref{appendix:human_evaluations}). Reported automatic metrics are deterministic on a fixed test set under greedy decoding.

\item {\bf Experiments compute resources}
    \item[] Question: For each experiment, does the paper provide sufficient information on the computer resources (type of compute workers, memory, time of execution) needed to reproduce the experiments?
    \item[] Answer: \answerYes{}
    \item[] Justification: Sec.~\ref{sec:exp_setup} states SFT uses $5\times$L40S GPUs and RL uses $4\times$L40S; Appendix~\ref{appendix:setup} reports per-backbone wall-clock training times (\eg, $\sim$30h for RL on Qwen3-VL-4B; multi-day SFT runs per backbone) and batch/sequence configurations.
    
\item {\bf Code of ethics}
    \item[] Question: Does the research conducted in the paper conform, in every respect, with the NeurIPS Code of Ethics \url{https://neurips.cc/public/EthicsGuidelines}?
    \item[] Answer: \answerYes{}
    \item[] Justification: We have reviewed the NeurIPS Code of Ethics and conform with it. Source figures are drawn from publicly available arXiv papers, no human-subject risks are introduced, and anonymity is preserved in the submission.

\item {\bf Broader impacts}
    \item[] Question: Does the paper discuss both potential positive societal impacts and negative societal impacts of the work performed?
    \item[] Answer: \answerYes{}
    \item[] Justification: Sec.~\ref{sec:intro} discusses the positive impact of the figure-to-SVG task: faster revision of scientific diagrams, lower barriers to professional visualization, and faithful cross-platform reuse. We do not identify direct paths to negative societal impacts: figure-to-SVG generation poses no obvious dual-use, fairness, privacy, or safety risks beyond those of generic vision-language models.
    
\item {\bf Safeguards}
    \item[] Question: Does the paper describe safeguards that have been put in place for responsible release of data or models that have a high risk for misuse (e.g., pre-trained language models, image generators, or scraped datasets)?
    \item[] Answer: \answerNA{}
    \item[] Justification: The released artifacts are figure-to-SVG models and curated arXiv-figure datasets. They do not produce natural images, faces, or generative content with elevated misuse risk, so no additional safeguards beyond standard licensing are required.

\item {\bf Licenses for existing assets}
    \item[] Question: Are the creators or original owners of assets (e.g., code, data, models), used in the paper, properly credited and are the license and terms of use explicitly mentioned and properly respected?
    \item[] Answer: \answerYes{}
    \item[] Justification: We cite all external assets used: backbone VLMs (Qwen3-VL, Qwen2.5-VL, InternVL3.5), training datasets (StarVector's SVG-Diagrams, Molmo2-Diagram, Paper2Fig), proprietary judge/reward APIs (Gemini-3, GPT-5.2), and the Semantic Scholar API for OOD curation. Source arXiv papers used for \compds{} and \benchood{} are publicly available preprints; we comply with their terms of use.

\item {\bf New assets}
    \item[] Question: Are new assets introduced in the paper well documented and is the documentation provided alongside the assets?
    \item[] Answer: \answerYes{}
    \item[] Justification: We release \dataset{} (66K filtered figure--SVG pairs), \bench{} (held-out test split), \benchood{} (198 manually curated OOD figures), and the \model{} model checkpoints. Datasets include a Croissant metadata record (CC-BY 1.0) with field-level documentation; data construction details are in Sec.~\ref{sec: dataset} and Appendix.

\item {\bf Crowdsourcing and research with human subjects}
    \item[] Question: For crowdsourcing experiments and research with human subjects, does the paper include the full text of instructions given to participants and screenshots, if applicable, as well as details about compensation (if any)?
    \item[] Answer: \answerYes{}
    \item[] Justification: The pairwise human evaluation on \benchood{} is summarized in Sec.~\ref{sec:exp_setup} (Metrics) and described in detail in Appendix~\ref{appendix:human_evaluations}, including the evaluation interface, instructions, full Elo summary, per-pair pairwise breakdown, and decisive-win heatmap. Annotators were the authors and consenting collaborators (no external paid crowdsourcing).

\item {\bf Institutional review board (IRB) approvals or equivalent for research with human subjects}
    \item[] Question: Does the paper describe potential risks incurred by study participants, whether such risks were disclosed to the subjects, and whether Institutional Review Board (IRB) approvals (or an equivalent approval/review based on the requirements of your country or institution) were obtained?
    \item[] Answer: \answerNA{}
    \item[] Justification: The human evaluation involves only the authors and consenting internal collaborators viewing publicly available scientific figures and rendered SVG outputs, with no personal data, no sensitive content, and no risks beyond ordinary screen viewing. No external crowdsourcing was used, so IRB approval was not required.

\item {\bf Declaration of LLM usage}
    \item[] Question: Does the paper describe the usage of LLMs if it is an important, original, or non-standard component of the core methods in this research? Note that if the LLM is used only for writing, editing, or formatting purposes and does \emph{not} impact the core methodology, scientific rigor, or originality of the research, declaration is not required.
    \item[] Answer: \answerYes{}
    \item[] Justification: LLMs are core to our methodology and are explicitly described: Gemini-3-Pro powers the describe-and-generate data pipeline (Sec.~\ref{sec:data_generation}), Gemini-3-Flash provides the rubric reward during RL (Sec.~\ref{sec:rl_reward}) and serves as one of two VLM judges, GPT-5.2 serves as the second judge (Sec.~\ref{sec:exp_setup}), and the trained \model{} family is itself fine-tuned from open-source VLMs (Qwen3-VL, Qwen2.5-VL, InternVL3.5).

\end{enumerate}


\end{document}